\documentclass[10pt,twocolumn,letterpaper]{article}

\usepackage[pagenumbers]{cvpr}

\usepackage{placeins}
\usepackage{stfloats}

\usepackage{graphicx,multirow,colortbl}

\definecolor{mkhi4d}{HTML}{BBD3EE}\definecolor{mkhi4dl}{HTML}{E1ECF8}
\definecolor{mkperp}{HTML}{F6CFA3}\definecolor{mkperpl}{HTML}{FCEBD7}
\definecolor{mksyn}{HTML}{D3C5EC}\definecolor{mksynl}{HTML}{EDE7F6}
\definecolor{mkcharge}{HTML}{F2BCCB}\definecolor{mkchargel}{HTML}{FBE4EA}
\definecolor{mkmcr}{HTML}{C9BEE8}\definecolor{mkmcrl}{HTML}{E9E4F6}
\definecolor{mkkub}{HTML}{BFE3D4}\definecolor{mkkubl}{HTML}{E2F3EC}
\definecolor{mkdex}{HTML}{F3E3A6}\definecolor{mkdexl}{HTML}{FAF2D6}
\definecolor{mkavg}{HTML}{D2D2D2}\definecolor{mkavgl}{HTML}{ECECEC}
\definecolor{mkgd}{HTML}{BBD3EE}\definecolor{mkgdl}{HTML}{E1ECF8}
\definecolor{mkgm}{HTML}{F6CFA3}\definecolor{mkgml}{HTML}{FCEBD7}
\definecolor{mkgs}{HTML}{D3C5EC}\definecolor{mkgsl}{HTML}{EDE7F6}
\definecolor{mkbest}{HTML}{B7E3B0}\definecolor{mksecond}{HTML}{E3F4DF}
\definecolor{mkours}{HTML}{FFF7D6}\definecolor{mkpanel}{HTML}{E8E8E8}\definecolor{mkfail}{HTML}{C62828}
\newcommand{\mkB}[1]{\cellcolor{mkbest}\textbf{#1}}
\newcommand{\mkS}[1]{\cellcolor{mksecond}#1}

\newcommand{\mksl}{\hspace{1pt}/\hspace{1pt}} 
\newcommand{\mksetup}{\centering\setlength{\aboverulesep}{0pt}\setlength{\belowrulesep}{0pt}\setlength{\extrarowheight}{1.5pt}}

\definecolor{cvprblue}{rgb}{0.21,0.49,0.74}
\usepackage[pagebackref,breaklinks,colorlinks,allcolors=cvprblue]{hyperref}

\title{Jojo4D: Bundle Adjustment Strikes Back in Dynamic Scenes}

\author{%
Johannes Weidenfeller \quad
Frano Raji\v{c} \quad
Dante Boeri \quad
Sergey Prokudin \quad
Siyu Tang\\[0.75ex]
{\small ETH Z\"urich}
}

\begin{document}
\twocolumn[{%
\renewcommand\twocolumn[1][]{#1}%
\maketitle
\begin{center}
    \centering
    \captionsetup{type=figure}
    \vspace{-0.81cm}
    \includegraphics[width=\textwidth]{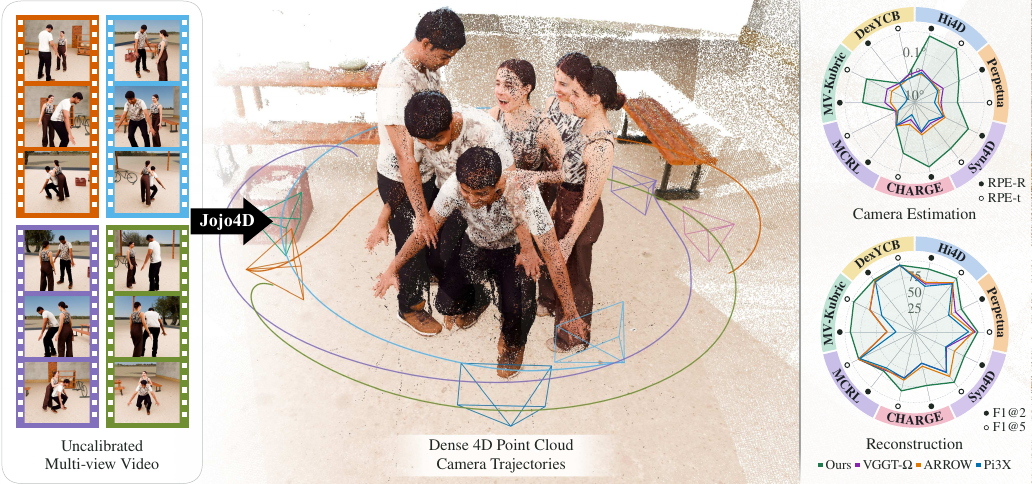}
    \captionof{figure}{\textbf{Jojo4D} takes synchronized, uncalibrated videos from static or moving cameras (left) and recovers camera trajectories and dense, time-varying geometry (middle), outperforming baselines in camera estimation and reconstruction across seven datasets (right).}
    \label{fig:teaser}
\end{center}%
}]
\begin{abstract}
Accurate 4D reconstruction from a few cameras offers a flexible way to reconstruct dynamic scenes for spatial measurement and robotic interaction. Cameras may be stationary, moving, or mixed, with unknown calibration, making joint recovery of camera parameters and evolving scene geometry in a shared coordinate frame essential. Recent feed-forward models make dense reconstruction from dynamic multi-camera video increasingly accessible, but plausible predictions alone do not ensure geometric consistency. Optimization-based methods further refine dynamic multi-camera reconstructions, yet substantial errors remain in camera estimation and scene geometry. We introduce \textbf{Jojo4D}, a robust framework for reconstructing dynamic scenes from as few as two synchronized, uncalibrated cameras. Starting from feed-forward estimates, \textbf{Jojo4D} jointly refines camera intrinsics, poses, and sparse geometry through robust bundle adjustment on tracks built from dense matches. These tracks exploit complementary constraints: synchronized views constrain static and moving surfaces alike, while stationary points connect observations across time. With the refined cameras fixed, we triangulate reliable anchors and align feed-forward point maps to them, extending reconstruction into regions without reliable matches. To assess both camera recovery and dense reconstruction, we establish a unified benchmark under a common protocol across seven real and synthetic datasets with varying camera counts and motion. \textbf{Jojo4D} substantially reduces camera estimation errors and delivers higher average dense-reconstruction F1 scores than feed-forward and optimization-based baselines. Project page: \url{https://ethz-vlg.github.io/jojo4d/}.
\end{abstract}

\section{Introduction}
\label{sec:introduction}

Consider a robot manipulation sequence recorded by a fixed camera overlooking a tabletop and a moving camera mounted on the robot. The external camera provides context, while the moving camera reveals details and surfaces that may otherwise be occluded. A similar need arises when two moving cameras record a human performance or interaction. In both cases, observing the same surfaces from different viewpoints at the same instant provides geometric constraints on their depth, without requiring those surfaces to stay still between frames. Reconstructing these scenes in 3D gives robots a geometric understanding of their surroundings~\cite{kerr2024rsrd}, supports rendering dynamic scenes from new viewpoints~\cite{wu2024fourdgs}, and provides training data for video generation and world models~\cite{ren2025gen3c,huang2026pointworld}.

For uncalibrated videos, realizing this potential requires estimating the cameras alongside the changing scene geometry. Classical structure-from-motion recovers cameras and geometry from \emph{static} scene points~\cite{schonberger2016colmap,pataki2026vidmap}, treating matches on moving objects as outliers rather than reconstructing their motion. \emph{Monocular} \mbox{methods recover cameras} and per-frame geometry in dynamic scenes~\cite{li2025megasam,huang2025vipe}, but do not jointly exploit observations from several cameras.

Recent feed-forward models~\cite{wang2026vggtomega,fradlin2026arrow,jiang2026omnix} reconstruct dynamic scenes from \emph{multi-camera} video. In this setting, Sun~\etal~\cite{sun2026multiview} refine camera poses and dense depth maps by balancing correspondence errors against deviations from monocular depth predictions. Despite these advances, substantial errors remain in camera calibration and reconstructed scene geometry.

To improve camera and reconstruction accuracy, we introduce \textbf{Jojo4D}, a method that estimates camera intrinsics, camera poses, and dense point clouds from synchronized, uncalibrated videos captured by static, moving, or mixed camera setups. We initialize the cameras with aligned feed-forward predictions~\cite{fradlin2026arrow} and refine them using sparse tracks built from RoMa~v2's correspondences across cameras and time~\cite{edstedt2026romav2}. These tracks provide two complementary constraints. Matches between cameras at the same \emph{instant} constrain their relative poses through both static and moving scene points, while tracks within each video connect observations across time. Joint bundle adjustment refines the cameras and one free 3D point per track, rather than dense depth maps regularized toward a learned prediction~\cite{sun2026multiview}. Strict reprojection thresholds suppress inconsistent observations without requiring a mask of moving objects.

With the refined cameras held fixed, we reconstruct the scene at each instant by triangulating anchor points where multiple matches agree on depth. We complete regions without reliable matches using dense feed-forward predictions that we align to nearby anchors in 3D. Both the feed-forward models and image matchers can be replaced, allowing Jojo4D to benefit from advances in either component.

To evaluate camera estimation and dense reconstruction together, we assemble a benchmark spanning seven real and synthetic datasets with varying camera counts and motion. We evaluate feed-forward and optimization-based methods under the same protocol on both tasks, across static, moving, and mixed camera configurations. Jojo4D substantially reduces relative camera rotation and translation-direction errors across datasets compared with feed-forward~and optimization-based baselines. For dense reconstruction, average point-cloud F1 improves over the best baseline from $65.9\%$ to $80.9\%$ at $2$\,cm and from $84.7\%$ to $91.5\%$ at $5$\,cm.

\section{Related Work}
\label{sec:related_work}

\paragraph{Static scene reconstruction.}
Classical structure-from-motion (SfM) recovers cameras and sparse geometry by matching image features and minimizing reprojection error through bundle adjustment~\cite{triggs1999bundle,schonberger2016colmap,pan2024glomap}. Learned matching~\cite{sarlin2020superglue,lindenberger2023lightglue,edstedt2024roma}, monocular depth and normal priors~\cite{pataki2025mpsfm,zhu2026mba}, and differentiable optimization~\cite{teed2021droidslam,wang2024vggsfm} improve the robustness of camera estimation. With the cameras recovered, multi-view stereo~\cite{schonberger2016pixelwise,yao2018mvsnet,gu2020casmvsnet} or surface reconstruction through differentiable rendering~\cite{wang2021neus,huang2024twodgs,weidenfeller2026adele} can provide dense geometry. These pipelines constrain each reconstructed point to a fixed 3D position. Jojo4D retains this constraint for tracks across time but also uses same-instant tracks, which constrain the cameras and geometry without requiring the observed surface to remain stationary.

Feed-forward models recover cameras and dense geometry from images~\cite{wang2024dust3r,leroy2024mast3r,keetha2026mapanything,wang2025vggt,lin2026depthanything3}. Hybrid approaches combine these learned predictions with geometric refinement for SfM and SLAM~\cite{duisterhof2025mast3rsfm,pan2026gluemap,murai2025mast3rslam,maggio2025vggtslam,leroy2026blast3r,chen2026ggpt,pataki2026vidmap}. Both groups assume a static scene. Applying them independently at each instant does not enforce a shared coordinate frame over time. Jojo4D also combines learned predictions with geometric refinement, but jointly estimates cameras across views and time.

\paragraph{Dynamic scene reconstruction.}
Monocular methods recover cameras and scene geometry from dynamic videos. Optimization-based approaches mask or down-weight moving observations during camera estimation~\cite{zhao2022particlesfm,li2025megasam,huang2025vipe}, or separate object motion from camera-induced motion before bundle adjustment~\cite{chen2025batrack}. Learned reconstruction models predict per-frame geometry and, in some cases, point motion~\cite{zhang2025monst3r,wang2025cut3r,zhang2026d4rt,karhade2026any4d}. Reconstructing each video independently, however, does not enforce consistency across cameras. Jojo4D instead uses synchronized views to constrain moving surfaces at each instant and connect the cameras in a shared reconstruction.

Recent feed-forward models also support dynamic multi-camera input~\cite{wang2026pi3,wang2026vggtomega,jiang2026omnix}. ARROW~\cite{fradlin2026arrow} unifies reconstruction and tracking through point queries over observations from different viewpoints and times. Such models predict geometry even in image regions without reliable matches, but their reconstructions can be geometrically inconsistent across views. Jojo4D initializes cameras from these predictions, refines them using matches, and fills gaps in triangulated geometry with depth from the same models.

Closest to our setting, Sun~\etal~\cite{sun2026multiview} extend visual SLAM to multiple moving cameras using a spatiotemporal connection graph and feed-forward initialization. They then refine camera poses and depth maps using wide-baseline optical flow and monocular depth priors, with fixed input intrinsics. These priors stabilize refinement but can bias it when the predicted depths conflict with image correspondences. Jojo4D instead optimizes intrinsics, poses, and free 3D track points without a depth-prior term in bundle adjustment. We use feed-forward predictions for initialization and dense completion, not as targets for the bundle-adjusted points. Jojo4D substantially improves pose accuracy and dense reconstruction over Sun~\etal~(\cref{tab:cameras,tab:point_cloud_results}).

\paragraph{Reconstruction benchmarks.}
Existing multi-camera datasets cover diverse scenes and camera configurations~\cite{chao2021dexycb,yin2023hi4d,jiang2026syn4d,nazarczuk2026charge,sun2026multiview}. The TAPVid-MV benchmark~\cite{koppula2026tapvidmv} unifies point tracking, camera estimation, and reconstruction evaluation, but dense ground-truth depth is available for only three subsets. We assemble seven existing datasets with ground-truth cameras and depth, combining its Hi4D~\cite{yin2023hi4d} and Perpetua~\cite{koppula2026tapvidmv} subsets with Syn4D~\cite{jiang2026syn4d}, CHARGE~\cite{nazarczuk2026charge}, MultiCamRobolab~\cite{sun2026multiview}, MV-Kubric~\cite{rajic2025mvtracker}, and DexYCB~\cite{chao2021dexycb} for joint evaluation of camera estimation and dense reconstruction under a unified protocol.

\section{Method}
\label{sec:method}

\begin{figure*}[t]
    \centering
    \includegraphics[width=\linewidth]{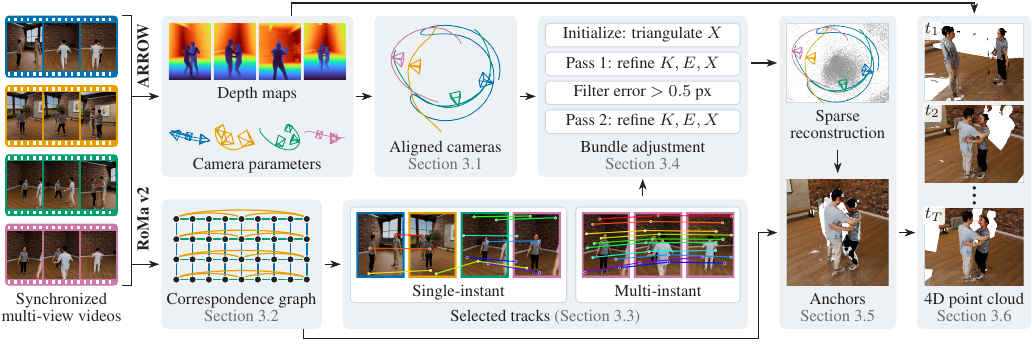}
    \caption{\textbf{Method overview.} Jojo4D takes synchronized, uncalibrated videos as input. We align per-video ARROW~\cite{fradlin2026arrow} camera predictions into a shared coordinate frame (\cref{sec:camera_init}) and build a RoMa~v2~\cite{edstedt2026romav2} correspondence graph (\cref{sec:correspondence_graph}), from which we select tracks (\cref{sec:track_selection}). Two bundle adjustment passes refine the cameras with reprojection-error filtering in between (\cref{sec:bundle_adjustment}). With the cameras fixed, we triangulate anchors from the dense matches (\cref{sec:anchor_triangulation}) and align feed-forward geometry to them, completing a point cloud \mbox{at each instant (\cref{sec:completion})}.}
    \label{fig:method}
    \vspace{-0.3cm}
\end{figure*}

Given $N$ synchronized, uncalibrated videos of a dynamic scene over $T$ time steps, Jojo4D estimates camera parameters and a dense point cloud at each time step in a shared coordinate frame. We denote the image from camera $i$ at time step $t$ by $I_{i,t}$ and call the set of images captured at the same time an \emph{instant}. \Cref{fig:method} outlines the pipeline. We initialize the cameras and construct a correspondence graph, then select tracks for joint bundle adjustment. With the refined cameras, we triangulate anchor points and complete the geometry using feed-forward predictions.

\subsection{Camera Initialization}
\label{sec:camera_init}

Bundle adjustment requires a reliable camera initialization. We first predict each camera's trajectory from its own video using ARROW~\cite{fradlin2026arrow}, processing chunks of at most 30 frames. Consecutive chunks share at least 15 frames, which we use to align their predictions into one trajectory per camera (see \cref{sec:chunk_stitching} for details). We identify static cameras from the rotation and translation in their predicted trajectories and replace each static trajectory by a single pose. These trajectories are initially expressed in separate coordinate frames, so they still need to be aligned across cameras.

To align these trajectories, we also run ARROW on all cameras at each instant. Every image then has two predicted poses, one from its video and one from its instant, which should agree once both are expressed in a shared frame. We therefore estimate one similarity transform per video and one per instant. Over all instants, we first robustly average rotations and reject inconsistent image pairs, then estimate scales from depth ratios and translations from camera centers using alternating medians. We retain the per-video trajectories, each aligned by one transform, and discard the auxiliary per-instant predictions (see \cref{sec:camera_alignment_suppl} for details).

We obtain one intrinsic matrix per camera by taking the element-wise median of its per-video ARROW estimates. Together with the aligned trajectories, these camera parameters initialize the bundle adjustment described in \cref{sec:bundle_adjustment}.

\subsection{Correspondence Graph}
\label{sec:correspondence_graph}
Camera refinement and dense reconstruction both rely on correspondences across views and over time. We represent the matched images as an undirected graph $\mathcal{G} = (\mathcal{V}, \mathcal{E})$, whose vertices are the frames $(i,t)$ and whose edges are the image pairs that we match. For every edge, we run RoMa~v2~\cite{edstedt2026romav2} bidirectionally, which yields a dense warp in each direction together with a per-pixel confidence.

Exhaustive matching includes $\binom{NT}{2}$ frame pairs, which quickly becomes infeasible. We therefore match only three kinds of frame pairs (see \cref{sec:matching_graph_suppl} for an illustration). First, we exhaustively match all camera pairs at each instant. Since these images capture the scene at the same time, their matches let us triangulate both static and moving points. Second, we connect consecutive frames of each video. Third, to bridge distant time steps without long chains of consecutive matches, we mark every fifth frame of a video as a keyframe and match it to the next five keyframes. Long-range matches connect keyframe indices $(0,8), (8,16), \ldots$, then $(0,16), (16,32), \ldots$, doubling the spacing at each level, so that any two keyframes are linked through $\mathcal{O}(\log T)$ matches. The resulting graph has $\mathcal{O}(N^2 T)$ edges and thus grows only linearly with the number of time steps.

\subsection{Track Selection}
\label{sec:track_selection}

We group the dense pairwise matches into a sparse set of tracks for bundle adjustment, each representing observations of one scene point. For an edge between two frames $a, b \in \mathcal{V}$, let $W_{a \to b}(p)$ denote the location in frame $b$ that RoMa~v2 predicts for a pixel $p$ of frame $a$, and let $c_{a \to b}(p) \in [0, 1]$ denote its certainty. We measure the consistency of a match by its round trip error
\begin{equation}
    r_{a \to b}(p) = \big\| W_{b \to a}\big( W_{a \to b}(p) \big) - p \big\|_2.
\end{equation}
We retain a match only if its certainty exceeds $0.5$ and its round-trip error is below $\tau_r$, a threshold of two pixels scaled from RoMa~v2's output grid to image coordinates.

To link the cameras at each instant, we construct single-instant tracks. In every image, we randomly sample up to $1024$ pixels that have an accepted match. Each sampled pixel forms one track together with its accepted matches in the other cameras at the same instant. Since all observations of such tracks are captured simultaneously, they constrain the cameras through static and moving scene points alike.

To constrain each moving camera's trajectory, we further construct multi-instant tracks. We sample $1024$ pixels per frame and follow them through the video by chaining RoMa warps, retaining only observations reached through accepted matches. We keep tracks that span at least four time steps. To limit the optimization cost, we subsample each track to at most $24$ observations (see \cref{sec:track_selection_suppl}).

\subsection{Bundle Adjustment}
\label{sec:bundle_adjustment}

We jointly refine the cameras and 3D points using both single-instant and multi-instant tracks. Each camera has time-invariant intrinsics $K_i = (f_{x,i}, f_{y,i}, c_{x,i}, c_{y,i})$ and poses $E_{i,t}$ that are shared across time for static cameras and estimated per frame for moving cameras. For each track $j$, we optimize one 3D point $X_j$ to explain its observations $u_{j,i,t}$ in frames $\mathcal{O}_j$ by minimizing
\begin{equation}
    \sum_{j} \sum_{(i,t) \in \mathcal{O}_j} \rho\Big( \big\| \pi(K_i, E_{i,t}, X_j) - u_{j,i,t} \big\|_2^2 \Big) + E_\text{prior},
    \label{eq:ba}
\end{equation}
where $\pi$ is the pinhole projection and $\rho$ is a Cauchy loss~\cite{barron2019general} with a strict scale of only $0.05$ pixels, reducing the influence of observations inconsistent with their track's 3D point. To stabilize intrinsics estimation without forcing all cameras to share one calibration, we softly couple the intrinsics $K_i$ to a shared, optimized anchor $\bar{K}$ through
\begin{equation}
    E_\text{prior} = w \sum_i \log\big( 1 + \| D^{-1} (K_i - \bar{K}) \|_2^2 \big),
\end{equation}
where $D$ is a diagonal scaling matrix that normalizes the parameter differences and $w$ controls the prior strength. This robust penalty encourages similar intrinsics while allowing cameras to differ when supported by the observations. 
We run two passes of bundle adjustment, each for up to $50$ iterations. After the first pass, we remove observations with reprojection errors above $0.5$ pixels and tracks with fewer than two remaining observations. The second pass refines the surviving tracks, starting from the cameras and points of the first pass (see \cref{sec:ba_details} for details and additional solver settings).

\subsection{Anchor Triangulation}
\label{sec:anchor_triangulation}

\begin{figure}[t]
    \centering
    \includegraphics[width=0.9\linewidth]{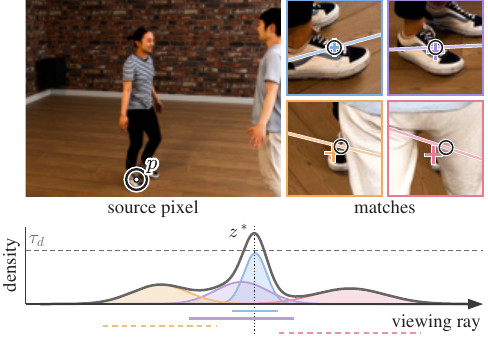}
    \caption{\textbf{Depth vote illustration.} The source pixel $p$ is matched to other frames (crosses). Circles mark projections of the same 3D point and lines show epipolar lines. Matches that agree (solid) or disagree (dashed) with the peak above $\tau_d$ are shown as colored curves of their weighted contributions, with their sum in gray.}
    \label{fig:depth_vote}
    \vspace{-0.3cm}
\end{figure}

With the cameras fixed by bundle adjustment, we triangulate anchor points for as many pixels as possible. We follow the dense warps to find matches that agree on depth.

Let $X(z) = c_{i,t} + z \, d(p)$ be the viewing ray of a pixel $p$ in frame $(i,t)$, on which we seek its depth $z$. We treat every match $m$ of $p$, with certainty $c_m$ and round-trip error $r_m$, as a noisy observation of this depth. A match constrains the depth less the higher its round-trip error, the further it lies from the epipolar line of $p$, and the lower its parallax.

To model this, we assign each match a noise level that grows with its round-trip error, $\sigma_m^2 = \sigma_0^2 + \tfrac{1}{2} r_m^2$. We model the position of each match as the projection of the true point perturbed by isotropic Gaussian noise, with standard deviation $\sigma_m$. Expressed along the viewing ray, this yields a density $p_m(z)$ over log depth for every match. Its peak is narrower and higher the stronger the parallax of the match, and its mass decays with the distance of the match from the epipolar line (\cref{fig:depth_vote}, see \cref{sec:anchor_triangulation_suppl} for the exact form).
\begin{figure*}[htpb]
    \centering
    \includegraphics[width=\linewidth]{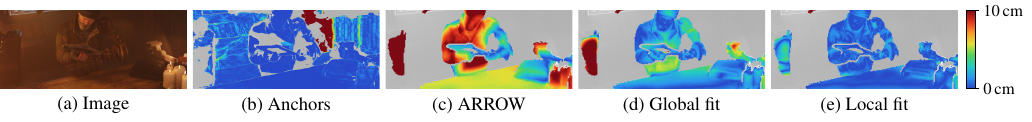}
    \vspace{-0.55cm}
    \caption{\textbf{Point cloud completion on CHARGE.} (a) Input, (b) triangulated anchors, (c) ARROW~\cite{fradlin2026arrow} predictions, (d) global alignment, and (e) local depth correction. Colors show nearest-ground-truth distances at the same instant (linear scale, capped at $10$\,cm).}
    \label{fig:completion}
\end{figure*}

For a set of matches $\mathcal{M}$, we average these densities weighted by the match certainty and approximate the mode,
\begin{equation}
    p_\mathcal{M}(z) = \frac{\sum_{m \in \mathcal{M}} c_m \, p_m(z)}{\sum_{m \in \mathcal{M}} c_m},
    \quad
    z^* \approx \arg\max_{z} \, p_\mathcal{M}(z).
    \label{eq:depth_density}
\end{equation}
We accept $z^*$ if $p_\mathcal{M}(z^*)$ reaches a threshold $\tau_d$ and retain this score as its confidence. Mode selection reduces the influence of incorrect matches as long as they do not agree with each other. We then refine the point by a weighted triangulation, in which each match is weighted by its certainty and its agreement with $z^*$ (see \cref{sec:anchor_triangulation_suppl}). We find same-instant matches more reliable and therefore first estimate the depth from them alone, repeating the procedure with the matches across time if this estimate is rejected.

\subsection{Point Cloud Completion}
\label{sec:completion}

The triangulated anchors are accurate, but they do not cover every pixel. We fill the gaps with ARROW~\cite{fradlin2026arrow} point predictions aligned to our reconstruction (\cref{fig:completion}). Throughout completion, we use our fixed, bundle-adjusted cameras.

We first pair each anchor with the ARROW point predicted at the same pixel, discarding pairs with low ARROW confidence. A single similarity transform fitted to all pairs then places the ARROW predictions in our coordinate frame. It cannot, however, correct errors in their relative depths, so we refine the alignment locally.

Let $\hat{z}(p)$ denote the aligned ARROW depth at pixel $p$, measured in our camera. We scale it by the depth ratios of nearby pairs,
\begin{equation}
    z(p) = \hat{z}(p) \, \exp\left(
    \frac{\sum_q w_q(p) \, \log \big(z^*(q)/\hat{z}(q)\big)}
         {1 + \sum_q w_q(p)} \right),
    \label{eq:completion_depth}
\end{equation}
where $z^*(q)$ is the anchor depth at pixel $q$ and the sums run over nearby anchors of the same image. The Gaussian weights $w_q(p)$ decrease with the 3D distance between the backprojections of $\hat{z}(p)$ and $\hat{z}(q)$, limiting the influence of nearby pixels on distant surfaces. We place each corrected depth on its pixel's viewing ray and fill only confident ARROW predictions away from depth discontinuities, keeping all anchors unchanged (\cref{sec:completion_suppl}).

\section{Experiments}
\label{sec:experiments}

\paragraph{Datasets.}
We establish a benchmark for camera estimation and dense reconstruction from seven existing multi-view datasets (\cref{tab:datasets}). All seven provide synchronized videos of dynamic scenes together with ground-truth camera parameters and depth maps. We use the first 150 frames of each sequence, or the full sequence if shorter. Combining these sequences with the listed camera subsets yields 373 evaluation clips with two to ten cameras (details in \cref{sec:supp_datasets}). ARROW, which Jojo4D also uses, was trained on Syn4D, so comparisons on this dataset are not training-data matched.

\begin{table}[t]
\caption{\textbf{Benchmark datasets.} Our benchmark spans diverse dynamic scenes and camera configurations. We list static (S) and moving (M) cameras per scene and the evaluated subset sizes.}
\label{tab:datasets}
\centering
\footnotesize\setlength{\tabcolsep}{2.5pt}
\resizebox{\columnwidth}{!}{%
\begin{tabular}{@{}lllcccc@{}}
\toprule
\textbf{Dataset} & \textbf{Data} & \textbf{Content} & \textbf{S} & \textbf{M} & \textbf{Subset sizes} & \textbf{Clips} \\
\midrule
Hi4D~\cite{yin2023hi4d} & rendered & two people & 4 & 4 & 2, 4, 8 & 144 \\
Perpetua~\cite{koppula2026tapvidmv} & synthetic & people indoors & 0 & 4 & 2, 4 & 42 \\
Syn4D~\cite{jiang2026syn4d} & synthetic & characters & 4 & 4 & 2, 4, 8 & 125 \\
CHARGE~\cite{nazarczuk2026charge} & synthetic & characters & 4 & 4 & 2, 4, 8 & 24 \\
MultiCamRobolab~\cite{sun2026multiview} & real & robots, people & 0 & 2 & 2 & 18 \\
MV-Kubric~\cite{rajic2025mvtracker} & synthetic & falling objects & 10 & 0 & 10 & 10 \\
DexYCB~\cite{chao2021dexycb} & real & hand, objects & 8 & 0 & 8 & 10 \\
\bottomrule
\end{tabular}}
\vspace{-0.3cm}
\end{table}

\begin{table*}[t]
\caption{\textbf{Camera estimation.} Jojo4D achieves the lowest average pose and intrinsics errors. Each setting reports the median over its scenes. Avg.\ first averages settings within each dataset, then averages over all seven datasets. Dark and light green mark the best and second best results. $R$, $t$ and FoV are in degrees, ATE$_n$ and $f$ in percent, and lower is better (additional results in \cref{sec:supp_cameras}).}
\label{tab:cameras}
\mksetup
\resizebox{\linewidth}{!}{%
\footnotesize\setlength{\tabcolsep}{2.5pt}
\begin{tabular}{l@{\hskip5pt}cccccc@{\hskip2.5pt{\color{gray!40}\vrule width 0.4pt}\hskip2.5pt}cccc@{\hskip2.5pt{\color{gray!40}\vrule width 0.4pt}\hskip2.5pt}cccccc@{\hskip2.5pt{\color{gray!40}\vrule width 0.4pt}\hskip2.5pt}cccccc@{\hskip2.5pt{\color{gray!40}\vrule width 0.4pt}\hskip2.5pt}cc@{\hskip2.5pt{\color{gray!40}\vrule width 0.4pt}\hskip2.5pt}w{c}{2.1em}w{c}{2.1em}@{\hskip2.5pt{\color{gray!40}\vrule width 0.4pt}\hskip2.5pt}cc@{\hskip2.5pt{\color{gray!40}\vrule width 0.4pt}\hskip2.5pt}ccccc}
\toprule
 & \multicolumn{6}{@{}c@{\hskip2.5pt{\color{gray!40}\vrule width 0.4pt}\hskip2.5pt}}{\cellcolor{mkhi4d}\textbf{Hi4D}} & \multicolumn{4}{@{}c@{\hskip2.5pt{\color{gray!40}\vrule width 0.4pt}\hskip2.5pt}}{\cellcolor{mkperp}\textbf{Perpetua}} & \multicolumn{6}{@{}c@{\hskip2.5pt{\color{gray!40}\vrule width 0.4pt}\hskip2.5pt}}{\cellcolor{mksyn}\textbf{Syn4D}} & \multicolumn{6}{@{}c@{\hskip2.5pt{\color{gray!40}\vrule width 0.4pt}\hskip2.5pt}}{\cellcolor{mkcharge}\textbf{CHARGE}} & \multicolumn{2}{@{}c@{\hskip2.5pt{\color{gray!40}\vrule width 0.4pt}\hskip2.5pt}}{\cellcolor{mkmcr}\textbf{MCRL}} & \multicolumn{2}{@{}c@{\hskip2.5pt{\color{gray!40}\vrule width 0.4pt}\hskip2.5pt}}{\cellcolor{mkkub}\textbf{MV-Kubric}} & \multicolumn{2}{@{}c@{\hskip2.5pt{\color{gray!40}\vrule width 0.4pt}\hskip2.5pt}}{\cellcolor{mkdex}\textbf{DexYCB}} & \multicolumn{5}{@{}c}{\cellcolor{mkavg}\textbf{Avg.}} \\
 & \multicolumn{2}{@{}c}{\cellcolor{mkhi4dl}2 cam} & \multicolumn{2}{c}{\cellcolor{mkhi4dl}4 cam} & \multicolumn{2}{c@{\hskip2.5pt{\color{gray!40}\vrule width 0.4pt}\hskip2.5pt}}{\cellcolor{mkhi4dl}8 cam} & \multicolumn{2}{@{}c}{\cellcolor{mkperpl}2 cam} & \multicolumn{2}{c@{\hskip2.5pt{\color{gray!40}\vrule width 0.4pt}\hskip2.5pt}}{\cellcolor{mkperpl}4 cam} & \multicolumn{2}{@{}c}{\cellcolor{mksynl}2 cam} & \multicolumn{2}{c}{\cellcolor{mksynl}4 cam} & \multicolumn{2}{c@{\hskip2.5pt{\color{gray!40}\vrule width 0.4pt}\hskip2.5pt}}{\cellcolor{mksynl}8 cam} & \multicolumn{2}{@{}c}{\cellcolor{mkchargel}2 cam} & \multicolumn{2}{c}{\cellcolor{mkchargel}4 cam} & \multicolumn{2}{c@{\hskip2.5pt{\color{gray!40}\vrule width 0.4pt}\hskip2.5pt}}{\cellcolor{mkchargel}8 cam} & \multicolumn{2}{@{}c@{\hskip2.5pt{\color{gray!40}\vrule width 0.4pt}\hskip2.5pt}}{\cellcolor{mkmcrl}2 cam} & \multicolumn{2}{@{}c@{\hskip2.5pt{\color{gray!40}\vrule width 0.4pt}\hskip2.5pt}}{\cellcolor{mkkubl}10 cam} & \multicolumn{2}{@{}c@{\hskip2.5pt{\color{gray!40}\vrule width 0.4pt}\hskip2.5pt}}{\cellcolor{mkdexl}8 cam} & \multicolumn{5}{@{}c}{\cellcolor{mkavgl}} \\
Method & \cellcolor{mkhi4dl}$R\!\downarrow$ & \cellcolor{mkhi4dl}$t\!\downarrow$ & \cellcolor{mkhi4dl}$R\!\downarrow$ & \cellcolor{mkhi4dl}$t\!\downarrow$ & \cellcolor{mkhi4dl}$R\!\downarrow$ & \cellcolor{mkhi4dl}$t\!\downarrow$ & \cellcolor{mkperpl}$R\!\downarrow$ & \cellcolor{mkperpl}$t\!\downarrow$ & \cellcolor{mkperpl}$R\!\downarrow$ & \cellcolor{mkperpl}$t\!\downarrow$ & \cellcolor{mksynl}$R\!\downarrow$ & \cellcolor{mksynl}$t\!\downarrow$ & \cellcolor{mksynl}$R\!\downarrow$ & \cellcolor{mksynl}$t\!\downarrow$ & \cellcolor{mksynl}$R\!\downarrow$ & \cellcolor{mksynl}$t\!\downarrow$ & \cellcolor{mkchargel}$R\!\downarrow$ & \cellcolor{mkchargel}$t\!\downarrow$ & \cellcolor{mkchargel}$R\!\downarrow$ & \cellcolor{mkchargel}$t\!\downarrow$ & \cellcolor{mkchargel}$R\!\downarrow$ & \cellcolor{mkchargel}$t\!\downarrow$ & \cellcolor{mkmcrl}$R\!\downarrow$ & \cellcolor{mkmcrl}$t\!\downarrow$ & \cellcolor{mkkubl}$R\!\downarrow$ & \cellcolor{mkkubl}$t\!\downarrow$ & \cellcolor{mkdexl}$R\!\downarrow$ & \cellcolor{mkdexl}$t\!\downarrow$ & \cellcolor{mkavgl}$R\!\downarrow$ & \cellcolor{mkavgl}$t\!\downarrow$ & \cellcolor{mkavgl}ATE$_n\!\downarrow$ & \cellcolor{mkavgl}$f\!\downarrow$ & \cellcolor{mkavgl}FoV$\downarrow$ \\
\midrule
\multicolumn{34}{l}{\textit{Feed-forward}} \\
VGGT~\cite{wang2025vggt} & 1.3 & 2.0 & 2.4 & 3.3 & 2.3 & 2.8 & 2.2 & 4.7 & 15 & 14 & .97 & 2.7 & 1.2 & 2.7 & 1.8 & 3.2 & .55 & 7.1 & 2.2 & 6.6 & 2.0 & 7.0 & .86 & 4.2 & 9.7 & 7.2 & 2.7 & 3.3 & 3.8 & 5.2 & 2.6 & 4.0 & 1.7 \\
VGGT-$\Omega$~\cite{wang2026vggtomega} & \mkS{.40} & 1.0 & \mkS{.64} & 1.1 & .69 & 1.1 & .38 & .87 & .93 & 1.5 & .53 & 1.2 & .56 & 1.1 & .81 & 1.4 & \mkS{.22} & 3.0 & .98 & 4.1 & 1.1 & 2.6 & \mkB{.84} & \mkB{2.2} & .81 & .75 & \mkS{1.5} & \mkB{1.0} & .83 & 1.5 & .49 & 1.3 & .57 \\
Pi3X~\cite{wang2026pi3} & .49 & 1.2 & .83 & 1.2 & 1.0 & 1.3 & .82 & 2.0 & 1.7 & 2.8 & .50 & 1.8 & .62 & 1.6 & .85 & 1.7 & .49 & 5.6 & .93 & 5.6 & 1.1 & 5.6 & .90 & 2.9 & 7.8 & 8.8 & \mkS{1.5} & 1.5 & 2.0 & 3.4 & 1.5 & 5.9 & 2.1 \\
$\pi^4$~\cite{wang2026pi4} & .53 & \mkS{.88} & .69 & \mkS{.92} & \mkS{.68} & \mkS{1.0} & .50 & .99 & \mkS{.62} & 1.1 & .29 & .80 & \mkS{.32} & .72 & \mkS{.31} & .51 & .42 & 2.8 & .84 & 2.0 & .90 & 2.0 & .88 & 2.4 & \mkS{.31} & \mkS{.26} & \mkB{1.2} & \mkS{1.1} & \mkS{.66} & \mkS{1.2} & \mkS{.37} & 1.1 & .44 \\
OmniX~\cite{jiang2026omnix} & 1.4 & 1.3 & 1.8 & 1.6 & 2.4 & 1.9 & 1.0 & 1.4 & 2.6 & 3.3 & .71 & 1.1 & .62 & 1.2 & 1.7 & 1.6 & .30 & 3.3 & 1.5 & 6.0 & .97 & 3.4 & .98 & \mkS{2.3} & 1.6 & 1.4 & 2.1 & 1.6 & 1.5 & 2.1 & .71 & 8.4 & 3.3 \\
ARROW-FF~\cite{fradlin2026arrow} & .70 & 1.1 & .94 & 1.0 & 1.1 & 1.3 & .64 & 1.3 & 1.5 & 1.9 & .41 & .92 & .48 & .80 & .66 & .78 & .36 & \mkS{2.3} & .73 & \mkS{1.7} & .68 & \mkS{1.7} & \mkS{.85} & 2.5 & 1.5 & 1.3 & 1.9 & 1.4 & 1.0 & 1.5 & .51 & \mkS{.96} & \mkS{.41} \\
\midrule
\multicolumn{34}{l}{\textit{Monocular, per video}} \\
Any4D~\cite{karhade2026any4d} & 2.3 & 4.4 & 2.3 & 3.0 & 2.0 & 2.7 & 3.0 & 7.2 & 3.1 & 5.7 & .98 & 4.9 & 1.0 & 4.0 & .89 & 2.3 & 1.9 & 26 & 1.6 & 9.6 & 1.4 & 10 & 1.4 & 7.9 & 1.1 & .92 & \mkS{1.5} & \mkS{1.1} & 1.7 & 5.5 & 2.2 & 11 & 4.6 \\
V-DPM~\cite{sucar2026vdpm} & 1.1 & 2.8 & 1.1 & 2.2 & 1.1 & 1.9 & 2.0 & 3.7 & 2.3 & 4.3 & 1.2 & 6.5 & .97 & 5.2 & .91 & 3.0 & 2.3 & 34 & 1.8 & 12 & 1.7 & 11 & 1.1 & 8.0 & 1.1 & .90 & \mkS{1.5} & \mkB{1.0} & 1.4 & 5.8 & 2.1 & 7.8 & 3.2 \\
MegaSaM~\cite{li2025megasam} & .64 & 1.1 & .68 & 1.1 & .76 & 1.2 & \mkS{.35} & .86 & .96 & 1.8 & \mkS{.28} & \mkS{.33} & .35 & \mkS{.57} & .37 & \mkS{.42} & .84 & 2.6 & 1.2 & 2.0 & .92 & 1.8 & .87 & 2.5 & 1.1 & .92 & \mkS{1.5} & \mkB{1.0} & .87 & 1.4 & .59 & 9.0 & 4.5 \\
\midrule
\multicolumn{34}{l}{\textit{Multi-view optimization}} \\
Sun et al.~\cite{sun2026multiview} & 1.2 & 2.2 & 1.2 & 1.6 & 1.3 & 1.7 & 2.6 & 4.3 & 7.3 & 11 & 1.1 & 1.9 & 1.0 & 1.5 & 1.3 & 1.3 & .86 & 6.1 & 2.0 & 3.7 & 1.2 & 3.4 & .91 & 2.8 & 1.4 & 1.7 & \mkS{1.5} & 1.2 & 1.8 & 3.0 & 1.7 & 5.8 & 2.1 \\
Jojo4D (init only) & .70 & 1.0 & .73 & .99 & .79 & 1.1 & .56 & \mkS{.79} & .72 & \mkS{.97} & .38 & .54 & .40 & .69 & .41 & .47 & .42 & 3.1 & \mkS{.70} & 2.2 & \mkS{.59} & 1.8 & .87 & 2.4 & 1.1 & .92 & \mkS{1.5} & \mkB{1.0} & .83 & 1.3 & .47 & 3.0 & 1.2 \\
\textbf{Jojo4D (ours)} & \mkB{.02} & \mkB{.02} & \mkB{.01} & \mkB{.01} & \mkB{.01} & \mkB{.01} & \mkB{.09} & \mkB{.21} & \mkB{.10} & \mkB{.17} & \mkB{.04} & \mkB{.06} & \mkB{.03} & \mkB{.03} & \mkB{.02} & \mkB{.02} & \mkB{.02} & \mkB{.13} & \mkB{.02} & \mkB{.06} & \mkB{.01} & \mkB{.04} & \mkB{.84} & 2.4 & \mkB{.07} & \mkB{.06} & \mkS{1.5} & \mkB{1.0} & \mkB{.36} & \mkB{.54} & \mkB{.13} & \mkB{.63} & \mkB{.29} \\
\bottomrule
\end{tabular}}
\vspace{-0.3cm}
\end{table*}

\paragraph{Baselines.} We compare Jojo4D with the feed-forward models VGGT~\cite{wang2025vggt}, VGGT-$\Omega$~\cite{wang2026vggtomega}, Pi3X~\cite{wang2026pi3}, $\pi^4$~\cite{wang2026pi4}, OmniX~\cite{jiang2026omnix} and ARROW~\cite{fradlin2026arrow}, which jointly process images from multiple cameras. We also evaluate the monocular methods Any4D~\cite{karhade2026any4d}, V-DPM~\cite{sucar2026vdpm} and MegaSaM~\cite{li2025megasam}, and the multi-view optimization-based method of Sun~\etal~\cite{sun2026multiview}, whose intrinsics are fixed to those of its Pi3X initialization. The monocular methods reconstruct each video independently, so we align their reconstructions using the same procedure as our camera initialization (\cref{sec:camera_init}).

\subsection{Camera Estimation}
\label{sec:exp_cameras}

We evaluate camera poses with relative rotation and translation-direction errors~\cite{wang2025vggt} (RPE-R and RPE-t), which require no alignment to the ground truth. For each scene, we average these errors over all valid image pairs across cameras and time. We also report the normalized absolute trajectory error (ATE$_n$), which measures the root mean square camera-center error after a single joint similarity alignment across cameras and time, normalized by the extent of the ground-truth cameras. For intrinsics, we report relative focal length and absolute field-of-view errors (see \cref{sec:supp_cameras}).

\Cref{tab:cameras} compares camera estimation across all seven datasets. Compared with the best baseline for each metric, Jojo4D reduces the dataset-averaged rotation error from $0.66^\circ$ to $0.36^\circ$ and the translation error from $1.2^\circ$ to $0.54^\circ$. Our initialization alone performs similarly to the feed-forward baselines, so this gain stems from the bundle adjustment. The relative focal length error drops from $0.96\%$ to $0.63\%$ and the field-of-view error from $0.41^\circ$ to $0.29^\circ$.

On MultiCamRobolab, our rotation error matches the best baseline at the reported precision. On DexYCB, our translation error matches the best baseline, but $\pi^4$ has lower rotation error ($1.2^\circ$ vs.\ $1.5^\circ$). On MultiCamRobolab, the methods agree more closely with each other than with the ground truth, which suggests that the ground-truth accuracy limits the comparison (see \cref{sec:supp_gt_accuracy}).

\begin{figure*}[t]
    \centering
    \includegraphics[width=\linewidth]{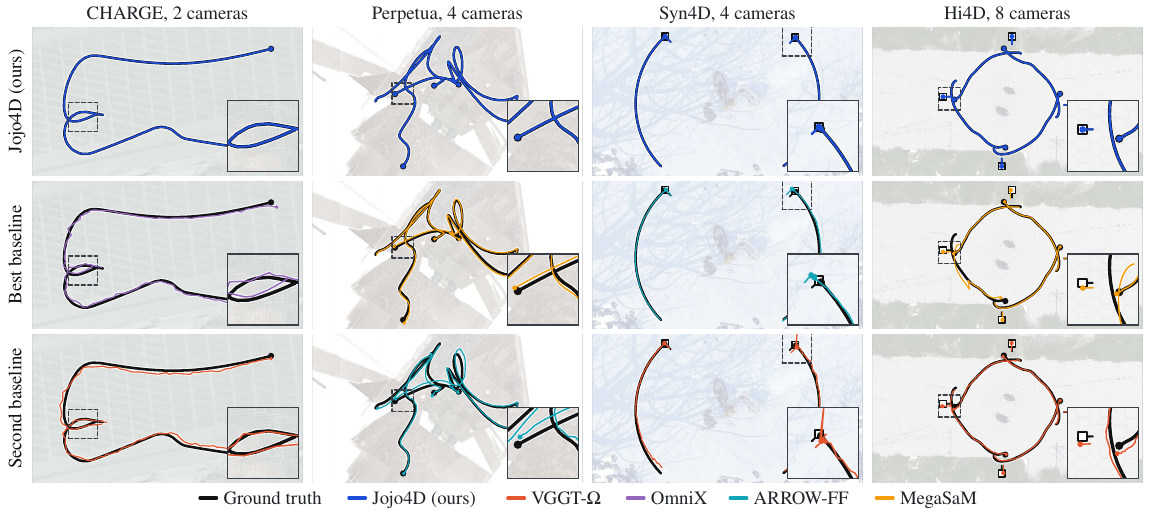}
    \vspace{-0.55cm}
    \caption{\textbf{Qualitative comparison of camera trajectories.} Camera trajectories estimated by our method (top row) and by the best and second-best baseline by RPE-R on each scene (middle and bottom rows), aligned to the ground truth (black).}
    \label{fig:trajectories}
\end{figure*}

\Cref{fig:trajectories} shows that our trajectories are better aligned with the ground truth and contain fewer outliers. The baselines often estimate the correct coarse shape of a trajectory, whereas ours are more accurate on a fine scale and typically better aligned with respect to one another. Additional camera-estimation metrics are reported in \cref{sec:supp_cameras}.

\subsection{Point Cloud Reconstruction}
\label{sec:exp_points}

We align each reconstructed scene to the ground truth with a single similarity transform, initialized from predicted and ground-truth 3D points at corresponding pixels and refined with trimmed ICP~\cite{besl1992icp,chetverikov2002trimmed}. We then report the F1 score of accuracy and completeness at distance thresholds of $2$\,cm and $5$\,cm. For methods with per-point confidence, we report the area under the accuracy--completeness curve (AUC) (see \cref{sec:supp_points}).

\begin{table*}[t]
\caption{\textbf{Point cloud reconstruction.} Jojo4D achieves the highest average F1 and AUC at both thresholds. Each setting reports the median over its scenes, and Avg.\ averages settings within each dataset, then over all seven datasets. Dark and light green mark the best and second best results. $\text{F}_2$, $\text{F}_5$ and $\text{A}_2$, $\text{A}_5$ are the F1 score and AUC in percent at $2$\,cm and $5$\,cm (higher is better). A dash indicates that AUC is unavailable for methods without confidence scores needed to trace the accuracy--completeness curve (additional results in \cref{sec:supp_points}).}
\label{tab:point_cloud_results}
\mksetup
\resizebox{\linewidth}{!}{%
\footnotesize\setlength{\tabcolsep}{2.5pt}
\begin{tabular}{l@{\hskip5pt}cccccc@{\hskip2.5pt{\color{gray!40}\vrule width 0.4pt}\hskip2.5pt}cccc@{\hskip2.5pt{\color{gray!40}\vrule width 0.4pt}\hskip2.5pt}cccccc@{\hskip2.5pt{\color{gray!40}\vrule width 0.4pt}\hskip2.5pt}cccccc@{\hskip2.5pt{\color{gray!40}\vrule width 0.4pt}\hskip2.5pt}cc@{\hskip2.5pt{\color{gray!40}\vrule width 0.4pt}\hskip2.5pt}w{c}{2.1em}w{c}{2.1em}@{\hskip2.5pt{\color{gray!40}\vrule width 0.4pt}\hskip2.5pt}cc@{\hskip2.5pt{\color{gray!40}\vrule width 0.4pt}\hskip2.5pt}cccc}
\toprule
 & \multicolumn{6}{@{}c@{\hskip2.5pt{\color{gray!40}\vrule width 0.4pt}\hskip2.5pt}}{\cellcolor{mkhi4d}\textbf{Hi4D}} & \multicolumn{4}{@{}c@{\hskip2.5pt{\color{gray!40}\vrule width 0.4pt}\hskip2.5pt}}{\cellcolor{mkperp}\textbf{Perpetua}} & \multicolumn{6}{@{}c@{\hskip2.5pt{\color{gray!40}\vrule width 0.4pt}\hskip2.5pt}}{\cellcolor{mksyn}\textbf{Syn4D}} & \multicolumn{6}{@{}c@{\hskip2.5pt{\color{gray!40}\vrule width 0.4pt}\hskip2.5pt}}{\cellcolor{mkcharge}\textbf{CHARGE}} & \multicolumn{2}{@{}c@{\hskip2.5pt{\color{gray!40}\vrule width 0.4pt}\hskip2.5pt}}{\cellcolor{mkmcr}\textbf{MCRL}} & \multicolumn{2}{@{}c@{\hskip2.5pt{\color{gray!40}\vrule width 0.4pt}\hskip2.5pt}}{\cellcolor{mkkub}\textbf{MV-Kubric}} & \multicolumn{2}{@{}c@{\hskip2.5pt{\color{gray!40}\vrule width 0.4pt}\hskip2.5pt}}{\cellcolor{mkdex}\textbf{DexYCB}} & \multicolumn{4}{@{}c}{\cellcolor{mkavg}\textbf{Avg.}} \\
 & \multicolumn{2}{@{}c}{\cellcolor{mkhi4dl}2 cam} & \multicolumn{2}{c}{\cellcolor{mkhi4dl}4 cam} & \multicolumn{2}{c@{\hskip2.5pt{\color{gray!40}\vrule width 0.4pt}\hskip2.5pt}}{\cellcolor{mkhi4dl}8 cam} & \multicolumn{2}{@{}c}{\cellcolor{mkperpl}2 cam} & \multicolumn{2}{c@{\hskip2.5pt{\color{gray!40}\vrule width 0.4pt}\hskip2.5pt}}{\cellcolor{mkperpl}4 cam} & \multicolumn{2}{@{}c}{\cellcolor{mksynl}2 cam} & \multicolumn{2}{c}{\cellcolor{mksynl}4 cam} & \multicolumn{2}{c@{\hskip2.5pt{\color{gray!40}\vrule width 0.4pt}\hskip2.5pt}}{\cellcolor{mksynl}8 cam} & \multicolumn{2}{@{}c}{\cellcolor{mkchargel}2 cam} & \multicolumn{2}{c}{\cellcolor{mkchargel}4 cam} & \multicolumn{2}{c@{\hskip2.5pt{\color{gray!40}\vrule width 0.4pt}\hskip2.5pt}}{\cellcolor{mkchargel}8 cam} & \multicolumn{2}{@{}c@{\hskip2.5pt{\color{gray!40}\vrule width 0.4pt}\hskip2.5pt}}{\cellcolor{mkmcrl}2 cam} & \multicolumn{2}{@{}c@{\hskip2.5pt{\color{gray!40}\vrule width 0.4pt}\hskip2.5pt}}{\cellcolor{mkkubl}10 cam} & \multicolumn{2}{@{}c@{\hskip2.5pt{\color{gray!40}\vrule width 0.4pt}\hskip2.5pt}}{\cellcolor{mkdexl}8 cam} & \multicolumn{4}{@{}c}{\cellcolor{mkavgl}} \\
Method & \cellcolor{mkhi4dl}$\text{F}_2\!\uparrow$ & \cellcolor{mkhi4dl}$\text{F}_5\!\uparrow$ & \cellcolor{mkhi4dl}$\text{F}_2\!\uparrow$ & \cellcolor{mkhi4dl}$\text{F}_5\!\uparrow$ & \cellcolor{mkhi4dl}$\text{F}_2\!\uparrow$ & \cellcolor{mkhi4dl}$\text{F}_5\!\uparrow$ & \cellcolor{mkperpl}$\text{F}_2\!\uparrow$ & \cellcolor{mkperpl}$\text{F}_5\!\uparrow$ & \cellcolor{mkperpl}$\text{F}_2\!\uparrow$ & \cellcolor{mkperpl}$\text{F}_5\!\uparrow$ & \cellcolor{mksynl}$\text{F}_2\!\uparrow$ & \cellcolor{mksynl}$\text{F}_5\!\uparrow$ & \cellcolor{mksynl}$\text{F}_2\!\uparrow$ & \cellcolor{mksynl}$\text{F}_5\!\uparrow$ & \cellcolor{mksynl}$\text{F}_2\!\uparrow$ & \cellcolor{mksynl}$\text{F}_5\!\uparrow$ & \cellcolor{mkchargel}$\text{F}_2\!\uparrow$ & \cellcolor{mkchargel}$\text{F}_5\!\uparrow$ & \cellcolor{mkchargel}$\text{F}_2\!\uparrow$ & \cellcolor{mkchargel}$\text{F}_5\!\uparrow$ & \cellcolor{mkchargel}$\text{F}_2\!\uparrow$ & \cellcolor{mkchargel}$\text{F}_5\!\uparrow$ & \cellcolor{mkmcrl}$\text{F}_2\!\uparrow$ & \cellcolor{mkmcrl}$\text{F}_5\!\uparrow$ & \cellcolor{mkkubl}$\text{F}_2\!\uparrow$ & \cellcolor{mkkubl}$\text{F}_5\!\uparrow$ & \cellcolor{mkdexl}$\text{F}_2\!\uparrow$ & \cellcolor{mkdexl}$\text{F}_5\!\uparrow$ & \cellcolor{mkavgl}$\text{F}_2\!\uparrow$ & \cellcolor{mkavgl}$\text{F}_5\!\uparrow$ & \cellcolor{mkavgl}$\text{A}_2\!\uparrow$ & \cellcolor{mkavgl}$\text{A}_5\!\uparrow$ \\
\midrule
\multicolumn{33}{l}{\textit{Feed-forward}} \\
VGGT~\cite{wang2025vggt} & 37.9 & 66.3 & 31.5 & 58.9 & 41.7 & 66.6 & 41.3 & 68.5 & 7.5 & 22.9 & 24.9 & 51.1 & 25.3 & 55.4 & 24.2 & 52.0 & 33.6 & 60.8 & 25.0 & 52.1 & 27.2 & 56.0 & 57.6 & 82.9 & 18.2 & 43.3 & 77.2 & 90.9 & 38.3 & 62.3 & 27.3 & 54.5 \\
VGGT-$\Omega$~\cite{wang2026vggtomega} & 66.6 & 86.8 & 63.9 & 84.7 & 67.4 & 84.4 & \mkS{68.6} & \mkS{90.0} & 53.6 & 81.3 & 40.4 & 68.1 & 48.4 & 70.3 & 46.1 & 70.6 & 41.7 & 66.1 & 35.6 & 60.0 & 41.3 & 62.9 & \mkS{65.4} & 88.2 & 33.0 & 68.0 & 90.6 & 96.0 & 57.2 & 79.4 & 49.6 & 77.1 \\
Pi3X~\cite{wang2026pi3} & 63.9 & 88.0 & 60.0 & 84.1 & 62.8 & 84.1 & 61.0 & 81.5 & 37.3 & 68.3 & 42.0 & 69.0 & 46.7 & 72.1 & 42.8 & 71.0 & 43.5 & 68.7 & 37.8 & 62.6 & 36.8 & 59.9 & 61.3 & 85.9 & 18.7 & 46.9 & 89.0 & \mkS{96.4} & 51.9 & 74.8 & 40.5 & 68.4 \\
$\pi^4$~\cite{wang2026pi4} & 66.6 & 87.4 & 65.3 & 84.1 & 71.6 & 88.2 & 68.3 & 87.7 & \mkS{64.9} & \mkS{85.2} & 55.6 & 74.8 & 59.7 & 78.0 & 61.2 & 77.5 & 44.1 & 70.4 & 43.8 & 65.8 & 51.6 & 70.5 & 64.2 & 87.9 & 64.6 & 89.5 & \mkS{92.4} & \mkB{96.9} & 65.9 & 84.7 & 59.5 & \mkS{83.7} \\
OmniX~\cite{jiang2026omnix} & 65.4 & 88.2 & 63.0 & 85.8 & 64.1 & 86.1 & 51.8 & 78.5 & 25.1 & 56.8 & 42.9 & 68.4 & 48.1 & 73.0 & 46.9 & 72.2 & \mkS{48.0} & 70.7 & 28.3 & 55.7 & 36.7 & 62.6 & \mkS{65.4} & \mkB{88.7} & 21.7 & 51.7 & 89.9 & 95.9 & 51.9 & 75.0 & 42.8 & 69.7 \\
ARROW-FF~\cite{fradlin2026arrow} & 68.8 & 88.8 & 68.0 & 88.2 & 68.1 & 86.1 & 59.9 & 87.1 & 47.3 & 78.5 & 55.9 & 77.1 & 61.3 & 79.4 & 60.8 & 79.4 & 46.4 & 67.7 & 41.7 & 64.5 & 44.8 & 67.9 & 63.8 & \mkS{88.3} & 33.0 & 68.4 & 90.9 & 95.4 & 59.0 & 81.1 & 49.5 & 78.1 \\
\midrule
\multicolumn{33}{l}{\textit{Monocular, per video}} \\
Any4D~\cite{karhade2026any4d} & 15.6 & 40.6 & 17.2 & 38.9 & 18.6 & 40.7 & 24.2 & 53.7 & 20.9 & 49.0 & 14.4 & 42.8 & 18.2 & 44.0 & 20.5 & 42.8 & 11.6 & 29.8 & 13.0 & 32.4 & 16.8 & 37.5 & 33.3 & 65.5 & 14.8 & 41.3 & 74.3 & 86.4 & 27.7 & 51.6 & 17.5 & 41.6 \\
V-DPM~\cite{sucar2026vdpm} & 41.5 & 68.7 & 32.1 & 59.8 & 37.8 & 65.3 & 27.5 & 59.1 & 25.6 & 50.5 & 17.2 & 39.7 & 15.9 & 37.3 & 17.0 & 36.8 & 18.4 & 40.1 & 12.4 & 30.3 & 17.3 & 38.7 & 49.2 & 74.3 & 16.1 & 42.5 & 69.2 & 85.0 & 33.0 & 56.5 & 21.7 & 48.2 \\
MegaSaM~\cite{li2025megasam} & 43.2 & 68.3 & 14.5 & 36.5 & 15.8 & 40.4 & 30.6 & 54.9 & 27.2 & 56.5 & 39.0 & 63.2 & 8.2 & 25.6 & 11.3 & 35.5 & 21.9 & 42.8 & 14.1 & 29.9 & 15.1 & 31.6 & 43.4 & 71.6 & 8.0 & 20.8 & 76.0 & 87.4 & 31.1 & 51.4 & -- & -- \\
\midrule
\multicolumn{33}{l}{\textit{Multi-view optimization}} \\
Sun et al.~\cite{sun2026multiview} & 22.5 & 51.7 & 37.3 & 60.7 & 35.9 & 58.2 & 9.3 & 23.6 & 5.8 & 17.4 & 10.2 & 29.0 & 24.5 & 50.0 & 25.9 & 49.9 & 31.2 & 59.0 & 24.0 & 47.9 & 38.8 & 61.0 & 63.0 & 82.7 & 12.3 & 32.9 & 89.9 & 94.8 & 36.6 & 55.2 & -- & -- \\
Jojo4D (anchors only) & \mkS{84.3} & \mkS{89.8} & \mkS{84.8} & \mkS{88.8} & \mkS{93.4} & \mkS{96.5} & 56.7 & 67.7 & 54.7 & 67.5 & \mkS{70.5} & \mkS{84.6} & \mkS{78.2} & \mkS{90.1} & \mkB{82.1} & \mkS{92.4} & \mkB{73.7} & \mkS{83.6} & \mkS{72.4} & \mkS{79.8} & \mkB{79.5} & \mkS{86.3} & 61.3 & 80.5 & \mkB{91.6} & \mkS{95.3} & 91.5 & 95.8 & \mkS{77.1} & \mkS{86.2} & \mkS{66.6} & 77.8 \\
\textbf{Jojo4D (ours)} & \mkB{88.6} & \mkB{96.6} & \mkB{87.1} & \mkB{93.5} & \mkB{94.1} & \mkB{98.1} & \mkB{72.9} & \mkB{90.2} & \mkB{70.3} & \mkB{89.3} & \mkB{70.7} & \mkB{86.9} & \mkB{78.3} & \mkB{90.8} & \mkS{82.0} & \mkB{92.8} & \mkB{73.7} & \mkB{84.5} & \mkB{72.9} & \mkB{80.4} & \mkS{79.2} & \mkB{86.9} & \mkB{67.7} & 87.9 & \mkS{91.5} & \mkB{96.7} & \mkB{93.2} & 96.3 & \mkB{80.9} & \mkB{91.5} & \mkB{74.2} & \mkB{87.9} \\
\bottomrule
\end{tabular}}
\vspace{-0.3cm}
\end{table*}

\begin{figure*}[t]
    \centering
    \includegraphics[width=\linewidth]{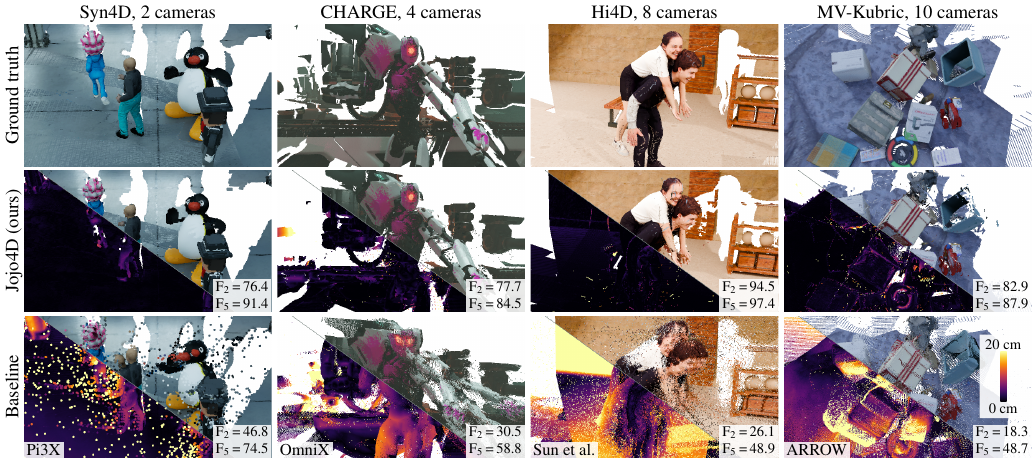}
    \caption{\textbf{Qualitative comparison of point clouds.} Each column shows one instant, with ground truth, Jojo4D and a baseline from top to bottom. Above the diagonal we show image colors, below it the nearest ground-truth error ($0$--$20$\,cm). Tags report $\text{F}_2$ and $\text{F}_5$ (\%).}
    \label{fig:point_clouds}
    \vspace{-0.3cm}
\end{figure*}

Jojo4D has the highest average F1 and AUC across the seven datasets at both distance thresholds (\cref{tab:point_cloud_results}). Compared with the best baseline, $\pi^4$, Jojo4D raises F1 from $65.9\%$ to $80.9\%$ at $2$\,cm and from $84.7\%$ to $91.5\%$ at $5$\,cm. The gain is largest at the stricter threshold, where the accuracy of the cameras and of the triangulated points matters most. Jojo4D also improves AUC over $\pi^4$ from $59.5\%$ to $74.2\%$ at $2$\,cm and from $83.7\%$ to $87.9\%$ at $5$\,cm.

\Cref{fig:point_clouds} shows that our reconstructions reach similar coverage to the feed-forward baselines while reducing geometric errors on both moving objects and the background.

\subsection{Camera Estimation Ablations}
\label{sec:ablation_cameras}

\Cref{tab:ablation_cameras} shows results for different ablations of our camera estimation algorithm. We ablate on $48$ scenes, six from each setting with two and four cameras of Hi4D, Perpetua, Syn4D and CHARGE, and average them as in \cref{tab:cameras}.

\textbf{Camera initialization.} We replace our camera initialization, which stitches per-video predictions and aligns them globally (\cref{sec:camera_init}), by a single ARROW prediction for the images of all cameras and time steps at once. On its own, this joint prediction is a less robust initialization than our stitched per-video trajectories. Over all benchmark scenes, its rotation error exceeds $2^\circ$ on $23$ scenes, compared to $11$ for our initialization. The bundle adjustment corrects most of these errors, so both initializations lead to similar final relative-pose errors, with a slightly higher RPE-R for the joint prediction ($0.15^\circ$ versus $0.12^\circ$). With Pi3X in place of ARROW in our initialization, the RPE-R increases from $0.12^\circ$ to $0.24^\circ$, while the intrinsics remain as accurate.

\textbf{Track selection.} We run the bundle adjustment with only single-instant tracks and with only multi-instant tracks (\cref{sec:track_selection}). Both variants increase the rotation and translation errors by a factor of at least $4.7$, so neither type of track alone constrains the cameras sufficiently.

\textbf{Shared intrinsics prior.} Without the prior that ties the intrinsics of cameras with similar initial intrinsics (\cref{sec:ba_details}), the intrinsics are estimated less accurately. The focal length error grows from $0.064\%$ to $0.17\%$ and the field-of-view error from $0.023^\circ$ to $0.065^\circ$. Since the poses are estimated jointly with the intrinsics, the rotations degrade as well, and RPE-R increases from $0.12^\circ$ to $0.15^\circ$, while RPE-t remains at $0.17^\circ$.

\textbf{Matching backbone.} Reducing RoMa v2's resolution to $512$\,px or replacing it with RoMa v1 increases RPE-R from $0.12^\circ$ to $0.18^\circ$ and $0.54^\circ$, respectively. This indicates that correspondence accuracy can limit camera accuracy.

\begin{table}[htbp]
\caption{\textbf{Ablation of the camera estimation.} 
Dark and light green mark the best and second best results.}
\label{tab:ablation_cameras}
\mksetup
\resizebox{\linewidth}{!}{%
\footnotesize\setlength{\tabcolsep}{2pt}
\begin{tabular}{lccccc}
\toprule
Variant & $R\!\downarrow$ & $t\!\downarrow$ & ATE$_n\!\downarrow$ & $f\!\downarrow$ & FoV$\downarrow$ \\
\midrule
ARROW~\cite{fradlin2026arrow} feed-forward initialization & .147 & .187 & .079 & \mkB{.047} & \mkS{.018} \\
Pi3X~\cite{wang2026pi3} instead of ARROW & .244 & .333 & .173 & \mkS{.048} & \mkB{.017} \\
\midrule
Single-instant tracks only & 3.13 & 6.83 & 8.00 & .534 & .190 \\
Multi-instant tracks only & .568 & 2.53 & 1.06 & 1.37 & .531 \\
\midrule
First BA pass only (with weak-frame fallback) & \mkS{.129} & .189 & .066 & .071 & .025 \\
Without shared intrinsics prior & .152 & \mkB{.172} & \mkS{.062} & .168 & .065 \\
\midrule
RoMa v2~\cite{edstedt2026romav2} at $512$\,px & .178 & .259 & .088 & .269 & .091 \\
RoMa v1~\cite{edstedt2024roma} & .540 & .860 & .489 & .109 & .039 \\
\midrule
\textbf{Jojo4D (ours)} & \mkB{.120} & \mkS{.174} & \mkB{.060} & .064 & .023 \\
\bottomrule
\end{tabular}
}
\vspace{-0.4cm}
\end{table}

\subsection{Depth Completion Ablations}
\label{sec:ablation_completion}

\Cref{tab:ablation_completion} ablates the dense reconstruction (\cref{sec:completion}) on the same scenes. Unless stated otherwise, all variants use the same cameras and the same ARROW predictions.

\begin{table}[t]
\caption{\textbf{Ablation of the dense reconstruction.} Metrics as in \cref{tab:point_cloud_results}, with the Chamfer distance (CD) in centimeters capped at $5$\,cm. Dark and light green mark the best and second best results.}
\label{tab:ablation_completion}
\mksetup
\resizebox{\linewidth}{!}{%
\footnotesize\setlength{\tabcolsep}{3pt}
\begin{tabular}{l@{\hskip5pt}cc@{\hskip6pt}cc@{\hskip6pt}cc@{\hskip6pt}cc@{\hskip6pt}c}
\toprule
 & \multicolumn{2}{c}{\cellcolor{mkpanel}F1$\uparrow$} & \multicolumn{2}{c}{\cellcolor{mkpanel}Acc.$\uparrow$} & \multicolumn{2}{c}{\cellcolor{mkpanel}Comp.$\uparrow$} & \multicolumn{2}{c}{\cellcolor{mkpanel}AUC$\uparrow$} & \cellcolor{mkpanel}CD$\downarrow$ \\
Variant & $2$\,cm & $5$\,cm & $2$\,cm & $5$\,cm & $2$\,cm & $5$\,cm & $2$\,cm & $5$\,cm &  \\
\midrule
Dense target, ARROW points only & 65.6 & 81.0 & 75.2 & 90.5 & 59.8 & 74.6 & 52.0 & 72.2 & 1.81 \\
Sparse target, ARROW points only & 63.4 & 79.6 & 71.8 & 89.1 & 58.0 & 73.6 & 49.8 & 70.8 & 1.91 \\
Sparse target, anchors + ARROW & 75.6 & \mkS{88.3} & 80.6 & 92.3 & \mkS{71.6} & \mkS{84.7} & \mkS{68.1} & \mkS{83.1} & 1.40 \\
One similarity per video & 74.1 & 87.3 & 78.8 & 91.0 & 70.3 & 84.0 & 67.0 & 82.4 & 1.47 \\
Scale and shift per image & 74.5 & 88.0 & 79.3 & 91.8 & 70.7 & 84.5 & 67.2 & 82.8 & 1.44 \\
\midrule
Pi3X~\cite{wang2026pi3} instead of ARROW & 72.6 & 84.2 & \mkS{82.2} & 92.6 & 65.1 & 77.1 & 62.6 & 75.9 & 1.50 \\
$\pi^4$~\cite{wang2026pi4} instead of ARROW & \mkS{75.8} & 85.8 & \mkB{88.5} & \mkB{96.4} & 67.0 & 77.7 & 65.0 & 77.0 & \mkB{1.37} \\
\midrule
\textbf{Jojo4D (ours)} & \mkB{76.0} & \mkB{88.7} & 81.0 & \mkS{92.8} & \mkB{71.9} & \mkB{85.1} & \mkB{68.4} & \mkB{83.4} & \mkS{1.38} \\
\bottomrule
\end{tabular}}
\vspace{-0.3cm}
\end{table}

\textbf{Triangulated points.} Replacing our triangulated points with aligned ARROW predictions at every pixel lowers F1 at $2$\,cm from $76.0\%$ to $65.6\%$. On all $48$ scenes, keeping our triangulated points gives higher F1 than using the aligned feed-forward points that would replace them. Across the full benchmark, completing our points with ARROW raises average F1 at $2$\,cm from $77.1\%$ to $80.9\%$ (\cref{tab:point_cloud_results}).

\textbf{Alignment target.} When our triangulated points are kept, aligning ARROW to the sparse points of the bundle adjustment instead of our dense anchors yields a similar F1 score at $2$\,cm ($75.6\%$ versus $76.0\%$). When ARROW replaces all our points, however, aligning it to the sparse points lowers F1 from $65.6\%$ to $63.4\%$ at $2$\,cm.

\textbf{Local fit.} Replacing our local fit by a single similarity transform per video or by a scale and shift per image lowers the F1 score at $2$\,cm from $76.0\%$ to $74.1\%$ and $74.5\%$, respectively. The gain is small on average but consistent, as the local fit improves $42$ of the $48$ scenes over a single similarity transform per video.

\textbf{Backbone.} Replacing ARROW with Pi3X for both cameras and completion yields slightly more accurate but less complete point clouds, lowering F1 at $2$\,cm from $76.0\%$ to $72.6\%$. With $\pi^4$, F1 nearly matches ARROW at $2$\,cm ($75.8\%$) but drops from $88.7\%$ to $85.8\%$ at $5$\,cm.

\subsection{Limitations}
\label{sec:limitations}

Jojo4D assumes synchronized videos and does not estimate timing offsets between cameras. Its clip-level matching and optimization run offline, not in real time (\cref{sec:supp_runtime}). Our ablations test individual Jojo4D components, but attributing the gains over Sun~\etal~\cite{sun2026multiview} to specific design choices remains difficult because the pipelines differ in matching, scene representation, and optimization. We compute local corrections per image without an explicit temporal-consistency constraint, so filled regions may flicker over time.

\section{Conclusion}
\label{sec:conclusion}

We introduce Jojo4D, a method for reconstructing dynamic scenes from synchronized, uncalibrated multi-camera video. Jojo4D uses image correspondences to refine cameras and triangulate geometry, then aligns feed-forward predictions to complete regions without reliable matches. This combines the precision of geometric refinement with the coverage of learned reconstruction. Our unified evaluation across seven existing datasets shows gains in camera estimation and dense reconstruction across static, moving, and mixed camera configurations.

\iftoggle{cvprfinal}{\section*{Acknowledgments}

This work was supported by the Swiss AI Initiative under project ID abs33 with computational resources provided by the Swiss National Supercomputing Centre on the Alps infrastructure. This work was also supported by the ETH Future Computing Laboratory. We thank Fabio Di Meo for his exploratory work on multi-view 4D reconstruction, Guido Caccianiga for sharing MV-dVRK surgical footage for our teaser, and Zador Pataki for proofreading the manuscript.
}{}
\section*{LLM Usage Statement}
We used OpenAI Codex and Anthropic Claude to assist with method development, coding and debugging, experiment orchestration and analysis, figure preparation, literature checks, and manuscript drafting and revision; the authors directed this work and take full responsibility for the submitted content, results, and scientific claims.

\appendix
\clearpage
\counterwithin{table}{section}
\counterwithin{figure}{section}
\maketitlesupplementary
\section*{Appendix Contents}
{\small
\newcommand{\suppentry}[1]{\noindent\hyperref[#1]{\ref*{#1}\hspace{1em}\nameref*{#1}\dotfill\pageref*{#1}}\par}
\suppentry{sec:implementation_details}
\suppentry{sec:supp_datasets}
\suppentry{sec:supp_dynamic_tracks}
\suppentry{sec:supp_runtime}
\suppentry{sec:supp_results}
\suppentry{sec:supp_tracking}
\suppentry{sec:supp_baselines}
\suppentry{sec:supp_single_camera}
\suppentry{sec:supp_failures}
\suppentry{sec:supp_qualitative}
}
\section*{Highlights}
After describing the implementation and benchmark datasets (\cref{sec:implementation_details,sec:supp_datasets}), we show that the filter between bundle adjustment passes preferentially removes observations of moving objects from tracks that span multiple instants, without using motion masks (\cref{sec:supp_dynamic_tracks}).

Jojo4D is slower than most feed-forward baselines. At $1024\times768$ on one GH200, it takes $5.3$\,s per time step for a two-camera clip of $30$ frames and $30.8$\,s for an eight-camera clip of $150$ frames (\cref{sec:supp_runtime}). Its time per step stays nearly constant as clips get longer, so its runtime grows roughly linearly with clip length, and with eight cameras and $150$ frames it needs less GPU memory than any baseline.

The detailed benchmark results confirm with additional metrics that Jojo4D outperforms all baselines, including the recent $\pi^4$ (\cref{sec:supp_results}). With ground-truth intrinsics as input, its camera errors decrease further (\cref{sec:supp_calibrated}). Jojo4D also runs on top of other feed-forward models and further improves their cameras, as we show for Pi3X and $\pi^4$ (\cref{sec:supp_cameras}). Our conclusions do not depend on reporting the median over scenes in the main paper, since with the mean Jojo4D still has the lowest camera errors and the highest point cloud F1 of all methods (\cref{sec:supp_cameras,sec:supp_points}).

On TAPVid-MV~\cite{koppula2026tapvidmv}, a benchmark for multi-view 3D point tracking, trackers using Jojo4D's cameras and depth achieve state-of-the-art results on four of the six available subsets: in both world and camera coordinates on Hi4D, Perpetua and the EgoExo4D pre-release, and in camera coordinates on PACE. With all six subsets weighted equally, trackers using Jojo4D also rank first overall in both world and camera coordinates (\cref{sec:supp_tracking}). On Hi4D, world-space tracking accuracy more than doubles, from $20.8\%$ to $48.7\%$.

We further compare Jojo4D with structure from motion and SLAM methods (\cref{sec:supp_baselines}). Even when structure from motion receives RoMa v2 matches on Jojo4D's pairs of synchronized images, its best rotation error on the $24$ two-camera clips is $0.66^\circ$, compared with $0.08^\circ$ for Jojo4D.

Jojo4D also works with a single camera (\cref{sec:supp_single_camera}). On the one-camera videos of our benchmark, it has the lowest average pose errors and the highest average point cloud F1, and on MPI Sintel~\cite{butler2012sintel} it has the lowest camera errors.

We trace Jojo4D's largest camera trajectory errors to limited overlap between views. When two camera pairs film different parts of a Perpetua room, Jojo4D can recover the camera motion within each pair but has too few matches between the pairs to align their reconstructions (\cref{sec:supp_failures}). 

We provide further qualitative results on all benchmark datasets and on a casual recording outside the benchmark (\cref{sec:supp_qualitative,sec:supp_real_demo}). For the latter, we filmed a guitar performance with four handheld phones and a drone. Despite camera shake, motion blur, color differences between devices, and imperfect manual synchronization, Jojo4D recovers all five camera paths and the moving 3D scene.

\smallskip
{\centering \(\scriptstyle\ast\)\hspace{.35em}\(\ast\)\hspace{.35em}\(\scriptstyle\ast\)\par}

\section{Implementation Details}
\label{sec:implementation_details}

\subsection{Stitching of per-video trajectories}
\label{sec:chunk_stitching}

As described in \cref{sec:camera_init}, we run ARROW on overlapping chunks of at most 30 frames and stitch their predictions into a single trajectory per video. For videos longer than 30 frames, chunks start every 15 frames, with an additional end-aligned chunk if needed to cover the final frame. ARROW estimates the cameras of a chunk from its predictions on a grid of $64 \times 64$ query pixels per frame, and the stitching below uses only these sparse predictions. For point cloud completion (\cref{sec:completion_suppl}), we reuse the chunk encodings to decode a dense point map for each frame at a two-pixel stride in both dimensions. When chunks overlap, we decode each shared frame from the chunk in which it lies closest to the middle.

Here, $R$ and $c$ denote the camera-to-world rotation and the center of a camera in the coordinate frame of the respective prediction, and $z$ denotes the depths at the query pixels, given by the third coordinate of each ARROW point in the frame's own camera coordinates. The first chunk defines the coordinate frame of the video, and each subsequent chunk $k$ is aligned to the frames stitched so far. Let $R^k(t)$, $c^k(t)$ and $z^k(t)$ be the prediction of chunk $k$ for frame $t$, and let $R(t)$, $c(t)$ and $z(t)$ be the stitched values. Each overlapping frame thus has two depth predictions for the same query pixels. To estimate the factor that scales one depth map $z'$ to match the other, $z$, we rank the pixels valid in both maps by the product of their ARROW point confidences. We retain the top half, $\Omega$, and take the median of their depth ratios:
\begin{equation}
    s(z, z') = \operatorname*{med}_{p \in \Omega} \frac{z(p)}{z'(p)}.
    \label{eq:depth_ratio}
\end{equation}

The chunk scale $\alpha_k$ is the median of these per-frame estimates over the overlapping frames $\mathcal{O}_k$. Together with the rotation $G_k$ and offset $\beta_k$, it maps chunk $k$ into the video coordinate frame:
\begin{align}
    \alpha_k &= \operatorname*{med}_{t \in \mathcal{O}_k} \, s\big(z(t), z^k(t)\big), \\
    G_k &= \operatorname*{mean}_{t \in \mathcal{O}_k} \, R(t) \, R^k(t)^\top, \\
    \beta_k &= \operatorname*{med}_{t \in \mathcal{O}_k} \, \big( c(t) - \alpha_k G_k \, c^k(t) \big),
\end{align}
where $G_k$ is the $\ell_1$ mean on $SO(3)$, which minimizes the sum of angular distances to the relative rotations of the overlapping frames, and $\beta_k$ is the component-wise median of the offsets over those frames. Frames that are new in chunk $k$ are then added as $R(t) = G_k R^k(t)$, $c(t) = \alpha_k G_k c^k(t) + \beta_k$ and $z(t) = \alpha_k z^k(t)$, together with their confidences from chunk $k$, while frames that are already stitched remain unchanged. We transform each chunk's dense points just like its camera centers.

We classify a camera as static if its stitched ARROW trajectory satisfies two conditions throughout the clip: its rotation stays within $0.3^\circ$ of its $\ell_1$ mean, and no per-frame camera center $c(t)$ differs from the coordinate-wise median of that camera's centers over all frames by more than $3\%$ of the video's median predicted scene depth. For these cameras, we replace the trajectory with a single pose before global alignment, using the mean rotation and median center from these tests.

\subsection{Global alignment of camera trajectories}
\label{sec:camera_alignment_suppl}
\paragraph{Bipartite alignment graph.}
After chunk stitching, each video trajectory resides in its own coordinate frame $C_i^{\mathrm{vid}}$. To bring them into one global coordinate frame $W$, we additionally run ARROW jointly on the $N$ images of every instant. This gives one prediction of the relative placement of all cameras per instant, expressed in a separate coordinate frame $C_t^{\mathrm{inst}}$, and we robustly average these predictions by message passing on the graph described below. Specifically, let $R_{i,t}^{\mathrm{vid}}$ and $c_{i,t}^{\mathrm{vid}}$ denote the rotation and center of camera $i$ at time step $t$ from the video prediction, and $R_{i,t}^{\mathrm{inst}}$ and $c_{i,t}^{\mathrm{inst}}$ the corresponding quantities from the per-instant prediction. We keep each trajectory fixed up to a global similarity and seek the similarity transforms
\begin{equation}
    T_i^{\mathrm{vid}}(x) = a_i Q_i x + b_i, \qquad T_t^{\mathrm{inst}}(x) = s_t P_t x + d_t
    \label{eq:node_similarities}
\end{equation}
that map $C_i^{\mathrm{vid}}$ and $C_t^{\mathrm{inst}}$ to $W$ while respecting the per-instant predictions.

\begin{figure}[t]
    \centering
    \includegraphics[width=\linewidth]{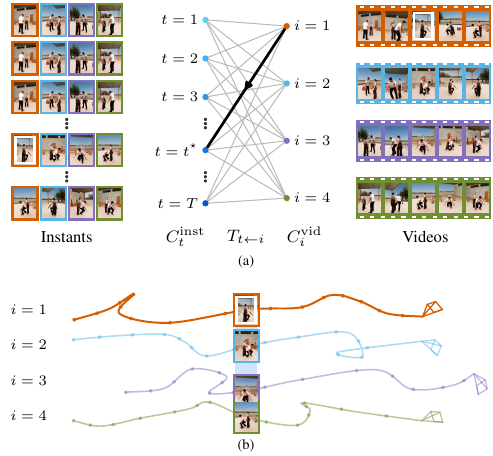}
    \caption{\textbf{Video-to-instant alignment.} We align the videos by message passing on a bipartite graph of instants and videos. (a) Each image links its instant (left) to its video (right) through the feed-forward predictions. (b) Each per-instant prediction places the trajectories relative to one another at that time. Message passing robustly averages these relative poses over the graph.}
    \label{fig:alignment_graph}
\end{figure}

Consider a bipartite graph in which one set of nodes represents the coordinate frames $C_t^{\mathrm{inst}}$ of the instants and the other set represents the coordinate frames $C_i^{\mathrm{vid}}$ of the videos (see \cref{fig:alignment_graph}). Every image $(i,t)$ appears in both the video prediction and the per-instant prediction and thus forms an edge between the nodes $i$ and $t$. Let $x$ be a 3D point expressed in the video's coordinate frame. To map it into the instant's frame, we first express it in the camera coordinates of their shared image
\begin{equation}
    x_{\mathrm{cam}} = \big(R_{i,t}^{\mathrm{vid}}\big)^\top
        \big(x-c_{i,t}^{\mathrm{vid}}\big).
\end{equation}
We then scale these camera coordinates by $\rho_{i,t}$ and place them with the instant's camera pose, which gives the point in $C_t^{\mathrm{inst}}$
\begin{equation}
    \begin{aligned}
        T_{t\leftarrow i}(x)
            &= \rho_{i,t} R_{i,t}^{\mathrm{inst}} x_{\mathrm{cam}} + c_{i,t}^{\mathrm{inst}}\\
            &= \rho_{i,t} R_{i,t}^{\mathrm{inst}} \big(R_{i,t}^{\mathrm{vid}}\big)^\top
                \big(x-c_{i,t}^{\mathrm{vid}}\big) + c_{i,t}^{\mathrm{inst}}\\
            &= \rho_{i,t} A_{i,t} \big(x-c_{i,t}^{\mathrm{vid}}\big) + c_{i,t}^{\mathrm{inst}},
    \end{aligned}
    \label{eq:edge_similarity}
\end{equation}
where $A_{i,t} = R_{i,t}^{\mathrm{inst}} \big( R_{i,t}^{\mathrm{vid}} \big)^\top$ is the relative rotation from video to instant coordinates. We use the same median depth ratio (\cref{eq:depth_ratio}) to scale the video prediction to the per-instant prediction of image $(i,t)$: $\rho_{i,t} = s\big(z_{i,t}^{\mathrm{inst}}, z_{i,t}^{\mathrm{vid}}\big)$. During alignment, we keep these edge transforms fixed.

If the predictions were consistent, mapping $x$ to world coordinates through the instant would give the same point as mapping it there directly from the video: $T_t^{\mathrm{inst}}\big(T_{t\leftarrow i}(x)\big) = T_i^{\mathrm{vid}}(x)$. Substituting the transforms from \cref{eq:node_similarities,eq:edge_similarity} into this identity gives three conditions on the unknown node rotations $(Q_i,P_t)$, scales $(a_i,s_t)$, and translations $(b_i,d_t)$, respectively:
\begin{align}
    Q_i &= P_t A_{i,t}, \label{eq:cons_rot} \\
    a_i &= s_t \, \rho_{i,t}, \label{eq:cons_scale} \\
    a_i Q_i c_{i,t}^{\mathrm{vid}} + b_i &= s_t P_t c_{i,t}^{\mathrm{inst}} + d_t. \label{eq:cons_pos}
\end{align}

\paragraph{Message passing.}
To robustly average the individual similarity transforms, we employ a message passing algorithm on this graph. A \emph{message} is a neighbor's estimate of how to map the receiving node into world coordinates. Each node updates its transform by robustly averaging these estimates. Rotation is independent of scale and position, and therefore also provides the criterion for identifying outliers. We thus first estimate all rotations, then select the inlier edges, and finally estimate scales and positions on these edges alone.

We choose a starting instant $h$ and initialize the rotations by setting $P_h = I$ and $Q_i = A_{i,h}$ for each video. We then alternate between updating all instant nodes and all video nodes for five iterations. Following \cref{eq:cons_rot}, an instant node $t$ receives rotation estimates $Q_i A_{i,t}^\top$ from its video nodes, and a video node $i$ receives estimates $P_t A_{i,t}$ from its instant nodes.

We define the agreement between two rotations $R$ and $S$ in terms of their angular distance $\theta(R,S)$:
\begin{equation}
    \kappa(R,S) = \max\!\left(0,\,1-\left(\frac{\theta(R,S)}{\tau}\right)^2\right), \qquad \tau = 3^\circ.
    \label{eq:inlier_score}
\end{equation}
Two rotations agree if this score is positive, that is, if they differ by less than $\tau$. At a given node, let $S_k$ be the rotation proposed by its $k$-th neighbor. We apply this score to pairs of proposals and sum the agreement for each one: $g_k = \sum_l \kappa(S_k,S_l)$. We take the highest-scoring rotation as the reference and keep all rotations that agree with it (i.e., differ from it by less than $3^\circ$). To update the node, we sum the retained rotation matrices with equal weight and project the sum onto $SO(3)$ by singular value decomposition.

Since the starting instant may be an outlier, we run the procedure with each instant in turn as the starting instant $h$ and keep the solution with the highest total agreement $\sum_{i,t} \kappa(Q_i,P_t A_{i,t})$. We mark each edge $(i,t)$ as an inlier if $Q_i$ and $P_t A_{i,t}$ agree. If an instant has fewer than two inlier edges, we exclude that instant from the scale and translation estimates.

On the inlier edges, we first estimate scales and then translations by alternating updates between instant and video nodes. For scale, each video predicts the scale of its neighboring instants and vice versa by \cref{eq:cons_scale}. Each node takes the median of the predictions it receives:
\begin{align}
    s_t &\leftarrow \operatorname*{med}_{i} \big( a_i / \rho_{i,t} \big), \\
    a_i &\leftarrow \operatorname*{med}_{t} \big( s_t \, \rho_{i,t} \big),
\end{align}
starting from $a_i = 1$. Multiplying all $a_i$ and $s_t$ by a common factor leaves \cref{eq:cons_scale} unchanged, so the edges determine the scales only up to the scale of $W$. After every iteration, we therefore divide all scales by the geometric mean of the $a_i$. This sets the scale of $W$ to that of a typical video and does not otherwise change the result.

With rotations and scales fixed, only the translations remain. The video translation $b_i$ and the instant translation $d_t$ must bring the two predictions of the same camera center to the same world position. We define $y_{i,t} = s_t P_t c_{i,t}^{\mathrm{inst}} - a_i Q_i c_{i,t}^{\mathrm{vid}}$, so that \cref{eq:cons_pos} becomes $b_i - d_t = y_{i,t}$. Each node then takes the per-coordinate median of the translations predicted by its neighbors
\begin{align}
    d_t &\leftarrow \operatorname*{med}_{i} \big( b_i - y_{i,t} \big), \\
    b_i &\leftarrow \operatorname*{med}_{t} \big( d_t + y_{i,t} \big),
\end{align}
starting from zero. We run 20 iterations for both the scale and the position. The instant nodes are discarded afterwards, and the initial pose of camera $i$ at time step $t$ is given by the rotation $Q_i R_{i,t}^{\mathrm{vid}}$ and the center $a_i Q_i c_{i,t}^{\mathrm{vid}} + b_i$.

\subsection{Matching graph}
\label{sec:matching_graph_suppl}

\begin{figure}[htbp]
    \centering
    \captionsetup{skip=5pt}
    \includegraphics[width=\linewidth]{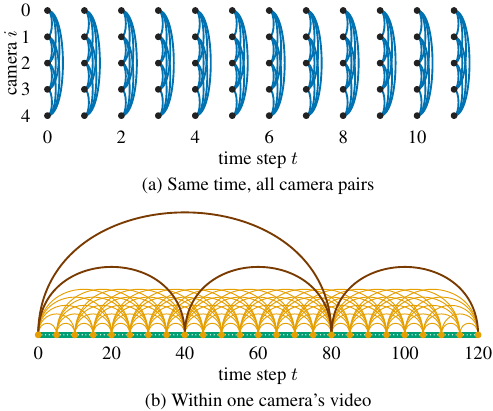}
    \caption{\textbf{Matching graph.} Dots denote frames. (a) We match all camera pairs at the same time step (blue). (b) Within a video, we match consecutive frames (teal), keyframes that are at most five keyframes apart (orange) and keyframes at power-of-two offsets (brown).}
    \label{fig:matching}
\end{figure}

\Cref{fig:matching} shows the frame pairs that we match, as described in \cref{sec:correspondence_graph}. We use the precise preset of RoMa~v2, which outputs dense warps on a $1280 \times 1280$ grid.

\subsection{Track selection}
\label{sec:track_selection_suppl}

\begin{figure}[htbp]
    \centering
    \includegraphics[width=\linewidth]{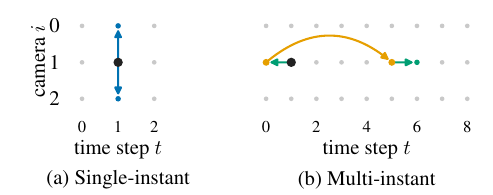}
    \caption{\textbf{Track types.} The black dot marks a source pixel, and orange dots mark keyframes. (a) A \emph{single-instant track} combines the matches of the source pixel to all other cameras at the same time step (vertical blue arrows). (b) A \emph{multi-instant track} reaches a target frame in the same video by chaining consecutive-frame warps (horizontal teal arrows) and direct keyframe warps (curved orange arrows).}
    \label{fig:tracks}
\end{figure}

\Cref{fig:tracks} illustrates the two track types introduced in \cref{sec:track_selection}. Both use the same match filter, which accepts a match if its certainty exceeds $0.5$ and its round-trip error is below $\tau_r$. For an image of size $H \times W$ and a RoMa~v2 output grid of size $h \times w$, we set the round-trip threshold in image pixels to $\tau_r = 2\max(W/w, H/h)$.

We form a multi-instant track by chaining warps. Since a chained position generally falls between pixels, we sample the next warp bilinearly at that position. We reach frames up to two time steps before or after the source by following consecutive-frame warps. To reach more distant frames, we always choose the shortest connecting path. We first follow these warps to the nearest keyframe, then use the fewest possible keyframe hops before following consecutive-frame warps to nearby frames. If a keyframe hop fails the match filter, we try other paths of the same length, but never a longer one. A consecutive-frame chain stops at its first rejected match.

Of the multi-instant tracks that reach at least four time steps, we randomly retain at most $133$ per source frame, which gives about $4000$ tracks for a video of $30$ frames. We subsample a track with more than $24$ observations by splitting its time-sorted observations into $24$ equal groups and drawing one at random from each. We build multi-instant tracks only for moving cameras, since all frames of a static camera share one pose.

\subsection{Bundle adjustment}
\label{sec:ba_details}

\paragraph{Initialization.}
We use a pinhole camera model with intrinsics $K_i = (f_{x,i}, f_{y,i}, c_{x,i}, c_{y,i})$ and do not model lens distortion. Each camera has one set of intrinsics for the whole clip. A static camera has a single pose $E_i$, and a moving camera has one pose $E_{i,t}$ per time step. The poses start from the aligned trajectories of \cref{sec:camera_alignment_suppl}. The intrinsics of camera $i$ start from the element-wise median of the ARROW intrinsics predicted for its video. We initialize the 3D point of each track by linear triangulation of its observations under these initial cameras and discard tracks for which the triangulation fails. The second pass starts from the cameras and points of the first pass.

\paragraph{Intrinsics priors.}
At the start of each pass, we initialize the free anchor $\bar{K}$ as the element-wise median of the current camera parameters. The normalization in the shared-intrinsics prior is $D = \operatorname{diag}(0.002 \bar{f}_x, 0.002 \bar{f}_y, 1, 1)$, where $\bar{f}_x$ and $\bar{f}_y$ are the anchor's focal lengths at the start of that pass and the principal point entries are in pixels. We keep $D$ fixed during the pass and set $w = 3 \cdot 10^{-5} n_\text{obs}/N$, where $n_\text{obs}$ counts the observations included in that pass. For single-camera clips, we instead use a Gaussian prior on the principal point, centered at its value at the start of each pass, with standard deviation $2.5\max(1,L/640)$ pixels where $L$ is the long side in pixels at the bundle-adjustment resolution.

\paragraph{Solver and filtering.}
We use the Cauchy loss $\rho(x) = \sigma_\rho^2 \log(1 + x/\sigma_\rho^2)$ with scale $\sigma_\rho = 0.05$ pixels and run each of the two passes for up to $50$ iterations. Between passes, we remove observations with reprojection errors above $0.5$ pixels and discard tracks with fewer than two remaining observations.

\paragraph{Weakly constrained frames.}
After the second pass, we mark a moving camera's frames as weak if fewer than $100$ observations from multi-instant tracks survived the first-pass filter. For such frames, the pose is interpolated between the nearest supported frames before and after it, using spherical linear interpolation for the camera-to-world rotation and linear interpolation for the camera center, with interpolation weights determined by frame indices. At either end of a video, we extrapolate the rotation and center with constant velocity from the two nearest supported frames.

\subsection{Anchor triangulation}
\label{sec:anchor_triangulation_suppl}

\paragraph{Match collection.}
For a pixel $p$ of frame $(i,t)$, we collect its matches in the other cameras at time step $t$, a five-frame window around $t$ in video $i$, its nearest keyframe, and all keyframes that lie within two keyframe edges of the nearest keyframe. At each of these other time steps, we additionally follow the warps from video $i$ into the other cameras. As in \cref{sec:anchor_triangulation}, every match $m \in \mathcal{M}$ carries a position $u_m$ in a frame with projection $\pi_m$, its certainty $c_m \in [0,1]$, and its round-trip error $r_m$ (\cref{sec:track_selection}). For a chained match, $c_m$ and $r_m$ are the smallest certainty and the largest round-trip error along its chain. 

\begin{figure}[t]
    \centering
    \includegraphics[width=\linewidth]{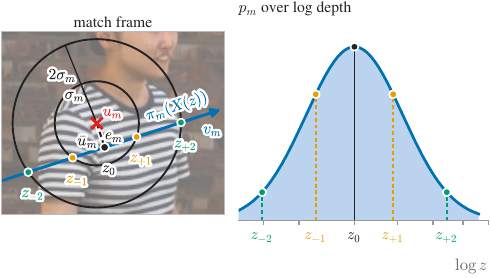}
    \caption{\textbf{Density of one match along the viewing ray.} On the left, the match $u_m$ lies at distance $e_m$ from the epipolar line $\pi_m(X(z))$. Its closest point $\bar{u}_m$ gives the candidate depth of this match $z_0$ with $s_m(z_0) = 0$. On the right, the density $p_m(z)$ of \cref{eq:match_density_derivation} drawn over log depth.}
    \label{fig:logdepth_density}
\end{figure}

\paragraph{Depth voting.}
The source pixel $p$ in frame $(i,t)$ tells us which ray the 3D point lies on, but not how far along it the point lies. \Cref{fig:logdepth_density} shows how a match in another image votes for depth along this ray. At a candidate depth $z>0$, its world position is
\begin{equation}
    X(z) = c_{i,t} + z \, d(p),
    \label{eq:ray}
\end{equation}
where $c_{i,t}$ is the source camera center and $d(p)$ points along the ray, scaled so that $z$ is depth along the source camera's optical axis. To test this depth, we project $X(z)$ into the image containing the matched pixel $u_m$ and compare the two pixel positions. The match may be in another camera or in another frame of the source camera. Applying that frame's projection matrix $P_m = K_m [R_m \,|\, t_m]$ to the ray gives a fixed term from the source camera center and a term proportional to depth,
\begin{equation}
    h_m(z) = P_m \begin{bmatrix} c_{i,t} \\ 1 \end{bmatrix} + z \, P_m \begin{bmatrix} d(p) \\ 0 \end{bmatrix} = h_m^0 + z \, h_m^d,
\end{equation}
and dividing by the third coordinate gives the predicted pixel in that image,
\begin{equation}
    \pi_m(X(z)) = \frac{1}{h_{m,3}(z)} \begin{bmatrix} h_{m,1}(z) \\ h_{m,2}(z) \end{bmatrix}.
    \label{eq:ray_projection}
\end{equation}
As $z$ varies, $\pi_m(X(z))$ moves along a straight line in the match image, the epipolar line of $p$. We assume that a correct match deviates from the projection of the true point by isotropic Gaussian noise whose variance grows with the round-trip error $r_m$,
\begin{equation}
    \sigma_m^2 = \sigma_0^2 + \tfrac{1}{2} \, r_m^2,
    \label{eq:match_noise}
\end{equation}
where $\sigma_0 = 0.4$\,px. Let $\bar{u}_m$ be the point of the epipolar line closest to the match $u_m$ and $v_m$ the unit direction of the line (see \cref{fig:logdepth_density}). Then
\begin{equation}
    \pi_m(X(z)) = \bar{u}_m + s_m(z) \, v_m,
    \label{eq:line_position}
\end{equation}
where $s_m(z)$ is its signed position along the line. Denote by $e_m = \| u_m - \bar{u}_m \|_2$ the distance of the match from the line. The likelihood of a position $s$ on the line under match $m$ is
\begin{equation}
    \begin{aligned}
    p(u_m \mid s) &= \mathcal{N}\big(u_m; \, \bar{u}_m + s \, v_m, \, \sigma_m^2 I_2\big) \\
           &= \mathcal{N}(e_m; 0, \sigma_m^2) \, \mathcal{N}(s; 0, \sigma_m^2).
    \end{aligned}
    \label{eq:match_likelihood}
\end{equation}
Here $e_m$ measures the error perpendicular to the epipolar line and $s$ the displacement along it, so the squared pixel error is $e_m^2+s^2$.
The second factor is a density over the position along the line. Changing variables from this position to the log depth multiplies it by $g_m(z) = |\mathrm{d} s_m / \mathrm{d} \log z|$, and we scale the first factor so that its maximum is one. This gives
\begin{equation}
    p_m(z) = \sqrt{2\pi}\,\sigma_m \; g_m(z) \; \mathcal{N}(e_m; 0, \sigma_m^2) \; \mathcal{N}\big(s_m(z); 0, \sigma_m^2\big),
    \label{eq:match_density_derivation}
\end{equation}
where the Jacobian factor $g_m(z) = \| \partial \, \pi_m(X(z)) / \partial \log z \|_2$ measures how far the projection moves per relative change of depth. From \cref{eq:ray_projection} it follows in closed form as
\begin{equation}
    g_m(z) = \frac{z}{h_{m,3}(z)^2} \, \Big\| h_{m,3}(z) \begin{bmatrix} h^d_{m,1} \\ h^d_{m,2} \end{bmatrix} - h^d_{m,3} \begin{bmatrix} h_{m,1}(z) \\ h_{m,2}(z) \end{bmatrix} \Big\|_2.
\end{equation}
Each factor of \cref{eq:match_density_derivation} has its own role. The first depends only on the distance $e_m$ of the match from the epipolar line and scales the whole density, so that a match far from the line contributes little. Integrated over all depths in front of both cameras, $p_m$ has mass at most $\exp(-e_m^2 / 2\sigma_m^2)$. The second places the peak at the depth whose projection is closest to the match, $s_m(z) = 0$. The Jacobian $g_m(z)$ sets the width and height of this peak. With strong parallax, the projection moves quickly along the line as the depth changes, and the density forms a narrow, high peak. With weak parallax, the same mass spreads over a wide range of depths.
For a set of matches $\mathcal{M}$, we average these densities with the match certainties as weights (\cref{eq:depth_density}). Unlike a product, which assumes that every match is an inlier, this certainty-weighted mixture is robust to outliers, since its mode lies where most matches agree. We average over log depth because the height of each density is then the relative depth precision of its match, which makes the heights comparable between near and far points and independent of the scale of the reconstruction.

We take the maximum of the averaged density as the depth $z^*$ and accept it if $p_\mathcal{M}(z^*) \geq \tau_d = 15$. The value $p_\mathcal{M}(z^*)$ becomes the confidence of the anchor. We first use only the matches at the same time step and turn to the matches across time if this estimate is rejected.

\paragraph{Point refinement.}
The vote places the point on the viewing ray of $p$. We refine it by triangulating it from all matches and the source pixel $p$, which we treat as a match with certainty $1$ and round-trip error $0$,
\begin{equation}
    X^*(p) = \arg\min_{X} \sum_{m} \omega_m \, \| \pi_m(X) - u_m \|_2^2,
    \label{eq:retriangulation}
\end{equation}
where
\begin{equation}
    \omega_m = \frac{c_m}{\sigma_m^2} \exp\Big( -\frac{\| \pi_m(X(z^*)) - u_m \|_2^2}{2 \sigma_m^2} \Big)
    \label{eq:retriangulation_weight}
\end{equation}
weights each match by its certainty, its noise level and its agreement with $z^*$. Matches that support another depth thus have almost no influence. If the refinement fails, we keep the point on the ray.

\subsection{Point cloud completion}
\label{sec:completion_suppl}

We refer to the points triangulated in \cref{sec:anchor_triangulation_suppl} as anchors. The anchors are accurate, but they do not cover every pixel. Regions without reliable matches, regions with little parallax, and moving regions that are visible in a single camera can remain without anchors. Completion aligns ARROW point predictions to the triangulated anchors while keeping our bundle-adjusted cameras fixed. We use the dense ARROW points of the camera initialization (\cref{sec:chunk_stitching}), which supply a point $\hat{X}(p)$ and confidence $\hat{c}(p)$ for each pixel of the ARROW prediction grid. The points of each video are expressed in the coordinate frame of its stitched trajectory. We first align each of these frames to our reconstruction, then correct the depths locally and fill pixels without anchors.

\paragraph{Valid cells.}
We compute the completion on a coarse grid whose cells cover two by two pixels of the ARROW prediction. Since ARROW predicts at a lower resolution than the input images, each cell covers a larger block of image pixels. We call a cell valid if at least half of these image pixels carry an anchor, that is, a triangulated point, if the $20$th and $80$th percentiles of the anchor depths differ by at most $3\%$ of their median, and if the ARROW confidence of the cell is at least $2$. A video with fewer than $50$ valid cells is not aligned and receives no filled points. We mark a cell as a depth discontinuity if its predicted depth differs from that of one of its four neighbors by more than $4\%$. Such cells are excluded from the alignment, the local correction and the filling.

\paragraph{Global alignment.}
For each video $i$, we estimate a single similarity transform for the whole video that maps the ARROW points of all its frames to the coordinate frame of our cameras. Its scale $a > 0$, rotation $Q \in SO(3)$ and translation $b \in \mathbb{R}^3$ minimize
\begin{equation}
    \sum_{t} \sum_{q \in \mathcal{B}_{i,t}} \frac{1}{z^*(q)^2} \, \big\| a \, Q \, \hat{X}(q) + b - X^*(q) \big\|_2^2.
    \label{eq:global_fit}
\end{equation}
where $\mathcal{B}_{i,t}$ is the set of valid cells of image $(i,t)$, $z^*(q)$ is the median depth of the anchors in cell $q$, and $X^*(q)$ is the point at this depth on the viewing ray of the cell center. The inverse-depth-squared weights account for errors growing with depth. We solve this weighted fit with iterative Umeyama alignment~\cite{umeyama1991least} of up to six rounds where pairs whose residual, divided by their depth, exceeds $2.5$ times the median over the remaining pairs, are discarded each round.

\paragraph{Local depth correction.}
Let $\hat{z}(p)$ denote the depth of the globally aligned ARROW point associated with pixel $p$ in our bundle-adjusted camera $(i,t)$. This point need not project to $p$, so we define $\hat{X}'(p) = c_{i,t} + \hat{z}(p) \, d(p)$ as the point at this depth on $p$'s viewing ray. At a valid cell $q$, the ratio $z^*(q) / \hat{z}(q)$ measures the required depth correction. We exclude from the local fit cells whose globally aligned depth deviates strongly from the anchor depth, $|\hat{z}(q)-z^*(q)| > 0.10 \, z^*(q)$. The local scale is a regularized weighted geometric mean of the ratios in the same image,
\begin{equation}
    \log \alpha(p) = \frac{\sum_{q \in \mathcal{B}_{i,t}} w_q(p) \, \log \big( z^*(q) / \hat{z}(q) \big)}{1 + \sum_{q \in \mathcal{B}_{i,t}} w_q(p)}.
    \label{eq:local_scale}
\end{equation}
Excluded cells have zero weight. For the others,
\begin{equation}
    w_q(p) = \exp\Big( -\frac{\| \hat{X}'(q) - \hat{X}'(p) \|_2^2}{2 \, (h \, \hat{z}(p))^2} \Big),
    \label{eq:local_weight}
\end{equation}
with $h = 0.08$. The weight of a cell thus decays with its distance to the point in 3D, so that cells that are close in the image but lie on another surface have little influence. The bandwidth $h\,\hat{z}(p)$ is proportional to the depth of the point, because both the 3D spacing between neighboring pixels and the errors of the predicted depths grow with depth. The constant $1$ in the denominator pulls $\alpha$ toward one, so that the global alignment is kept where few valid cells lie nearby.

For efficiency, the sums in \cref{eq:local_scale} run only over the $512$ valid cells nearest to the point in 3D. A pixel whose correction exceeds $|\log\alpha(p)| > 0.10$ keeps its globally aligned depth, and an image with fewer than eight valid cells keeps the global alignment as a whole.

\paragraph{Filling.}
We fill only pixels without an anchor and leave the anchors unchanged. A cell takes part in the filling if its ARROW confidence is at least $5$. Since a cell covers several image pixels, we interpolate the corrected cell depths to the pixels. A pixel is filled only if its nearest cell takes part, and its depth is the bilinear interpolation of the surrounding cells whose depth lies within $4\%$ of that of the nearest cell. A filled point receives the confidence $\tau_d \, \min(\hat{c} / 40, 1)$, where $\hat{c}$ is the ARROW confidence of its nearest cell. It never exceeds $\tau_d$, the lowest confidence an anchor can have, so that any confidence threshold removes filled points before anchors.

\subsection{Input and output resolutions}
\label{sec:supp_resolutions}

\Cref{tab:resolutions} distinguishes the image resolutions used by each method from its reconstructed and evaluated point grids. ARROW decodes points every second pixel in each dimension, whereas Jojo4D reconstructs points on the benchmark image grid using matches and ARROW predictions computed at their respective resolutions.

\begin{table*}[t]
\centering
\caption{\textbf{Image and point-grid resolutions.} Dimensions are height $\times$ width in pixels. The raw MultiCamRobolab (MCRL) images are cropped from $384\times512$ to $364\times512$ for every method to remove black borders at the top and bottom. VGGT-$\Omega$~\cite{wang2026vggtomega} center-crops CHARGE to an aspect ratio of $1:2$ before resizing, and MegaSaM~\cite{li2025megasam} crops the bottom and right edges of its resized images to multiples of eight. Jojo4D uses the listed ARROW~\cite{fradlin2026arrow} inputs for initialization and completion. RoMa~v2~\cite{edstedt2026romav2} resizes images to the listed square grids, while camera-graph and bundle-adjustment observations use the listed working coordinates. To reduce computation, we evaluate the local completion fit on a coarse grid and interpolate the corrected depths to image pixels. The final rows give evaluation grids after resampling, while other baselines retain their reconstructed sampling.}
\label{tab:resolutions}
\footnotesize
\definecolor{resLow}{HTML}{F7CBD2}
\definecolor{resMid}{HTML}{FFF0B3}
\definecolor{resHigh}{HTML}{CCEBD5}
\newcommand{\resgrid}[3]{\ifnum#1<50\cellcolor{resLow!\the\numexpr100-2*#1\relax!resMid}\else\cellcolor{resHigh!\the\numexpr2*#1-100\relax!resMid}\fi$#2\times#3$}
\setlength{\tabcolsep}{3pt}
\begin{tabular}{lccccccc}
\toprule
Method or stage & Hi4D & Perpetua & Syn4D & CHARGE & MCRL & MV-Kubric & DexYCB \\
\midrule
Raw source images & \resgrid{79}{1024}{768} & \resgrid{44}{512}{512} & \resgrid{85}{720}{1280} & \resgrid{100}{858}{2048} & \resgrid{41}{384}{512} & \resgrid{44}{512}{512} & \resgrid{53}{480}{640} \\
\midrule
\multicolumn{8}{@{}l}{\textit{Feed-forward baselines}} \\
VGGT / Pi3X & \resgrid{57}{686}{518} & \resgrid{44}{518}{518} & \resgrid{38}{294}{518} & \resgrid{35}{224}{518} & \resgrid{40}{364}{518} & \resgrid{44}{518}{518} & \resgrid{41}{392}{518} \\
VGGT-$\Omega$ & \resgrid{37}{480}{368} & \resgrid{31}{416}{416} & \resgrid{42}{320}{560} & \resgrid{44}{288}{592} & \resgrid{37}{352}{496} & \resgrid{31}{416}{416} & \resgrid{37}{368}{480} \\
OmniX & \resgrid{55}{672}{504} & \resgrid{43}{504}{504} & \resgrid{36}{280}{504} & \resgrid{33}{210}{504} & \resgrid{39}{364}{504} & \resgrid{43}{504}{504} & \resgrid{39}{378}{504} \\
$\pi^4$ & \resgrid{40}{512}{384} & \resgrid{44}{512}{512} & \resgrid{37}{288}{512} & \resgrid{33}{208}{512} & \resgrid{40}{368}{512} & \resgrid{44}{512}{512} & \resgrid{40}{384}{512} \\
ARROW-FF input & \resgrid{39}{504}{378} & \resgrid{43}{504}{504} & \resgrid{36}{280}{504} & \resgrid{33}{210}{504} & \resgrid{39}{364}{504} & \resgrid{43}{504}{504} & \resgrid{39}{378}{504} \\
ARROW-FF point grid & \resgrid{0}{252}{189} & \resgrid{0}{252}{252} & \resgrid{0}{140}{252} & \resgrid{0}{105}{252} & \resgrid{0}{182}{252} & \resgrid{0}{252}{252} & \resgrid{0}{189}{252} \\
\midrule
\multicolumn{8}{@{}l}{\textit{Monocular baselines}} \\
Any4D / V-DPM & \resgrid{57}{686}{518} & \resgrid{44}{518}{518} & \resgrid{38}{294}{518} & \resgrid{35}{224}{518} & \resgrid{40}{364}{518} & \resgrid{44}{518}{518} & \resgrid{41}{392}{518} \\
MegaSaM & \resgrid{40}{512}{384} & \resgrid{34}{440}{440} & \resgrid{44}{328}{584} & \resgrid{47}{280}{680} & \resgrid{40}{368}{520} & \resgrid{34}{440}{440} & \resgrid{40}{384}{512} \\
\midrule
\multicolumn{8}{@{}l}{\textit{Multi-view optimization baseline}} \\
Sun~\etal (SLAM / CVD) & \resgrid{40}{512}{384} & \resgrid{34}{440}{440} & \resgrid{45}{336}{592} & \resgrid{48}{288}{688} & \resgrid{41}{376}{528} & \resgrid{34}{440}{440} & \resgrid{40}{384}{512} \\
\midrule
\multicolumn{8}{@{}l}{\textit{Jojo4D components}} \\
ARROW inputs & \resgrid{39}{504}{378} & \resgrid{43}{504}{504} & \resgrid{36}{280}{504} & \resgrid{33}{210}{504} & \resgrid{39}{364}{504} & \resgrid{43}{504}{504} & \resgrid{39}{378}{504} \\
RoMa coarse matching & \resgrid{73}{800}{800} & \resgrid{71}{800}{800} & \resgrid{76}{800}{800} & \resgrid{76}{800}{800} & \resgrid{74}{800}{800} & \resgrid{71}{800}{800} & \resgrid{73}{800}{800} \\
RoMa refinement & \resgrid{100}{1280}{1280} & \resgrid{100}{1280}{1280} & \resgrid{100}{1280}{1280} & \resgrid{98}{1280}{1280} & \resgrid{100}{1280}{1280} & \resgrid{100}{1280}{1280} & \resgrid{100}{1280}{1280} \\
Camera graph / BA coordinates & \resgrid{79}{1024}{768} & \resgrid{44}{512}{512} & \resgrid{85}{720}{1280} & \resgrid{100}{856}{2048} & \resgrid{40}{368}{512} & \resgrid{44}{512}{512} & \resgrid{53}{480}{640} \\
Dense anchors & \resgrid{79}{1024}{768} & \resgrid{44}{512}{512} & \resgrid{85}{720}{1280} & \resgrid{100}{858}{2048} & \resgrid{39}{364}{512} & \resgrid{44}{512}{512} & \resgrid{53}{480}{640} \\
Local completion fit & \resgrid{0}{252}{189} & \resgrid{0}{252}{252} & \resgrid{0}{140}{252} & \resgrid{0}{105}{252} & \resgrid{0}{182}{252} & \resgrid{0}{252}{252} & \resgrid{0}{189}{252} \\
Completed point grid & \resgrid{79}{1024}{768} & \resgrid{44}{512}{512} & \resgrid{85}{720}{1280} & \resgrid{100}{858}{2048} & \resgrid{39}{364}{512} & \resgrid{44}{512}{512} & \resgrid{53}{480}{640} \\
\midrule
\multicolumn{8}{@{}l}{\textit{Evaluation grids after resampling}} \\
Jojo4D / anchors only & \resgrid{57}{686}{518} & \resgrid{44}{512}{512} & \resgrid{38}{294}{518} & \resgrid{35}{224}{518} & \resgrid{39}{364}{512} & \resgrid{44}{512}{512} & \resgrid{41}{392}{518} \\
ARROW-FF & \resgrid{57}{686}{518} & \resgrid{44}{518}{518} & \resgrid{38}{294}{518} & \resgrid{35}{224}{518} & \resgrid{40}{364}{518} & \resgrid{44}{518}{518} & \resgrid{41}{392}{518} \\
\bottomrule
\end{tabular}
\end{table*}

\section{Benchmark Datasets}
\label{sec:supp_datasets}

This section describes the seven datasets of our benchmark and the camera settings that we form from them. From every sequence we use the first 150 frames, or all frames if the sequence is shorter. The numbers of sequences and cameras are summarized in \cref{tab:datasets}, and \cref{fig:datasets} shows example frames.

\newlength{\dsw}\newlength{\dsl}
\setlength{\dsl}{14pt}
\newcommand{\dslabel}[1]{\smash{\rotatebox[origin=c]{90}{\scriptsize\begin{tabular}{@{}c@{}}#1\end{tabular}}}}
\newcommand{\dsrow}[2]{\noindent\makebox[\dsl]{\dslabel{#1}}\hspace{2pt}\raisebox{-0.5\height}{#2}\par\vspace{2pt}}
\begin{figure*}[t]
    \centering
    \begin{minipage}[t]{0.493\textwidth}
        \setlength{\dsw}{\dimexpr\linewidth-\dsl-2pt\relax}
        \dsrow{Hi4D}{\includegraphics[height=0.19038\dsw]{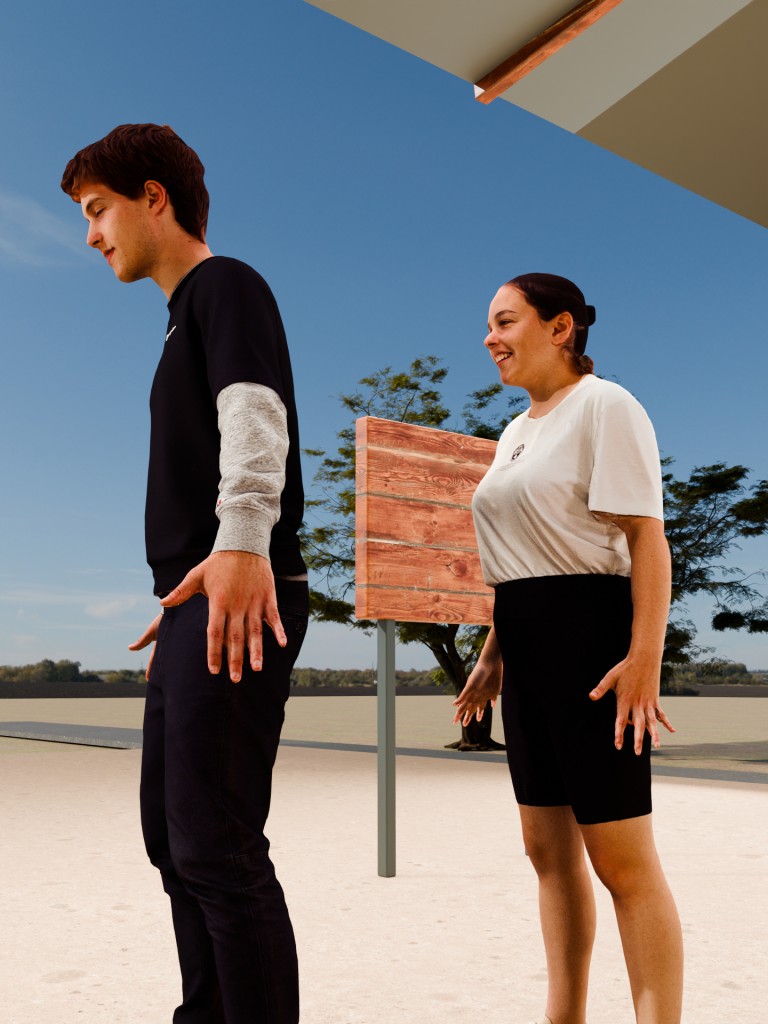}\includegraphics[height=0.19038\dsw]{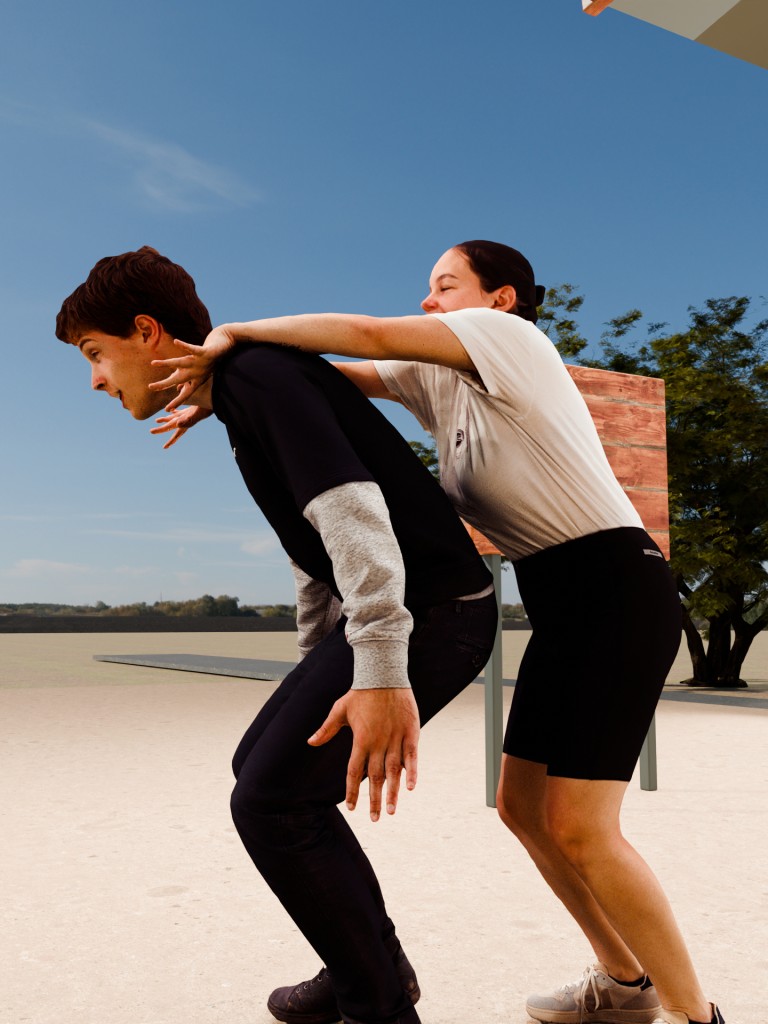}\includegraphics[height=0.19038\dsw]{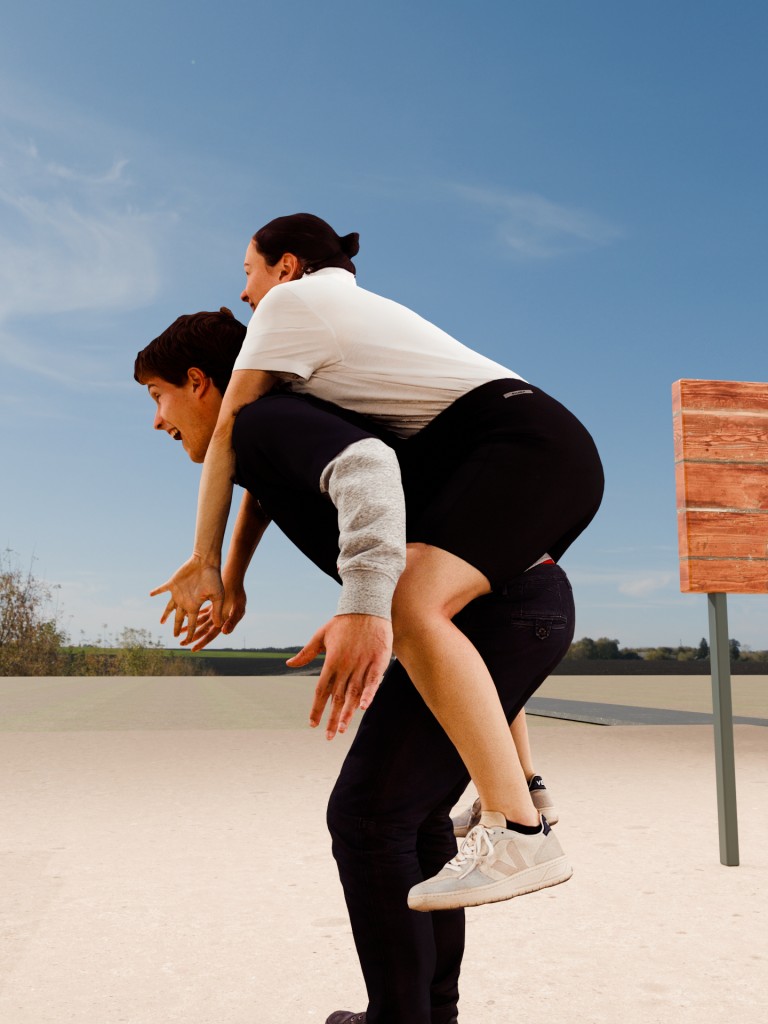}\includegraphics[height=0.19038\dsw]{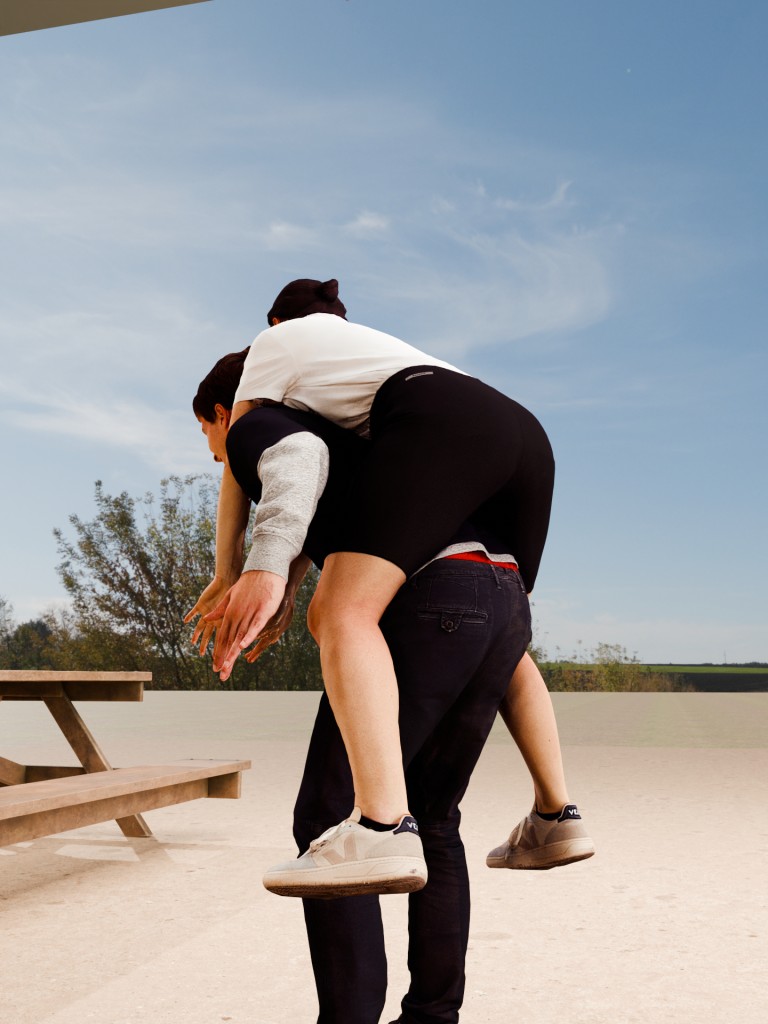}\includegraphics[height=0.19038\dsw]{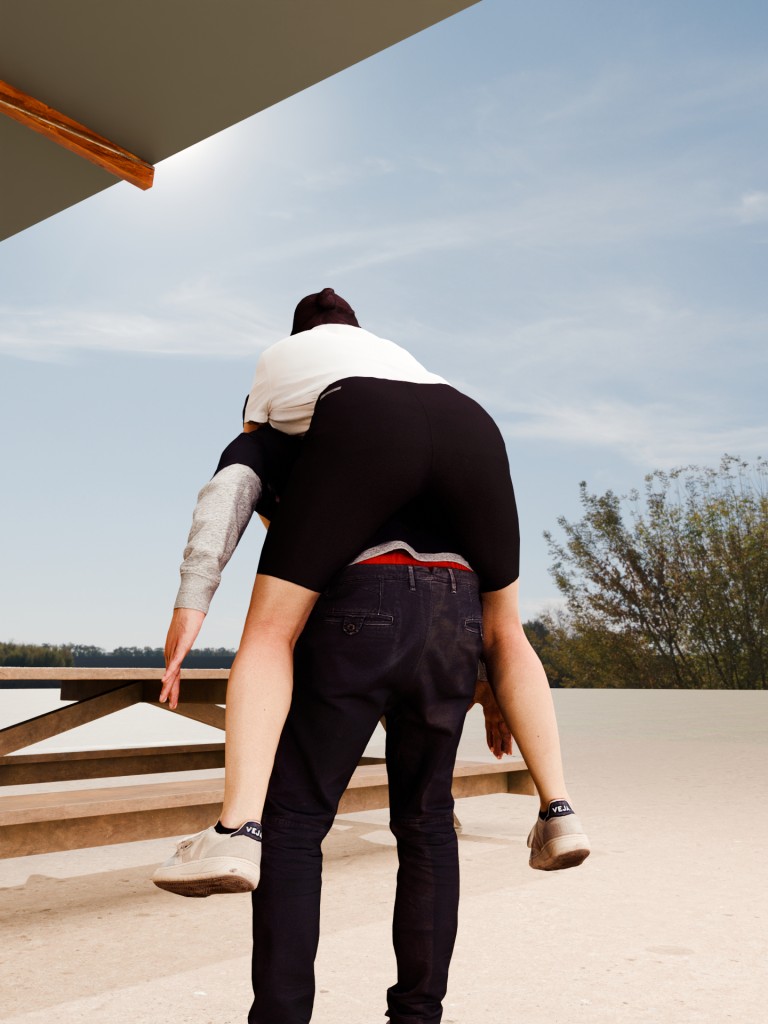}\includegraphics[height=0.19038\dsw]{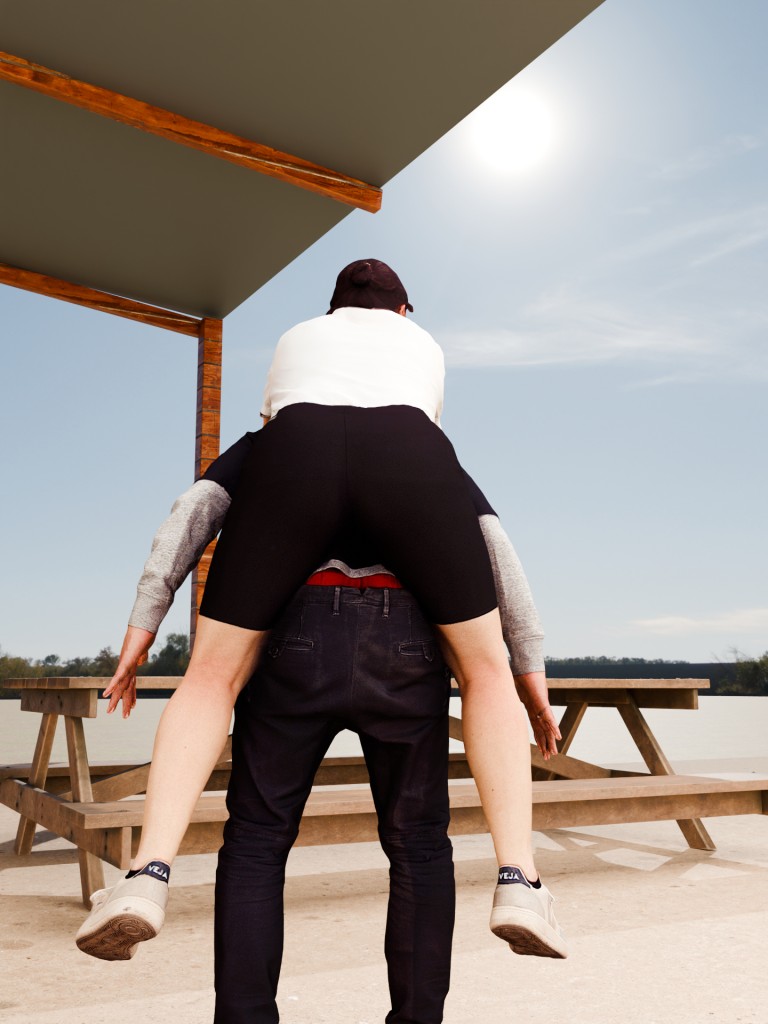}\includegraphics[height=0.19038\dsw]{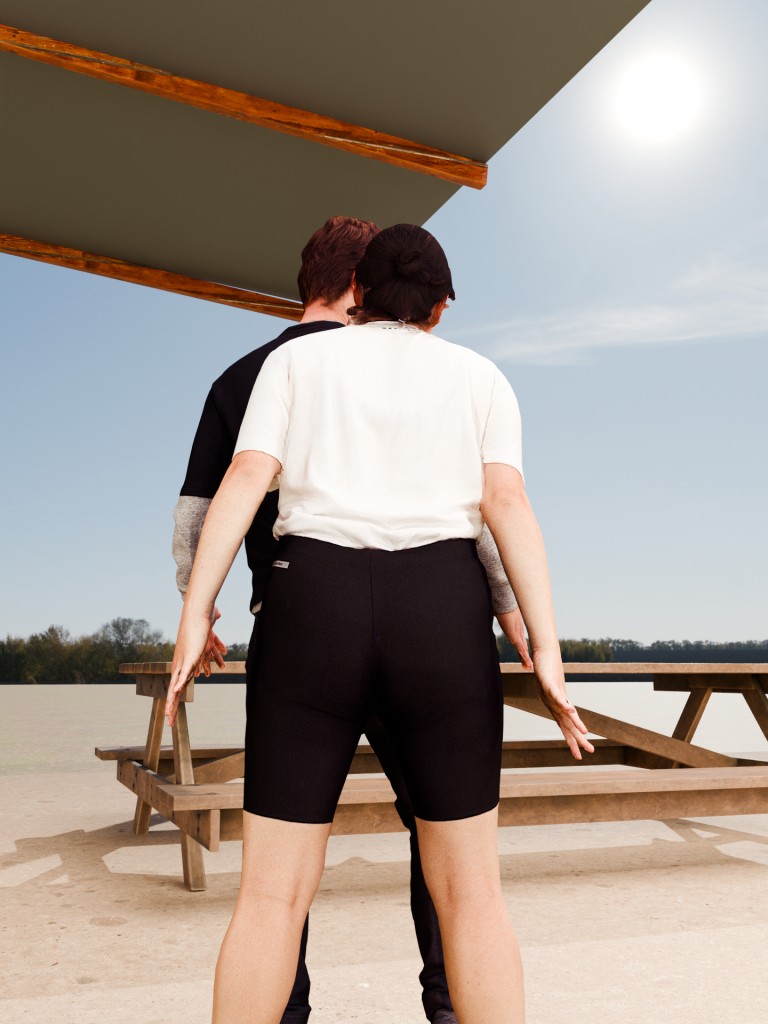}}
        \dsrow{Perpetua}{\includegraphics[height=0.19990\dsw]{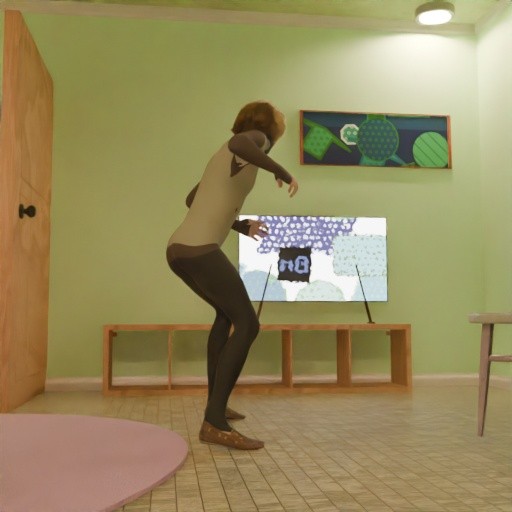}\includegraphics[height=0.19990\dsw]{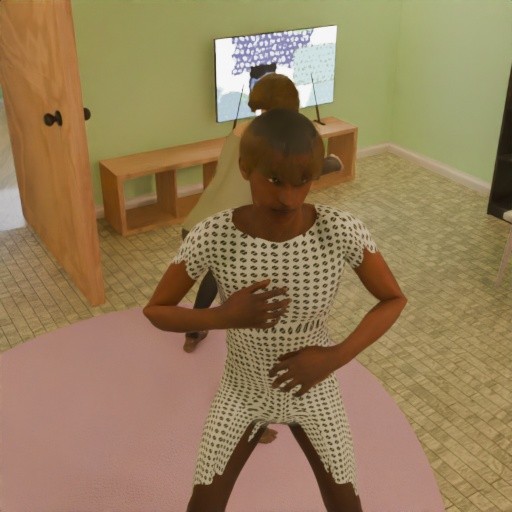}\includegraphics[height=0.19990\dsw]{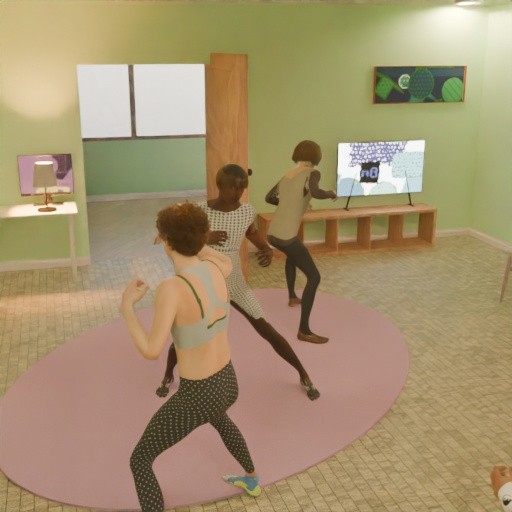}\includegraphics[height=0.19990\dsw]{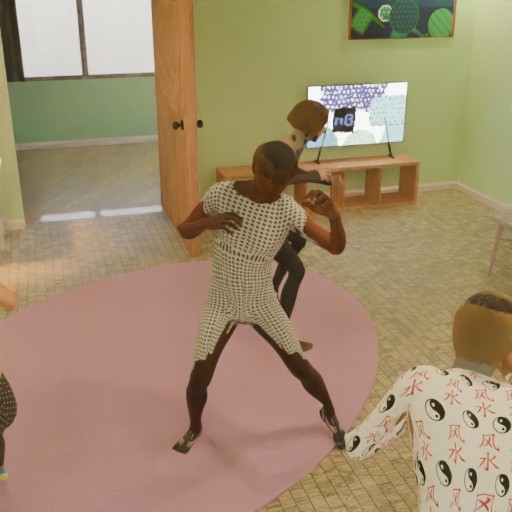}\includegraphics[height=0.19990\dsw]{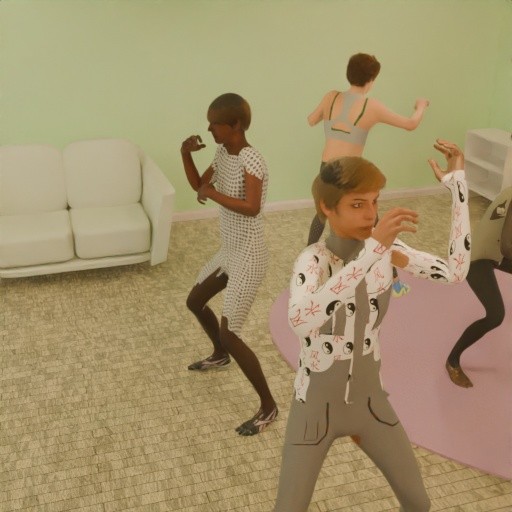}}
        \dsrow{Syn4D}{\includegraphics[height=0.18741\dsw]{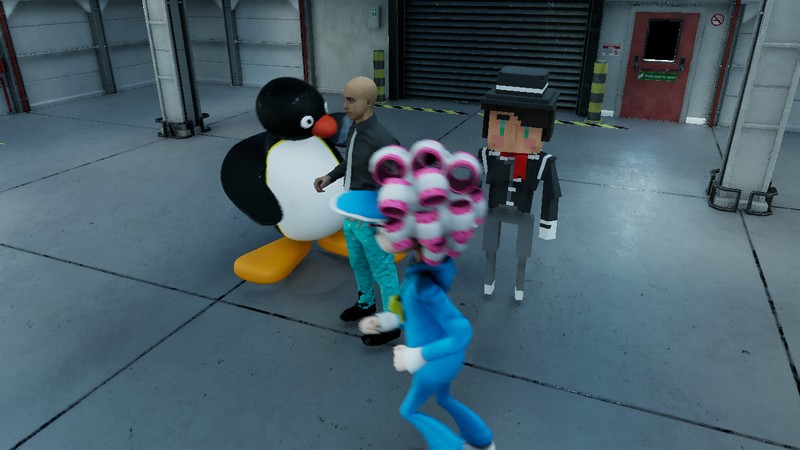}\includegraphics[height=0.18741\dsw]{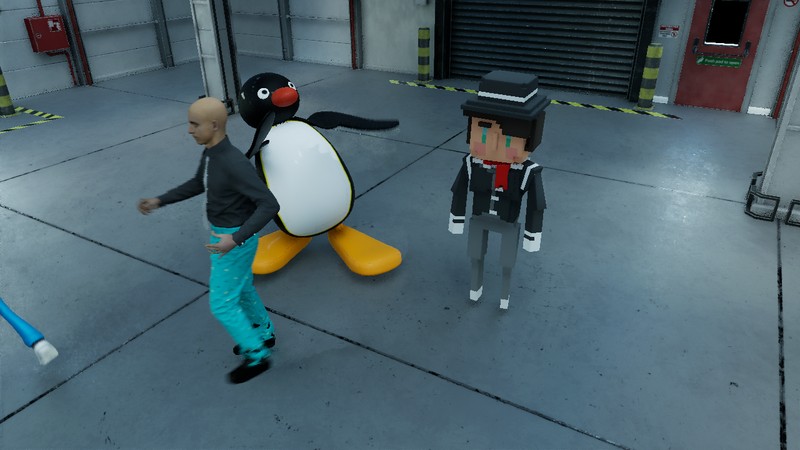}\includegraphics[height=0.18741\dsw]{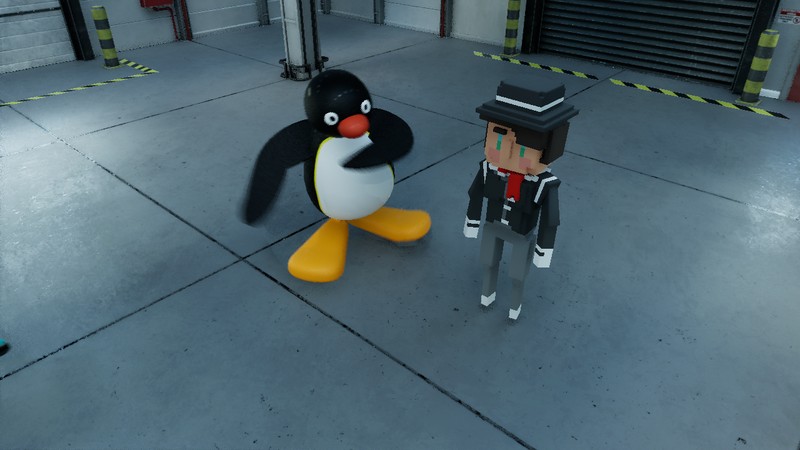}}
    \end{minipage}\hfill
    \begin{minipage}[t]{0.493\textwidth}
        \setlength{\dsw}{\dimexpr\linewidth-\dsl-2pt\relax}
        \dsrow{MultiCam-\\Robolab}{\includegraphics[height=0.17765\dsw]{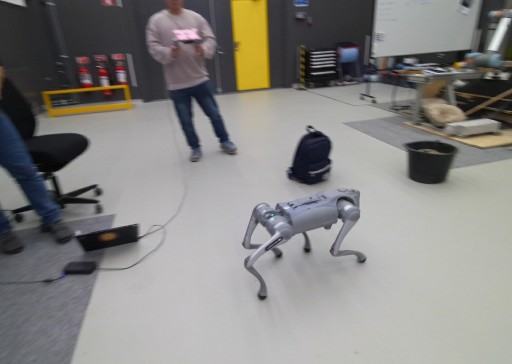}\includegraphics[height=0.17765\dsw]{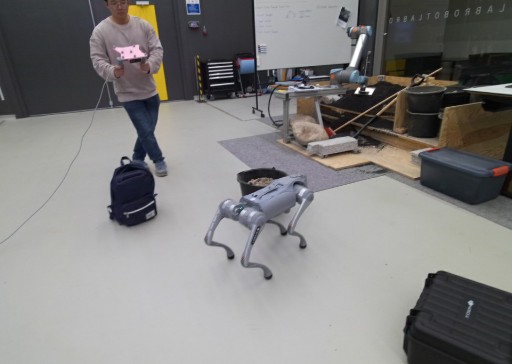}\includegraphics[height=0.17765\dsw]{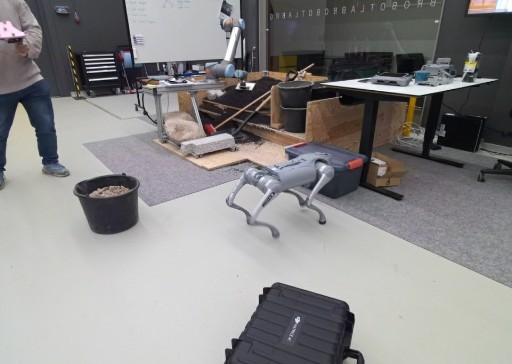}\includegraphics[height=0.17765\dsw]{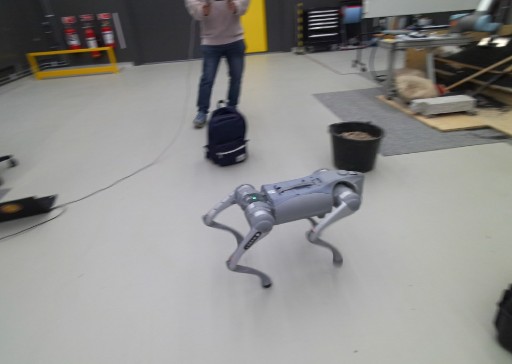}}
        \dsrow{MV-Kubric}{\includegraphics[height=0.19990\dsw]{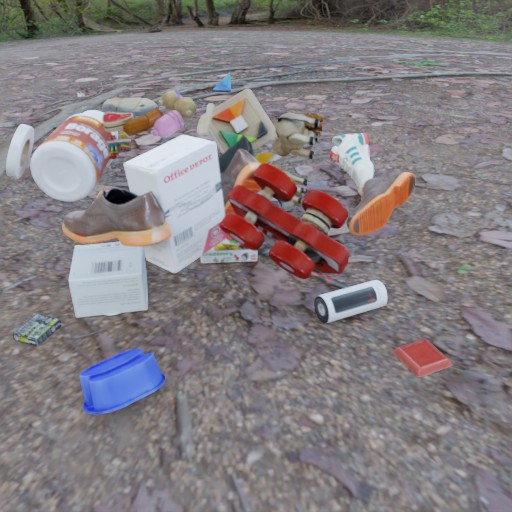}\includegraphics[height=0.19990\dsw]{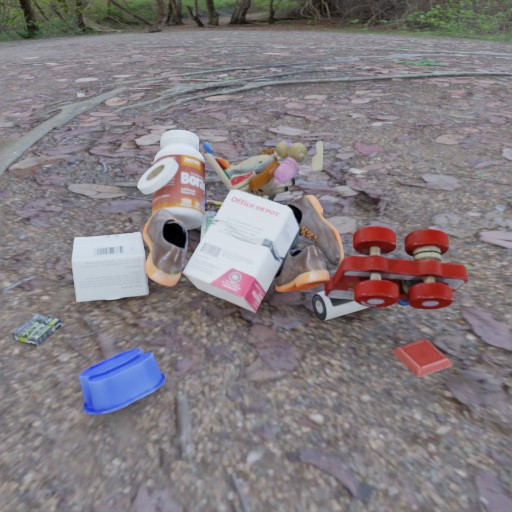}\includegraphics[height=0.19990\dsw]{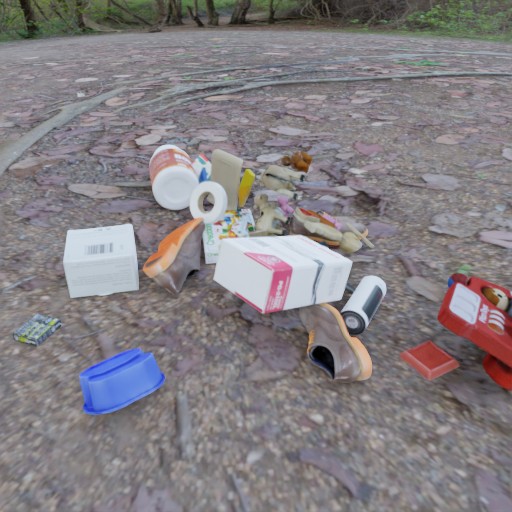}\includegraphics[height=0.19990\dsw]{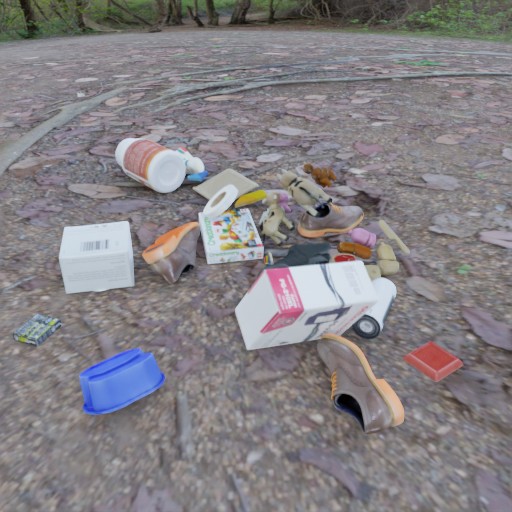}\includegraphics[height=0.19990\dsw]{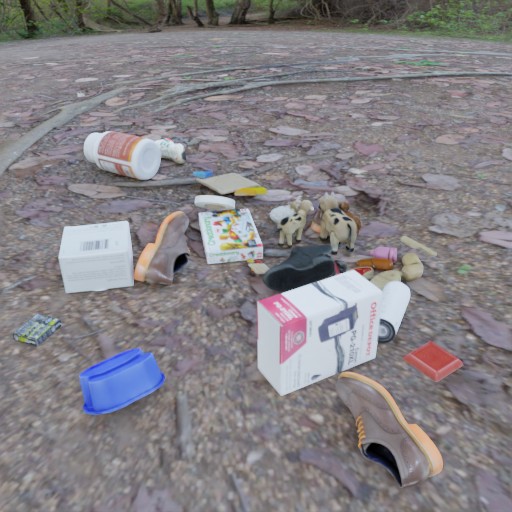}}
        \dsrow{DexYCB}{\includegraphics[height=0.18741\dsw]{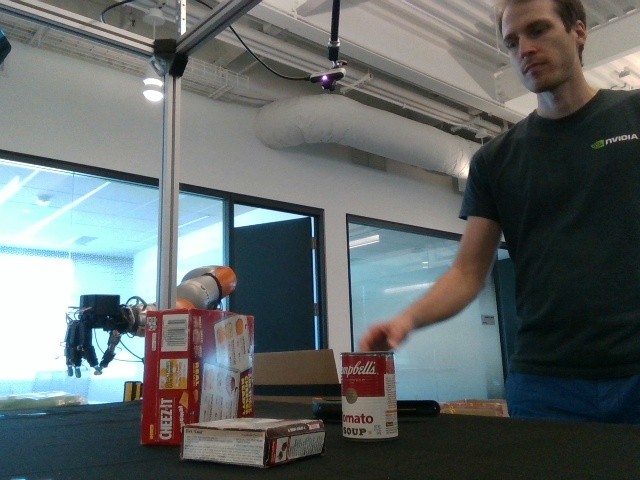}\includegraphics[height=0.18741\dsw]{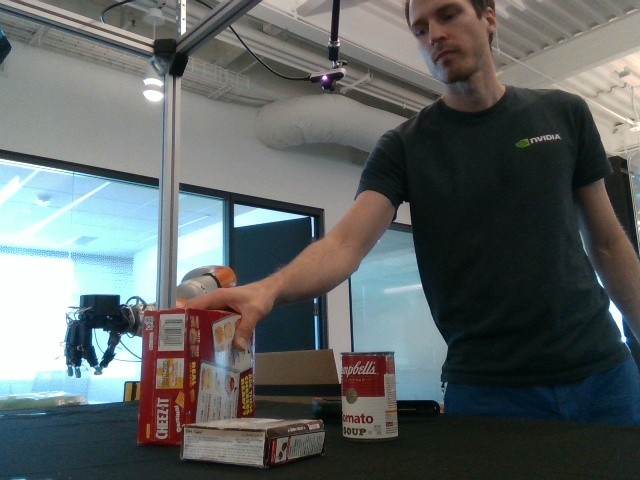}\includegraphics[height=0.18741\dsw]{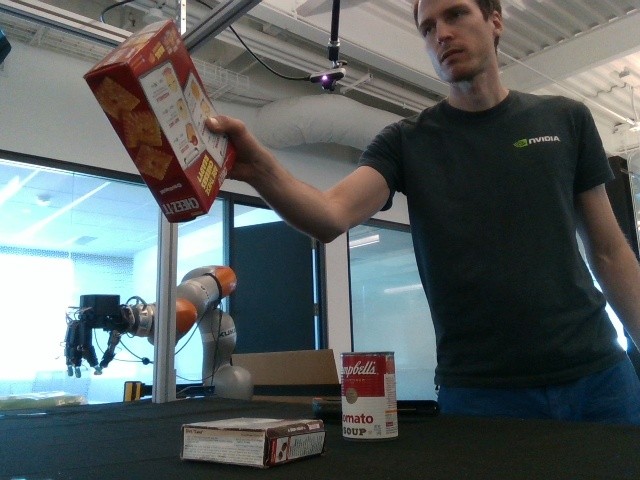}\includegraphics[height=0.18741\dsw]{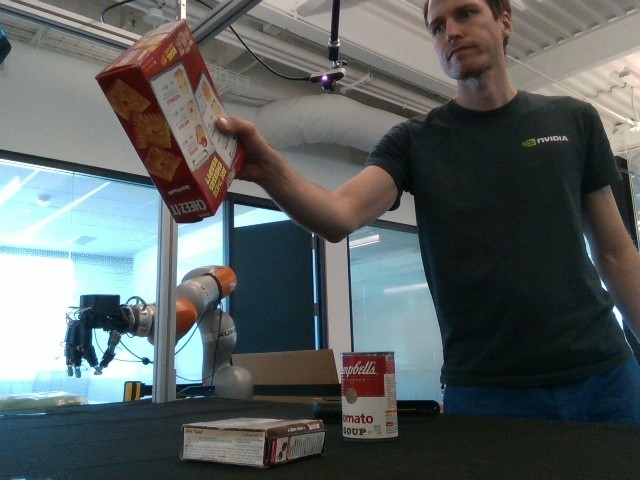}}
    \end{minipage}\hspace*{\fontdimen2\font}\par\nointerlineskip
    \begin{minipage}{0.493\textwidth}
        \setlength{\dsw}{\dimexpr\linewidth-\dsl-2pt\relax}
        \dsrow{CHARGE}{\includegraphics[height=0.20927\dsw]{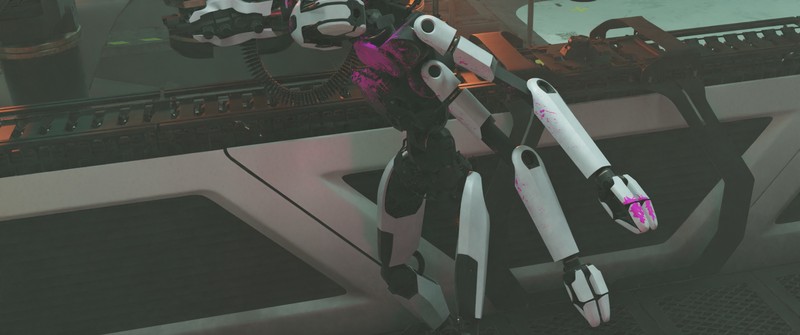}\includegraphics[height=0.20927\dsw]{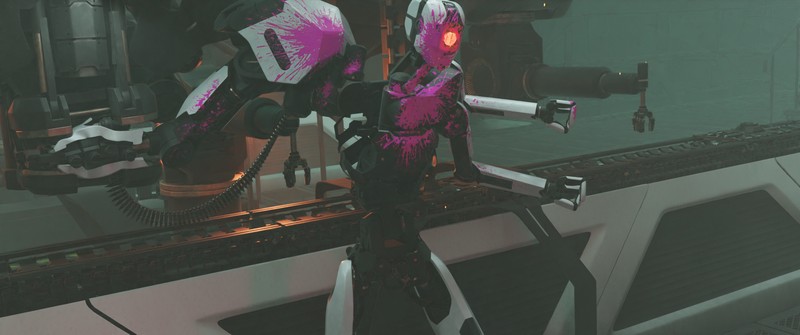}}
    \end{minipage}
    \vspace{-2pt}
    \caption{\textbf{Benchmark datasets.} Each strip shows frames from one camera in temporal order.}
    \label{fig:datasets}
\end{figure*}

\paragraph{Hi4D.} Hi4D~\cite{yin2023hi4d} contains textured 4D scans of two people in close interaction. We use the version of TAPVid-MV~\cite{koppula2026tapvidmv}, which renders the scans inside a virtual environment from four of the original static cameras and from four additional moving cameras. Our benchmark includes all 48 sequences, which are 70 to 150 frames long. The setting with two cameras uses the same two of the four moving cameras in every sequence. The setting with four cameras combines two static with two moving cameras, and the setting with eight cameras uses the full rig.

\paragraph{Perpetua.} Perpetua~\cite{koppula2026tapvidmv} is a procedurally generated synthetic dataset of indoor rooms with animated human actors. Every scene is observed by four moving cameras. We use all 21 scenes from five different room types. Besides the setting with all four cameras, we form a setting with the two cameras whose views overlap the most.
\paragraph{Syn4D.} Syn4D~\cite{jiang2026syn4d} uses Unreal Engine to render animated people, animals, and other characters from eight viewpoints. Our two-camera setting uses two moving cameras, while the four- and eight-camera settings use equal numbers of static and moving cameras. We evaluate each of the 20 warehouse sequences and each of the 20 winter sequences with two, four, and eight cameras. All eight cameras move in the five office sequences, so we evaluate them only in the two-camera setting.

\paragraph{CHARGE.} CHARGE~\cite{nazarczuk2026charge} renders shots from the Blender animated short film \emph{Charge} from multiple viewpoints for novel view synthesis. We use all eight shots, with evaluated clips ranging from 93 to 150 frames. We use the same camera subsets for every shot. The two-camera setting uses the fast spline and fast random-walk cameras. The four-camera setting adds the first two cameras of the static rig. The eight-camera setting uses the first four static cameras and all four moving cameras: two follow spline paths and two follow random walks, with one fast and one slow camera in each pair.

\paragraph{MultiCamRobolab.} MultiCamRobolab~\cite{sun2026multiview} contains real-world RGB-D recordings of robots and people in a laboratory. Every sequence is captured by two moving cameras, whose poses are measured by a motion capture system. We use 18 of its 19 sequences and exclude one in which a camera provides almost no valid depth. The undistorted images contain black borders at the top and bottom, which we crop for all methods.

\paragraph{MV-Kubric.} MV-Kubric~\cite{rajic2025mvtracker} consists of synthetic scenes generated with Kubric~\cite{greff2022kubric}, in which rigid objects fall onto the ground and collide. All cameras of this dataset are static. We use 10 scenes with all ten cameras.

\paragraph{DexYCB.} DexYCB~\cite{chao2021dexycb} contains real-world RGB-D recordings of people grasping objects on a table. Every sequence is captured by eight calibrated cameras, all of which are static. We use 10 sequences with all eight cameras.

\section{Suppression of Dynamic Tracks}
\label{sec:supp_dynamic_tracks}

Our bundle adjustment does not use a mask of moving objects. Multi-instant tracks on moving content are instead suppressed by the tight Cauchy loss and by the reprojection filter between the two passes (\cref{sec:bundle_adjustment}). This section analyzes how well this suppression works.

We select ten sequences each from Hi4D, Perpetua, and Syn4D and all eight shots of CHARGE, the four datasets with moving cameras that provide ground-truth masks of dynamic content. We run every sequence with two and with four cameras. A track is labeled as dynamic if its source pixel lies on moving content and as static otherwise. Note that since the masks are defined per object, a point on a labeled object counts as dynamic even if it stays at rest for the whole track, such as the foot of a standing person.

\begin{figure*}[t]
    \centering
    \includegraphics[width=\linewidth]{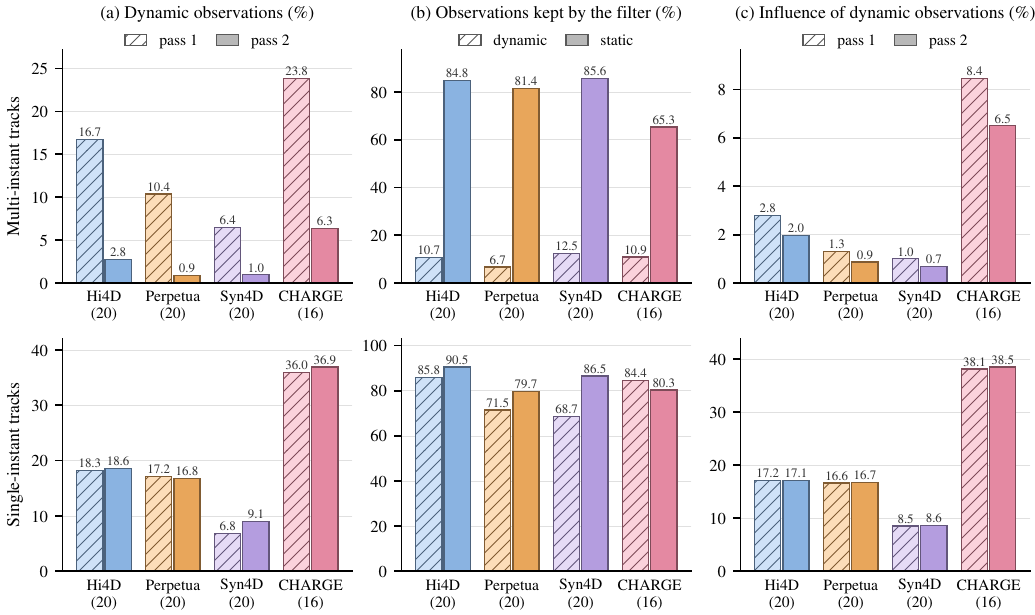}
    \caption{\textbf{Suppression of dynamic tracks.} On all four datasets, the filter removes most observations on moving objects from multi-instant tracks but retains most of them in single-instant tracks. Multi-instant tracks (top) and single-instant tracks (bottom). Per dataset, (a) the share of observations on dynamic content that enter the first and the second pass of bundle adjustment, (b) the share of dynamic and static observations that the reprojection filter between the passes retains, and (c) the share of total influence due to dynamic observations within each track type at the end of each pass (where an observation's influence is the norm of its loss gradient with respect to the scaled camera parameters). We average the percentages over clips within each dataset, counting each two-camera or four-camera clip once.}
    \label{fig:supp_dynamic_bars}
\end{figure*}

\begin{figure*}[t]
    \centering
    \includegraphics[width=\linewidth]{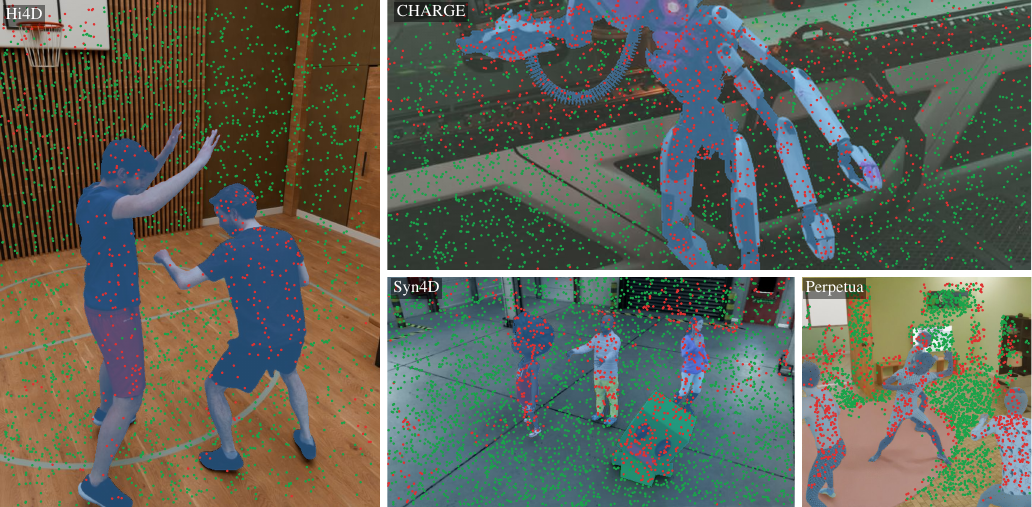}
    \caption{\textbf{The reprojection filter removes observations on moving objects while retaining much of the background.} Each dot is an observation from a multi-instant track in the first pass. Green dots enter the second pass. Red dots are removed. Blue marks the ground-truth mask of moving objects.}
    \label{fig:supp_dynamic_tracks}
\end{figure*}

Between the two passes, the filter preferentially removes observations on moving objects from multi-instant tracks (\cref{fig:supp_dynamic_bars,fig:supp_dynamic_tracks}). For single-instant tracks, it retains dynamic and static observations at similar rates on most datasets. Dynamic multi-instant observations already have little influence on the cameras after the first pass and even less after the second. We measure the influence of an observation by the gradient of its contribution to the bundle adjustment error with respect to the camera parameters, each scaled such that a unit change moves the observations of its camera by one pixel on average.

\begin{figure}[t]
    \centering
    \includegraphics[width=\linewidth]{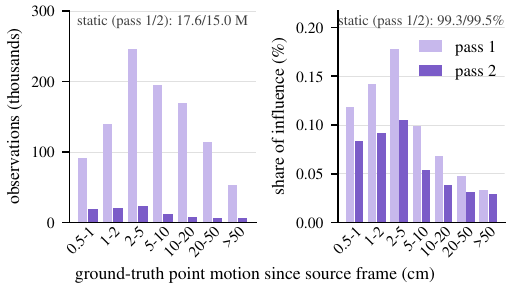}
    \caption{\textbf{Ground-truth motion of multi-instant observations.} For each pass, we group observations from all 20 Syn4D clips by how far the point moved since its source frame. Bars show the total count in each bin (left) and its share of the total influence on the cameras (right). The static labels give the values for pass 1 and pass 2, in that order, for motion below $0.5$\,cm (M = million).}
    \label{fig:supp_track_motion}
\end{figure}

The object masks do not tell us if and how far a point actually moves. On Syn4D, mesh-face trajectories let us measure this displacement from the source frame to each observation. When the point moves more than a few centimeters, the filter removes most of its multi-instant observations. Remaining points have little influence on the cameras (\cref{fig:supp_track_motion}).

To test whether the remaining dynamic observations harm the cameras, we rerun our bundle adjustment with ground-truth masks (\cref{tab:supp_dynamic_oracle}). Removing all dynamic multi-instant tracks before the first pass lowers the mean rotation error over the clips from $0.052^\circ$ to $0.048^\circ$, and removing only the retained dynamic observations before the second pass lowers it to $0.050^\circ$. 

\begin{table}[t]
\caption{\textbf{Removing dynamic observations with ground-truth masks.} Mean RPE-R\mksl{}RPE-t in degrees of our cameras and of two reruns that remove the dynamic multi-instant observations with the ground-truth masks, either all of them before the first pass or the retained ones before the second pass. Either removal changes the mean rotation error of each dataset by at most $0.012^\circ$. Four Perpetua four-camera clips on which our cameras have errors above $1^\circ$ are excluded as outliers.}
\label{tab:supp_dynamic_oracle}
\mksetup
\footnotesize\setlength{\tabcolsep}{3pt}
\resizebox{\ifdim\width>\linewidth\linewidth\else\width\fi}{!}{%
\begin{tabular}{@{}lcccc}
\toprule
RPE-R$\,\downarrow$\mksl{}RPE-t$\,\downarrow$ & \cellcolor{mkhi4d}\textbf{Hi4D} & \cellcolor{mkperp}\textbf{Perpetua} & \cellcolor{mksyn}\textbf{Syn4D} & \cellcolor{mkcharge}\textbf{CHARGE} \\
\midrule
Jojo4D (ours) & .022\mksl{}.020 & .088\mksl{}.145 & .048\mksl{}.058 & .057\mksl{}.225 \\
GT removal before pass 1 & .021\mksl{}.019 & .082\mksl{}.137 & .049\mksl{}.060 & .045\mksl{}.169 \\
GT removal before pass 2 & .021\mksl{}.019 & .085\mksl{}.140 & .048\mksl{}.059 & .050\mksl{}.174 \\
\bottomrule
\end{tabular}}
\end{table}

\section{Runtime and Memory}
\label{sec:supp_runtime}
\definecolor{rtLow}{HTML}{F7CBD2}\definecolor{rtMid}{HTML}{FFF0B3}\definecolor{rtHigh}{HTML}{CCEBD5}\newcommand{\rtheat}[2]{\ifnum#1<50\cellcolor{rtLow!\the\numexpr100-2*#1\relax!rtMid}\else\cellcolor{rtHigh!\the\numexpr2*#1-100\relax!rtMid}\fi#2}
\newcommand{\runtimeSD}[1]{\,{\scriptsize$\pm$\,#1}}

We compare runtime and peak memory on four Hi4D sequences drawn at random without replacement from the eight with at least $150$ frames in all three camera configurations. Each run uses one NVIDIA GH200 Grace Hopper superchip with a $96$\,GB H100 GPU and a $72$-core Grace CPU. We measure whole-process wall-clock time, including data and model loading but excluding evaluation.

\begin{figure}[htpb]
\centering
\includegraphics[width=\linewidth]{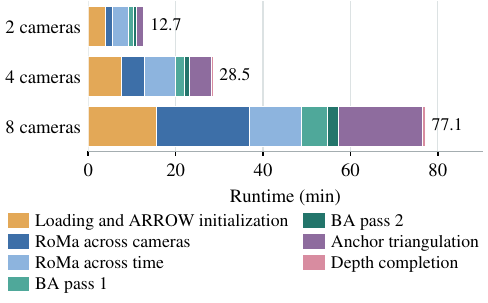}
\caption{\textbf{Runtime breakdown of Jojo4D.} Bars show mean stage times across four Hi4D sequences with $150$ frames per camera. Matching across cameras and anchor triangulation grow fastest with the number of cameras and take about half of the runtime with eight cameras.}
\label{fig:runtime_stages}
\end{figure}

\Cref{fig:runtime_stages} breaks down Jojo4D's runtime by stage for a 150-frame sequence. Matching across cameras grows fastest with the number of cameras, since every pair of cameras is matched at every time step. With $150$ frames per camera, it takes $1.5$ minutes with two cameras and $21.2$ minutes with eight cameras, the largest share of the total of $77.1$ minutes. We believe that a sparser correspondence graph, which matches only a subset of the camera pairs, could reach similar accuracy at a lower cost.

\Cref{tab:runtime} compares how runtime and GPU memory scale across methods. Jojo4D takes more time per step than most baselines. With eight cameras, however, its time per step stays nearly constant as the clips get longer: $27.1$\,s at $30$ frames and $30.8$\,s at $150$. For ARROW-FF, the time per step grows almost linearly with clip length, from $25.3$\,s to $104.0$\,s, so five times as many frames take about twenty times as long and its total runtime grows roughly quadratically over this range. The GPU memory of Jojo4D also scales favorably. With eight cameras and $150$ frames, its peak is $36.7$\,GiB, compared to $48.3$\,GiB for VGGT-$\Omega$, the baseline with the lowest peak.

\begin{table*}[t]
\centering
\caption{\textbf{Runtime and memory on Hi4D.} Runtimes are seconds per synchronized time step, given as means $\pm$ one sample standard deviation across the four sequences. GPU memory values are the largest peaks across the four sequences.}
\label{tab:runtime}
\small
\setlength{\tabcolsep}{2.5pt}
\newcommand{\runtimebranch}{\hspace{0.5em}\rule[-0.8ex]{0.3pt}{2.8ex}\rule[0.4ex]{0.7em}{0.3pt}\hspace{0.3em}}
\newcommand{\runtimeend}{\hspace{0.5em}\rule[0.4ex]{0.3pt}{1.6ex}\rule[0.4ex]{0.7em}{0.3pt}\hspace{0.3em}}
\resizebox{\ifdim\width>\linewidth\linewidth\else\width\fi}{!}{%
\begin{tabular}{@{}lccc@{\hspace{8pt}}ccc@{\hspace{8pt}}ccc@{}}
\toprule
\textbf{Runtime (s per time step)$\downarrow$} & \multicolumn{3}{c}{2 cameras} & \multicolumn{3}{c}{4 cameras} & \multicolumn{3}{c}{8 cameras} \\
\cmidrule(lr){2-4}\cmidrule(lr){5-7}\cmidrule(l){8-10}
Frames per camera & 30 & 90 & 150 & 30 & 90 & 150 & 30 & 90 & 150 \\
\midrule
VGGT~\cite{wang2025vggt} & \rtheat{77}{0.95\runtimeSD{0.10}} & \rtheat{79}{0.88\runtimeSD{0.21}} & \rtheat{81}{0.74\runtimeSD{0.08}} & \rtheat{64}{2.19\runtimeSD{0.16}} & \rtheat{71}{1.36\runtimeSD{0.04}} & \rtheat{70}{1.44\runtimeSD{0.09}} & \rtheat{56}{3.43\runtimeSD{0.35}} & \rtheat{56}{3.52\runtimeSD{0.16}} & \rtheat{56}{3.50\runtimeSD{0.14}} \\
Pi3X~\cite{wang2026pi3} & \rtheat{77}{0.95\runtimeSD{0.10}} & \rtheat{84}{0.64\runtimeSD{0.02}} & \rtheat{87}{0.54\runtimeSD{0.03}} & \rtheat{70}{1.48\runtimeSD{0.51}} & \rtheat{71}{1.42\runtimeSD{0.21}} & \rtheat{73}{1.24\runtimeSD{0.05}} & \rtheat{60}{2.71\runtimeSD{0.11}} & \rtheat{58}{3.00\runtimeSD{0.22}} & \rtheat{59}{2.92\runtimeSD{0.13}} \\
OmniX~\cite{jiang2026omnix} & \rtheat{81}{0.77\runtimeSD{0.11}} & \rtheat{85}{0.58\runtimeSD{0.02}} & \rtheat{87}{0.53\runtimeSD{0.01}} & \rtheat{74}{1.15\runtimeSD{0.05}} & \rtheat{74}{1.15\runtimeSD{0.01}} & \rtheat{75}{1.12\runtimeSD{0.01}} & \rtheat{61}{2.55\runtimeSD{0.00}} & \rtheat{59}{2.82\runtimeSD{0.02}} & \rtheat{59}{2.88\runtimeSD{0.09}} \\
VGGT-$\Omega$~\cite{wang2026vggtomega} & \rtheat{81}{0.76\runtimeSD{0.04}} & \rtheat{96}{0.30\runtimeSD{0.02}} & \rtheat{100}{0.24\runtimeSD{0.01}} & \rtheat{81}{0.78\runtimeSD{0.08}} & \rtheat{89}{0.47\runtimeSD{0.01}} & \rtheat{90}{0.43\runtimeSD{0.01}} & \rtheat{76}{1.05\runtimeSD{0.05}} & \rtheat{79}{0.88\runtimeSD{0.02}} & \rtheat{79}{0.88\runtimeSD{0.03}} \\
ARROW-FF~\cite{fradlin2026arrow} & \rtheat{63}{2.29\runtimeSD{0.00}} & \rtheat{50}{4.89\runtimeSD{0.06}} & \rtheat{43}{7.77\runtimeSD{0.11}} & \rtheat{45}{6.97\runtimeSD{0.13}} & \rtheat{28}{18.5\runtimeSD{0.07}} & \rtheat{20}{30.8\runtimeSD{0.27}} & \rtheat{23}{25.3\runtimeSD{0.45}} & \rtheat{6}{73.9\runtimeSD{0.14}} & \rtheat{0}{104.0\runtimeSD{1.02}} \\
\addlinespace[3pt]
Jojo4D (ours), total & \rtheat{49}{5.33\runtimeSD{0.06}} & \rtheat{50}{4.92\runtimeSD{0.34}} & \rtheat{50}{5.09\runtimeSD{0.15}} & \rtheat{38}{10.4\runtimeSD{0.17}} & \rtheat{37}{11.2\runtimeSD{0.29}} & \rtheat{36}{11.4\runtimeSD{0.13}} & \rtheat{22}{27.1\runtimeSD{0.15}} & \rtheat{19}{32.3\runtimeSD{0.39}} & \rtheat{20}{30.8\runtimeSD{0.36}} \\
\runtimebranch Camera initialization & \rtheat{69}{1.58\runtimeSD{0.05}} & \rtheat{71}{1.36\runtimeSD{0.01}} & \rtheat{71}{1.37\runtimeSD{0.01}} & \rtheat{61}{2.58\runtimeSD{0.03}} & \rtheat{61}{2.57\runtimeSD{0.02}} & \rtheat{60}{2.65\runtimeSD{0.02}} & \rtheat{51}{4.77\runtimeSD{0.04}} & \rtheat{50}{5.04\runtimeSD{0.02}} & \rtheat{49}{5.17\runtimeSD{0.02}} \\
\runtimebranch Matching and bundle adjustment & \rtheat{58}{3.06\runtimeSD{0.04}} & \rtheat{60}{2.80\runtimeSD{0.33}} & \rtheat{59}{2.87\runtimeSD{0.15}} & \rtheat{46}{6.19\runtimeSD{0.19}} & \rtheat{46}{6.37\runtimeSD{0.23}} & \rtheat{46}{6.29\runtimeSD{0.12}} & \rtheat{29}{17.4\runtimeSD{0.22}} & \rtheat{28}{18.9\runtimeSD{0.24}} & \rtheat{30}{16.7\runtimeSD{0.35}} \\
\runtimeend Dense reconstruction & \rtheat{83}{0.69\runtimeSD{0.07}} & \rtheat{81}{0.76\runtimeSD{0.02}} & \rtheat{79}{0.85\runtimeSD{0.02}} & \rtheat{68}{1.64\runtimeSD{0.05}} & \rtheat{63}{2.26\runtimeSD{0.06}} & \rtheat{62}{2.48\runtimeSD{0.03}} & \rtheat{50}{4.96\runtimeSD{0.05}} & \rtheat{42}{8.33\runtimeSD{0.24}} & \rtheat{40}{8.94\runtimeSD{0.14}} \\
\midrule
\textbf{Peak GPU memory (GiB)$\downarrow$} & \multicolumn{3}{c}{2 cameras} & \multicolumn{3}{c}{4 cameras} & \multicolumn{3}{c}{8 cameras} \\
\cmidrule(lr){2-4}\cmidrule(lr){5-7}\cmidrule(l){8-10}
Frames per camera & 30 & 90 & 150 & 30 & 90 & 150 & 30 & 90 & 150 \\
\midrule
VGGT~\cite{wang2025vggt} & \rtheat{63}{36.8} & \rtheat{2}{86.1} & \rtheat{0}{87.3} & \rtheat{35}{59.7} & \rtheat{1}{86.5} & \rtheat{1}{86.5} & \rtheat{1}{86.5} & \rtheat{1}{86.5} & \rtheat{1}{86.5} \\
Pi3X~\cite{wang2026pi3} & \rtheat{25}{67.7} & \rtheat{17}{73.5} & \rtheat{17}{74.1} & \rtheat{18}{72.7} & \rtheat{17}{73.9} & \rtheat{17}{73.9} & \rtheat{17}{73.9} & \rtheat{17}{73.9} & \rtheat{17}{73.9} \\
OmniX~\cite{jiang2026omnix} & \rtheat{74}{28.6} & \rtheat{29}{63.9} & \rtheat{24}{68.4} & \rtheat{56}{42.3} & \rtheat{24}{68.4} & \rtheat{24}{68.4} & \rtheat{24}{68.1} & \rtheat{24}{68.1} & \rtheat{24}{68.1} \\
VGGT-$\Omega$~\cite{wang2026vggtomega} & \rtheat{96}{10.4} & \rtheat{84}{20.7} & \rtheat{70}{31.2} & \rtheat{90}{15.5} & \rtheat{64}{36.5} & \rtheat{48}{48.7} & \rtheat{77}{26.0} & \rtheat{49}{48.3} & \rtheat{49}{48.3} \\
ARROW-FF~\cite{fradlin2026arrow} & \rtheat{92}{14.0} & \rtheat{78}{25.0} & \rtheat{66}{35.0} & \rtheat{85}{19.2} & \rtheat{57}{41.7} & \rtheat{27}{65.5} & \rtheat{72}{30.2} & \rtheat{13}{77.3} & \rtheat{11}{78.4} \\
\addlinespace[3pt]
Jojo4D (ours), total & \rtheat{78}{25.2} & \rtheat{68}{33.4} & \rtheat{68}{33.4} & \rtheat{75}{27.5} & \rtheat{68}{33.4} & \rtheat{68}{33.4} & \rtheat{73}{29.1} & \rtheat{72}{29.7} & \rtheat{63}{36.7} \\
\runtimebranch Camera initialization & \rtheat{78}{25.2} & \rtheat{68}{33.4} & \rtheat{68}{33.4} & \rtheat{75}{27.5} & \rtheat{68}{33.4} & \rtheat{68}{33.4} & \rtheat{73}{29.1} & \rtheat{73}{29.1} & \rtheat{73}{29.1} \\
\runtimebranch Matching and bundle adjustment & \rtheat{84}{20.4} & \rtheat{84}{20.5} & \rtheat{84}{20.4} & \rtheat{85}{19.2} & \rtheat{85}{19.6} & \rtheat{85}{19.3} & \rtheat{80}{23.3} & \rtheat{79}{24.0} & \rtheat{80}{23.4} \\
\runtimeend Dense reconstruction & \rtheat{100}{7.6} & \rtheat{99}{8.7} & \rtheat{96}{10.9} & \rtheat{97}{10.3} & \rtheat{93}{13.4} & \rtheat{89}{16.6} & \rtheat{84}{20.4} & \rtheat{72}{29.7} & \rtheat{63}{36.7} \\
\bottomrule
\end{tabular}}
\end{table*}

\section{Detailed Benchmark Results}
\label{sec:supp_results}

\subsection{Camera estimation}
\label{sec:supp_cameras}

\paragraph{Metrics.}

\emph{RPE-R and RPE-t.} The relative pose errors~\cite{wang2025vggt} are the camera metrics of the main paper (\cref{tab:cameras}). They compare the relative pose of two images with the ground truth and require no alignment. For two images $a$ and $b$ with camera-to-world rotations $R_a$, $R_b$ and camera centers $c_a$, $c_b$, let $R_{ab} = R_a^\top R_b$ and $d_{ab} = R_a^\top (c_b - c_a)$ denote the ground-truth relative rotation and translation vector, and $\hat{R}_{ab}$, $\hat{d}_{ab}$ their estimates. The errors of the pair are
\begin{align}
    e_R(a,b) &= \arccos\Big( \tfrac{1}{2} \big( \operatorname{tr}\big( \hat{R}_{ab}^\top R_{ab} \big) - 1 \big) \Big), \label{eq:rot_error} \\
    e_t(a,b) &= \arccos\bigg( \frac{\big| \hat{d}_{ab}^\top d_{ab} \big|}{\| \hat{d}_{ab} \|_2 \, \| d_{ab} \|_2} \bigg). \label{eq:trans_error}
\end{align}
 We evaluate all pairs of images of a scene, across cameras and time steps, and report the mean errors in degrees as RPE-R and RPE-t. Every image contributes one pose, so a static camera is counted once per time step. We exclude pairs from the same static camera at different times from both metrics because their translation direction is undefined.

\emph{ATE$_n$.} The normalized absolute trajectory error measures the positions of all cameras of a scene jointly. We align the camera centers of all cameras and time steps to the ground truth with a single similarity transform and compute the root mean square distance of the aligned centers. Dividing it by the extent of the ground-truth cameras, which we define as the diagonal of the bounding box of their centers, yields ATE$_n$ in percent. The main paper reports its average over all datasets, and \cref{tab:supp_camera_metrics} lists it per dataset.

\emph{RRA$_1$, RTA$_1$ and AUC$_5$.} These metrics measure how many image pairs are estimated accurately. RRA$_1$ and RTA$_1$ are the shares of image pairs whose relative rotation and translation errors (\cref{eq:rot_error,eq:trans_error}) lie below $1^\circ$. AUC$_5$ combines both errors over a range of thresholds~\cite{wang2025vggt}. A pair counts as accurate at a threshold $\tau$ if both of its errors lie below $\tau$, and AUC$_5$ averages the share of accurate pairs over the thresholds from $1^\circ$ to $5^\circ$,
\begin{align}
    \mathrm{AUC}_5 &= \frac{1}{5} \sum_{\tau = 1^\circ}^{5^\circ} r(\tau), \label{eq:pose_auc} \\
    r(\tau) &= \frac{\big| \{ (a,b) \in \mathcal{Q} \mid \max( e_R, e_t ) < \tau \} \big|}{| \mathcal{Q} |}, \nonumber
\end{align}
where $r(\tau)$ is the share of the evaluated image pairs $\mathcal{Q}$ that are accurate at $\tau$. It reaches $100\%$ only if both errors of every pair lie below $1^\circ$.

\emph{Intrinsics.} We report the relative error of the focal length ($f$) and the absolute error of the field of view (FoV), both averaged over the two image axes and all cameras of a scene, as well as the distance between the estimated and true principal point (pp) in percent of the image diagonal, averaged over all cameras of a scene. All three compare the estimated intrinsics after scaling them to the resolution of the ground-truth images. We compute FoV from focal length and image size alone, so it does not measure errors in the principal point. The main paper reports the averages of the first two metrics over all datasets.

\paragraph{Results.}
Complementing the camera results of the main paper (\cref{tab:cameras}), \cref{tab:supp_camera_metrics} reports additional pose metrics per dataset, \cref{tab:supp_intrinsics} the errors of the intrinsics, and \cref{tab:supp_cameras_mean} the main camera table with means instead of medians over the scenes. Jojo4D quickly saturates the pair metrics in \cref{tab:supp_camera_metrics}. Its RRA$_1$ reaches $99.7\%$ to $100\%$ on every dataset except MultiCamRobolab and DexYCB.

On the synthetic datasets, some baselines have a lower principal-point error (\cref{tab:supp_intrinsics}), because they assume the principal point at the image center, which is exact for these renders. On the real datasets, our field of view is less accurate than that of the best feed-forward methods.

The means in \cref{tab:supp_cameras_mean} are dominated by a few four-camera Perpetua clips where the camera groups have little overlap and are hard to place relative to one another. To evaluate the motion within each video independently of that placement, we align each trajectory separately to the ground truth. With this alignment, Jojo4D has the lowest mean per-video ATE$_n$ on Perpetua ($0.21\%$ versus $0.49\%$ for $\pi^4$, the best baseline). \Cref{sec:supp_failures} discusses these failures and how often the baselines fail on the same clips.

\paragraph{Accuracy of the ground-truth cameras.}
\label{sec:supp_gt_accuracy}
On MultiCamRobolab, even the lowest rotation error in \cref{tab:cameras} is $0.84^\circ$, higher than on every other dataset (except DexYCB, $1.2^\circ$). To check whether the ground truth could explain part of this gap, we use the same relative pose errors (\cref{eq:rot_error,eq:trans_error}) to compare the methods' camera estimates with each other. Our cameras and those of VGGT-$\Omega$ and Pi3X differ from each other by only $0.19^\circ$ to $0.28^\circ$, but from the ground truth by $0.84^\circ$ to $0.90^\circ$.

The discrepancy is consistent with an error in the calibration between a camera and its motion capture markers, since such an error would introduce the same rotation offset for that camera at every time step. To test this explanation, we fit one such rotation per camera in each clip using only the Pi3X estimates. Applying these rotations to the ground truth roughly halves the rotation error of our method, from $0.84^\circ$ to $0.48^\circ$, and of VGGT-$\Omega$, from $0.84^\circ$ to $0.41^\circ$. On MV-Kubric, whose ground truth is exact, fitting the same correction to Pi3X increases the median rotation error of every other method tested.

\begin{figure}[t]
\centering
\includegraphics[width=\linewidth]{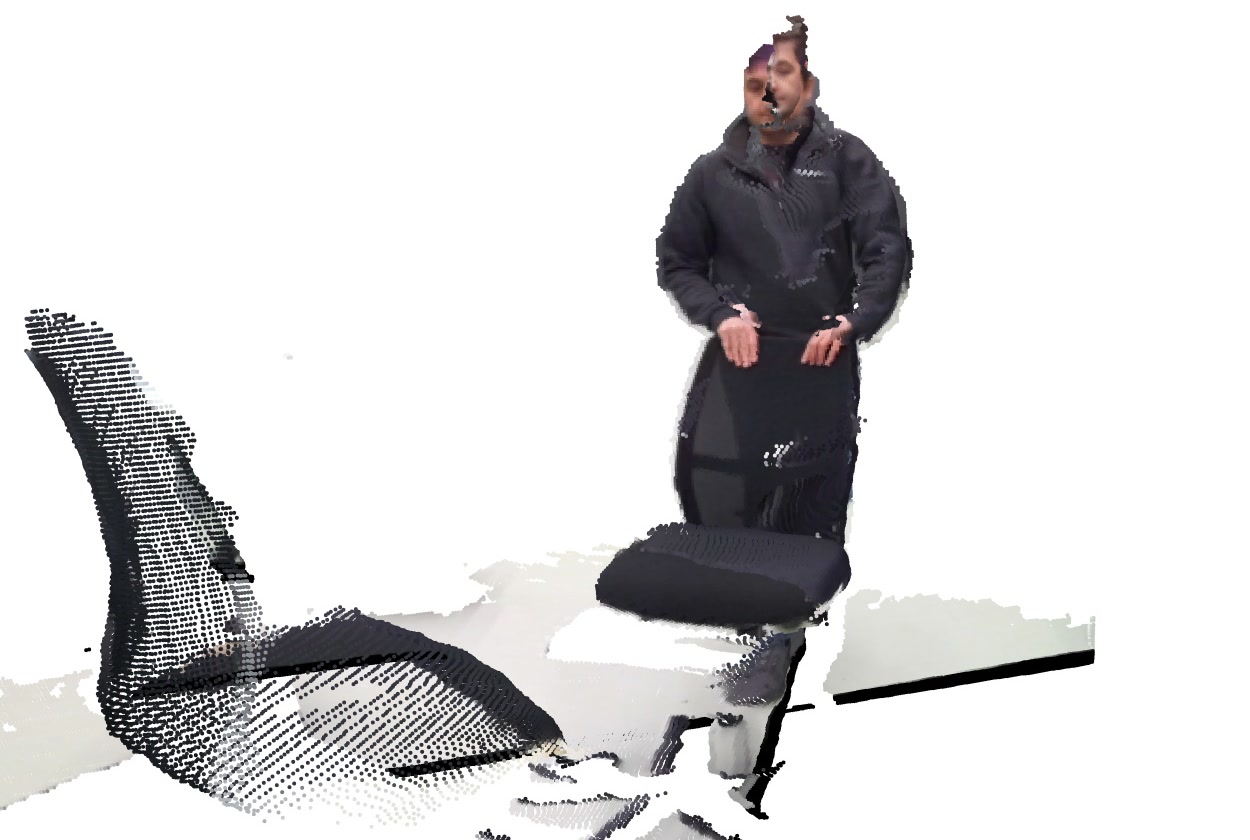}
\caption{\textbf{Duplicate surfaces in the ground truth of MultiCamRobolab.} The face of the person appears twice, shifted sideways, indicating a slight  mismatch in the ground truth poses or point data.}
\label{fig:mcr_gt_duplicate}
\end{figure}

The MultiCamRobolab ground-truth point clouds also show a mismatch between views. Both cameras observe the same surfaces at the same instant, but the ground-truth poses place them as two shifted copies (\cref{fig:mcr_gt_duplicate}).

Together, these checks suggest that part of the measured rotation error on MultiCamRobolab comes from the ground-truth camera poses. The dataset still distinguishes methods with large camera errors, but a difference below about one degree is weak evidence that one method estimates rotations more accurately than another.

\begin{table*}[p]
\caption{\textbf{Additional camera benchmark metrics.} Averaged over all datasets, Jojo4D outperforms every baseline on every metric, and it reaches near-perfect RRA$_1$ and RTA$_1$ on all datasets except MultiCamRobolab and DexYCB. ATE$_n$ is in percent of the extent of the GT cameras (lower is better). RRA$_1$ and RTA$_1$ are the rotation and translation accuracy at $1^\circ$ and AUC$_5$ is the pose AUC up to $5^\circ$, all in percent (higher is better). Each value averages the setting medians of a dataset. Dark and light green mark the best and second best results.}
\label{tab:supp_camera_metrics}
\mksetup
\resizebox{\linewidth}{!}{%
\footnotesize\setlength{\tabcolsep}{2.5pt}
\begin{tabular}{l@{\hskip5pt}cccc@{\hskip2.5pt{\color{gray!40}\vrule width 0.4pt}\hskip2.5pt}cccc@{\hskip2.5pt{\color{gray!40}\vrule width 0.4pt}\hskip2.5pt}cccc@{\hskip2.5pt{\color{gray!40}\vrule width 0.4pt}\hskip2.5pt}cccc@{\hskip2.5pt{\color{gray!40}\vrule width 0.4pt}\hskip2.5pt}cccc@{\hskip2.5pt{\color{gray!40}\vrule width 0.4pt}\hskip2.5pt}cccc@{\hskip2.5pt{\color{gray!40}\vrule width 0.4pt}\hskip2.5pt}cccc@{\hskip2.5pt{\color{gray!40}\vrule width 0.4pt}\hskip2.5pt}cccc}
\toprule
 & \multicolumn{4}{@{}c@{\hskip2.5pt{\color{gray!40}\vrule width 0.4pt}\hskip2.5pt}}{\cellcolor{mkhi4d}\textbf{Hi4D}} & \multicolumn{4}{@{}c@{\hskip2.5pt{\color{gray!40}\vrule width 0.4pt}\hskip2.5pt}}{\cellcolor{mkperp}\textbf{Perpetua}} & \multicolumn{4}{@{}c@{\hskip2.5pt{\color{gray!40}\vrule width 0.4pt}\hskip2.5pt}}{\cellcolor{mksyn}\textbf{Syn4D}} & \multicolumn{4}{@{}c@{\hskip2.5pt{\color{gray!40}\vrule width 0.4pt}\hskip2.5pt}}{\cellcolor{mkcharge}\textbf{CHARGE}} & \multicolumn{4}{@{}c@{\hskip2.5pt{\color{gray!40}\vrule width 0.4pt}\hskip2.5pt}}{\cellcolor{mkmcr}\textbf{MCRL}} & \multicolumn{4}{@{}c@{\hskip2.5pt{\color{gray!40}\vrule width 0.4pt}\hskip2.5pt}}{\cellcolor{mkkub}\textbf{MV-Kubric}} & \multicolumn{4}{@{}c@{\hskip2.5pt{\color{gray!40}\vrule width 0.4pt}\hskip2.5pt}}{\cellcolor{mkdex}\textbf{DexYCB}} & \multicolumn{4}{@{}c}{\cellcolor{mkavg}\textbf{Avg.}} \\
Method & \cellcolor{mkhi4dl}ATE$_n$ & \cellcolor{mkhi4dl}RRA$_1$ & \cellcolor{mkhi4dl}RTA$_1$ & \cellcolor{mkhi4dl}AUC$_5$ & \cellcolor{mkperpl}ATE$_n$ & \cellcolor{mkperpl}RRA$_1$ & \cellcolor{mkperpl}RTA$_1$ & \cellcolor{mkperpl}AUC$_5$ & \cellcolor{mksynl}ATE$_n$ & \cellcolor{mksynl}RRA$_1$ & \cellcolor{mksynl}RTA$_1$ & \cellcolor{mksynl}AUC$_5$ & \cellcolor{mkchargel}ATE$_n$ & \cellcolor{mkchargel}RRA$_1$ & \cellcolor{mkchargel}RTA$_1$ & \cellcolor{mkchargel}AUC$_5$ & \cellcolor{mkmcrl}ATE$_n$ & \cellcolor{mkmcrl}RRA$_1$ & \cellcolor{mkmcrl}RTA$_1$ & \cellcolor{mkmcrl}AUC$_5$ & \cellcolor{mkkubl}ATE$_n$ & \cellcolor{mkkubl}RRA$_1$ & \cellcolor{mkkubl}RTA$_1$ & \cellcolor{mkkubl}AUC$_5$ & \cellcolor{mkdexl}ATE$_n$ & \cellcolor{mkdexl}RRA$_1$ & \cellcolor{mkdexl}RTA$_1$ & \cellcolor{mkdexl}AUC$_5$ & \cellcolor{mkavgl}ATE$_n$ & \cellcolor{mkavgl}RRA$_1$ & \cellcolor{mkavgl}RTA$_1$ & \cellcolor{mkavgl}AUC$_5$ \\
\midrule
\multicolumn{33}{l}{\textit{Feed-forward}} \\
VGGT~\cite{wang2025vggt} & 1.4 & 22.5 & 19.2 & 54.7 & 4.6 & 20.9 & 12.9 & 26.5 & 1.0 & 38.0 & 28.9 & 62.4 & 2.4 & 47.1 & 1.8 & 25.1 & .63 & 52.5 & 16.2 & 55.3 & 6.3 & 5.0 & 11.8 & 28.2 & 1.9 & 5.6 & 16.0 & 39.5 & 2.6 & 27.4 & 15.3 & 41.7 \\
VGGT-$\Omega$~\cite{wang2026vggtomega} & .28 & \mkS{91.6} & 56.8 & 88.8 & .45 & 85.9 & 74.5 & 86.8 & .38 & 84.6 & 75.5 & 87.1 & .96 & 67.4 & 17.0 & 52.6 & \mkB{.34} & \mkB{55.7} & \mkB{27.4} & \mkB{73.6} & .54 & \mkS{71.5} & 77.1 & \mkS{90.7} & .47 & 22.4 & \mkB{56.5} & 77.3 & .49 & 68.4 & 55.0 & 79.6 \\
Pi3X~\cite{wang2026pi3} & .43 & 72.5 & 51.9 & 83.8 & 1.1 & 56.6 & 42.6 & 69.3 & .49 & 83.6 & 60.6 & 80.7 & 1.6 & 66.5 & 7.6 & 37.6 & .52 & 53.1 & 15.9 & 63.9 & 5.9 & 0.3 & 2.0 & 4.7 & .73 & 24.3 & 31.7 & 72.2 & 1.5 & 51.0 & 30.3 & 58.9 \\
$\pi^4$~\cite{wang2026pi4} & \mkS{.22} & 88.0 & \mkS{69.3} & \mkS{91.0} & .41 & 89.4 & 71.9 & 88.9 & .15 & \mkB{100} & 87.3 & 94.9 & .73 & 70.5 & 30.1 & 67.4 & .40 & 52.2 & \mkS{22.6} & 70.5 & \mkS{.17} & \mkB{100} & \mkS{99.9} & \mkB{100} & .49 & \mkB{34.5} & 49.3 & \mkB{81.4} & .37 & 76.4 & 61.5 & 84.9 \\
OmniX~\cite{jiang2026omnix} & .41 & 27.3 & 34.7 & 65.4 & 1.1 & 48.4 & 41.2 & 64.3 & .39 & 73.2 & 70.4 & 82.3 & 1.1 & 64.8 & 12.1 & 47.4 & .45 & 50.4 & 20.9 & 68.3 & .78 & 24.7 & 43.6 & 72.0 & .72 & 12.3 & 26.8 & 62.0 & .71 & 43.0 & 35.7 & 66.0 \\
ARROW-FF~\cite{fradlin2026arrow} & .28 & 59.8 & 52.2 & 84.7 & .59 & 68.2 & 60.0 & 80.3 & .24 & \mkS{94.8} & 84.8 & 92.4 & \mkS{.66} & \mkS{84.2} & 41.2 & \mkS{74.1} & .40 & 54.0 & 21.1 & 68.0 & .70 & 25.9 & 45.9 & 74.5 & .67 & 10.4 & 36.2 & 67.4 & .51 & 56.8 & 48.8 & 77.3 \\
\midrule
\multicolumn{33}{l}{\textit{Monocular, per video}} \\
Any4D~\cite{karhade2026any4d} & 1.4 & 23.4 & 22.1 & 49.1 & 3.1 & 16.8 & 10.9 & 33.1 & 1.1 & 60.7 & 46.6 & 63.7 & 6.0 & 42.2 & 16.3 & 29.8 & 2.5 & 34.3 & 3.0 & 27.3 & .59 & 48.9 & 62.2 & 84.0 & \mkS{.43} & 19.6 & 51.8 & 77.9 & 2.2 & 35.1 & 30.4 & 52.1 \\
V-DPM~\cite{sucar2026vdpm} & .98 & 51.2 & 29.6 & 68.3 & 1.8 & 20.6 & 16.3 & 44.2 & 1.4 & 54.7 & 38.0 & 59.1 & 7.5 & 28.6 & 12.7 & 30.4 & 2.1 & 48.5 & 6.5 & 30.2 & .58 & 48.9 & 62.2 & 84.0 & \mkS{.43} & 19.6 & \mkS{53.6} & 77.9 & 2.1 & 38.9 & 31.3 & 56.3 \\
MegaSaM~\cite{li2025megasam} & .33 & 82.8 & 55.2 & 87.3 & .96 & 86.5 & 69.4 & 85.7 & \mkS{.13} & \mkB{100} & \mkS{95.6} & \mkS{98.2} & 1.1 & 65.3 & 38.4 & 70.0 & .55 & 53.0 & 20.0 & \mkS{70.9} & .60 & 48.9 & 61.1 & 84.0 & .44 & 19.6 & 50.0 & 77.1 & .59 & 65.2 & 55.7 & 81.9 \\
\midrule
\multicolumn{33}{l}{\textit{Multi-view optimization}} \\
Sun et al.~\cite{sun2026multiview} & 1.1 & 41.2 & 33.7 & 69.8 & 6.3 & 23.4 & 11.1 & 28.9 & .69 & 43.7 & 43.0 & 71.6 & 1.5 & 48.1 & 9.6 & 47.4 & .55 & 51.1 & 16.0 & 64.1 & 1.0 & 21.0 & 27.2 & 70.1 & .60 & 19.2 & 40.8 & 77.0 & 1.7 & 35.4 & 25.9 & 61.3 \\
Jojo4D (init only) & .26 & 78.7 & 61.0 & 87.7 & .44 & 84.4 & 75.8 & 91.4 & .15 & \mkB{100} & 91.0 & 96.7 & .92 & \mkS{84.2} & 38.5 & 66.6 & .51 & \mkS{55.4} & 18.2 & 69.0 & .59 & 48.9 & 61.1 & 84.0 & \mkS{.43} & 19.6 & 51.8 & 77.9 & .47 & 67.3 & 56.8 & 81.9 \\
Jojo4D (Pi3X) & \mkB{.003} & \mkB{100} & \mkB{100} & \mkB{100} & \mkS{.11} & 99.3 & \mkS{97.4} & \mkS{98.8} & \mkB{.01} & \mkB{100} & \mkB{99.9} & \mkB{100} & \mkB{.02} & \mkB{100} & \mkS{99.4} & \mkB{99.8} & \mkS{.37} & 53.8 & 21.8 & 70.2 & \mkB{.03} & \mkB{100} & \mkB{100} & \mkB{100} & .45 & \mkS{26.8} & 48.2 & 76.8 & \mkS{.14} & \mkB{82.8} & \mkS{81.0} & \mkS{92.2} \\
Jojo4D ($\pi^4$) & \mkB{.003} & \mkB{100} & \mkB{100} & \mkB{100} & .13 & \mkB{99.8} & 96.6 & 98.7 & \mkB{.01} & \mkB{100} & \mkB{99.9} & \mkB{100} & \mkB{.02} & \mkB{100} & \mkB{99.5} & \mkB{99.8} & \mkS{.37} & 53.4 & 21.7 & 70.1 & \mkB{.03} & \mkB{100} & \mkB{100} & \mkB{100} & \mkB{.42} & 17.9 & 48.2 & 74.3 & \mkS{.14} & 81.6 & 80.8 & 91.8 \\
\textbf{Jojo4D (ours)} & \mkB{.003} & \mkB{100} & \mkB{100} & \mkB{100} & \mkB{.08} & \mkS{99.7} & \mkB{98.3} & \mkB{99.2} & \mkB{.01} & \mkB{100} & \mkB{99.9} & \mkB{100} & \mkB{.02} & \mkB{100} & \mkB{99.5} & \mkB{99.8} & \mkS{.37} & 53.2 & 21.6 & 70.1 & \mkB{.03} & \mkB{100} & \mkB{100} & \mkB{100} & .44 & 23.2 & 51.8 & \mkS{78.6} & \mkB{.13} & \mkS{82.3} & \mkB{81.6} & \mkB{92.5} \\
\bottomrule
\end{tabular}}
\vspace{4mm}
\caption{\textbf{Camera intrinsics per dataset.} Jojo4D has the lowest average error in focal length, field of view and principal point, and recovers the intrinsics almost exactly on all datasets except MultiCamRobolab and DexYCB. On DexYCB, the static cameras mostly observe a small tabletop scene, which likely constrains the intrinsics only weakly. With ground-truth intrinsics, our rotation error on DexYCB halves (\cref{sec:supp_calibrated}). $f$ is the relative focal length error in percent, FoV the field-of-view error in degrees and pp the principal-point distance in percent of the image diagonal (lower is better). For each dataset, we take the median over the clips of each camera setting and average these medians. Avg.\ equals the $f$ and FoV columns of \cref{tab:cameras}. Sun~\etal use the intrinsics of their Pi3X initialization.}
\label{tab:supp_intrinsics}
\mksetup
\centering\resizebox{0.86\linewidth}{!}{%
\footnotesize\setlength{\tabcolsep}{2.5pt}
\begin{tabular}{l@{\hskip5pt}ccc@{\hskip2.5pt{\color{gray!40}\vrule width 0.4pt}\hskip2.5pt}ccc@{\hskip2.5pt{\color{gray!40}\vrule width 0.4pt}\hskip2.5pt}ccc@{\hskip2.5pt{\color{gray!40}\vrule width 0.4pt}\hskip2.5pt}ccc@{\hskip2.5pt{\color{gray!40}\vrule width 0.4pt}\hskip2.5pt}ccc@{\hskip2.5pt{\color{gray!40}\vrule width 0.4pt}\hskip2.5pt}w{c}{2.1em}w{c}{2.1em}w{c}{2.1em}@{\hskip2.5pt{\color{gray!40}\vrule width 0.4pt}\hskip2.5pt}ccc@{\hskip2.5pt{\color{gray!40}\vrule width 0.4pt}\hskip2.5pt}ccc}
\toprule
 & \multicolumn{3}{@{}c@{\hskip2.5pt{\color{gray!40}\vrule width 0.4pt}\hskip2.5pt}}{\cellcolor{mkhi4d}\textbf{Hi4D}} & \multicolumn{3}{@{}c@{\hskip2.5pt{\color{gray!40}\vrule width 0.4pt}\hskip2.5pt}}{\cellcolor{mkperp}\textbf{Perpetua}} & \multicolumn{3}{@{}c@{\hskip2.5pt{\color{gray!40}\vrule width 0.4pt}\hskip2.5pt}}{\cellcolor{mksyn}\textbf{Syn4D}} & \multicolumn{3}{@{}c@{\hskip2.5pt{\color{gray!40}\vrule width 0.4pt}\hskip2.5pt}}{\cellcolor{mkcharge}\textbf{CHARGE}} & \multicolumn{3}{@{}c@{\hskip2.5pt{\color{gray!40}\vrule width 0.4pt}\hskip2.5pt}}{\cellcolor{mkmcr}\textbf{MCRL}} & \multicolumn{3}{@{}c@{\hskip2.5pt{\color{gray!40}\vrule width 0.4pt}\hskip2.5pt}}{\cellcolor{mkkub}\textbf{MV-Kubric}} & \multicolumn{3}{@{}c@{\hskip2.5pt{\color{gray!40}\vrule width 0.4pt}\hskip2.5pt}}{\cellcolor{mkdex}\textbf{DexYCB}} & \multicolumn{3}{@{}c}{\cellcolor{mkavg}\textbf{Avg.}} \\
Method & \cellcolor{mkhi4dl}$f\!\downarrow$ & \cellcolor{mkhi4dl}FoV$\downarrow$ & \cellcolor{mkhi4dl}pp$\downarrow$ & \cellcolor{mkperpl}$f\!\downarrow$ & \cellcolor{mkperpl}FoV$\downarrow$ & \cellcolor{mkperpl}pp$\downarrow$ & \cellcolor{mksynl}$f\!\downarrow$ & \cellcolor{mksynl}FoV$\downarrow$ & \cellcolor{mksynl}pp$\downarrow$ & \cellcolor{mkchargel}$f\!\downarrow$ & \cellcolor{mkchargel}FoV$\downarrow$ & \cellcolor{mkchargel}pp$\downarrow$ & \cellcolor{mkmcrl}$f\!\downarrow$ & \cellcolor{mkmcrl}FoV$\downarrow$ & \cellcolor{mkmcrl}pp$\downarrow$ & \cellcolor{mkkubl}$f\!\downarrow$ & \cellcolor{mkkubl}FoV$\downarrow$ & \cellcolor{mkkubl}pp$\downarrow$ & \cellcolor{mkdexl}$f\!\downarrow$ & \cellcolor{mkdexl}FoV$\downarrow$ & \cellcolor{mkdexl}pp$\downarrow$ & \cellcolor{mkavgl}$f\!\downarrow$ & \cellcolor{mkavgl}FoV$\downarrow$ & \cellcolor{mkavgl}pp$\downarrow$ \\
\midrule
\multicolumn{25}{l}{\textit{Feed-forward}} \\
VGGT~\cite{wang2025vggt} & 10 & 4.3 & 1.2 & 1.9 & .68 & .10 & 3.0 & 1.3 & .12 & 5.0 & 1.6 & .12 & 1.8 & .88 & .40 & 3.7 & 1.7 & .10 & 2.9 & 1.2 & \mkS{.97} & 4.0 & 1.7 & .43 \\
VGGT-$\Omega$~\cite{wang2026vggtomega} & 2.5 & 1.1 & 1.2 & .36 & .13 & .12 & 1.1 & .53 & .11 & 1.3 & .47 & .09 & 1.2 & \mkS{.61} & .40 & 1.2 & .52 & .12 & 1.4 & \mkS{.56} & \mkS{.97} & 1.3 & .57 & .43 \\
Pi3X~\cite{wang2026pi3} & .68 & \mkS{.31} & 1.1 & 1.1 & .40 & \mkB{0} & 1.1 & .52 & \mkB{0} & 3.5 & 1.1 & \mkB{0} & 2.2 & 1.1 & .51 & 30 & 11 & \mkB{0} & 1.8 & .78 & \mkB{.96} & 5.9 & 2.1 & .37 \\
$\pi^4$~\cite{wang2026pi4} & 1.4 & .62 & 1.1 & 1.1 & .40 & \mkB{0} & .68 & .31 & \mkB{0} & 1.8 & .61 & \mkB{0} & \mkB{.71} & \mkB{.38} & .51 & \mkS{.43} & \mkS{.19} & \mkB{0} & \mkS{1.3} & \mkS{.56} & \mkB{.96} & 1.1 & .44 & .37 \\
OmniX~\cite{jiang2026omnix} & 15 & 6.5 & 1.1 & 11 & 3.7 & \mkB{0} & 1.3 & .57 & \mkB{0} & 12 & 4.0 & \mkB{0} & 4.1 & 1.9 & .51 & 10 & 4.3 & \mkB{0} & 5.5 & 2.2 & \mkB{.96} & 8.4 & 3.3 & .37 \\
ARROW-FF~\cite{fradlin2026arrow} & .70 & .32 & 1.1 & .75 & .28 & .02 & \mkS{.64} & \mkS{.26} & \mkS{.02} & 1.1 & \mkS{.38} & \mkS{.02} & \mkS{1.1} & .62 & .51 & 1.3 & .55 & .03 & \mkB{1.1} & \mkB{.48} & .98 & .96 & .41 & .38 \\
\midrule
\multicolumn{25}{l}{\textit{Monocular, per video}} \\
Any4D~\cite{karhade2026any4d} & 20 & 7.6 & 1.2 & 2.4 & .88 & .11 & 9.4 & 4.4 & .14 & 16 & 5.7 & .57 & 22 & 9.8 & .89 & 5.4 & 2.3 & .15 & 2.7 & 1.2 & \mkS{.97} & 11 & 4.6 & .57 \\
V-DPM~\cite{sucar2026vdpm} & 17 & 6.6 & 1.2 & 5.4 & 2.1 & .10 & 5.9 & 2.6 & .12 & 13 & 4.5 & .12 & 4.7 & 2.3 & .40 & 5.8 & 2.5 & .10 & 3.5 & 1.5 & \mkS{.97} & 7.8 & 3.2 & .43 \\
MegaSaM~\cite{li2025megasam} & 6.7 & 3.6 & 1.2 & .21 & \mkS{.08} & .11 & 7.9 & 4.3 & .10 & 11 & 4.4 & .10 & 1.7 & .88 & .40 & 24 & 13 & .11 & 12 & 5.3 & \mkS{.97} & 9.0 & 4.5 & .43 \\
\midrule
\multicolumn{25}{l}{\textit{Multi-view optimization}} \\
Sun et al.~\cite{sun2026multiview} & .77 & .35 & 1.1 & .78 & .29 & 0 & 1.5 & .72 & 0 & 3.2 & 1.0 & 0 & 2.2 & 1.1 & .51 & 30 & 11 & 0 & 1.8 & .78 & .96 & 5.8 & 2.1 & .37 \\
Jojo4D (init only) & 2.7 & 1.2 & 1.1 & .77 & .28 & \mkS{.005} & 1.0 & .45 & .07 & 4.1 & 1.4 & .09 & 1.7 & .89 & .51 & 5.5 & 2.5 & \mkS{.01} & 4.9 & 2.0 & \mkS{.97} & 3.0 & 1.2 & .39 \\
Jojo4D (Pi3X) & \mkB{.01} & \mkB{.01} & \mkS{.02} & \mkB{.05} & \mkB{.02} & .07 & \mkB{.05} & \mkB{.02} & .05 & \mkS{.06} & \mkB{.02} & .03 & 1.4 & .75 & \mkS{.25} & \mkB{.09} & \mkB{.04} & .10 & 2.7 & 1.2 & 1.4 & \mkS{.63} & \mkS{.29} & \mkS{.27} \\
Jojo4D ($\pi^4$) & \mkS{.02} & \mkB{.01} & \mkB{.01} & \mkB{.05} & \mkB{.02} & .07 & \mkB{.05} & \mkB{.02} & .05 & \mkB{.05} & \mkB{.02} & .04 & 1.4 & .74 & \mkS{.25} & \mkB{.09} & \mkB{.04} & .10 & 2.5 & 1.1 & 1.3 & \mkB{.60} & \mkB{.28} & \mkB{.26} \\
\textbf{Jojo4D (ours)} & \mkB{.01} & \mkB{.01} & \mkB{.01} & \mkS{.06} & \mkB{.02} & .06 & \mkB{.05} & \mkB{.02} & .05 & \mkB{.05} & \mkB{.02} & .03 & 1.4 & .75 & \mkB{.24} & \mkB{.09} & \mkB{.04} & .09 & 2.7 & 1.2 & 1.4 & \mkS{.63} & \mkS{.29} & \mkS{.27} \\
\bottomrule
\end{tabular}}
\vspace{4mm}
{\renewenvironment{table*}[1][]{}{}
\begin{table*}[t]
\caption{\textbf{Camera estimation, mean over scenes.} With the mean instead of the median, Jojo4D still has the lowest average errors of all methods. Its high mean errors on Perpetua are dominated by a few clips with limited view overlap between cameras (\cref{sec:supp_failures}). We also show Jojo4D initialized from Pi3X and $\pi^4$. Avg.\ first averages settings within each dataset, then averages over all seven datasets. Dark and light green mark the best and second best results.}
\label{tab:supp_cameras_mean}
\mksetup
\resizebox{\linewidth}{!}{%
\footnotesize\setlength{\tabcolsep}{2.5pt}
\begin{tabular}{l@{\hskip5pt}cccccc@{\hskip2.5pt{\color{gray!40}\vrule width 0.4pt}\hskip2.5pt}cccc@{\hskip2.5pt{\color{gray!40}\vrule width 0.4pt}\hskip2.5pt}cccccc@{\hskip2.5pt{\color{gray!40}\vrule width 0.4pt}\hskip2.5pt}cccccc@{\hskip2.5pt{\color{gray!40}\vrule width 0.4pt}\hskip2.5pt}cc@{\hskip2.5pt{\color{gray!40}\vrule width 0.4pt}\hskip2.5pt}w{c}{2.1em}w{c}{2.1em}@{\hskip2.5pt{\color{gray!40}\vrule width 0.4pt}\hskip2.5pt}cc@{\hskip2.5pt{\color{gray!40}\vrule width 0.4pt}\hskip2.5pt}ccccc}
\toprule
 & \multicolumn{6}{@{}c@{\hskip2.5pt{\color{gray!40}\vrule width 0.4pt}\hskip2.5pt}}{\cellcolor{mkhi4d}\textbf{Hi4D}} & \multicolumn{4}{@{}c@{\hskip2.5pt{\color{gray!40}\vrule width 0.4pt}\hskip2.5pt}}{\cellcolor{mkperp}\textbf{Perpetua}} & \multicolumn{6}{@{}c@{\hskip2.5pt{\color{gray!40}\vrule width 0.4pt}\hskip2.5pt}}{\cellcolor{mksyn}\textbf{Syn4D}} & \multicolumn{6}{@{}c@{\hskip2.5pt{\color{gray!40}\vrule width 0.4pt}\hskip2.5pt}}{\cellcolor{mkcharge}\textbf{CHARGE}} & \multicolumn{2}{@{}c@{\hskip2.5pt{\color{gray!40}\vrule width 0.4pt}\hskip2.5pt}}{\cellcolor{mkmcr}\textbf{MCRL}} & \multicolumn{2}{@{}c@{\hskip2.5pt{\color{gray!40}\vrule width 0.4pt}\hskip2.5pt}}{\cellcolor{mkkub}\textbf{MV-Kubric}} & \multicolumn{2}{@{}c@{\hskip2.5pt{\color{gray!40}\vrule width 0.4pt}\hskip2.5pt}}{\cellcolor{mkdex}\textbf{DexYCB}} & \multicolumn{5}{@{}c}{\cellcolor{mkavg}\textbf{Avg.}} \\
 & \multicolumn{2}{@{}c}{\cellcolor{mkhi4dl}2 cam} & \multicolumn{2}{c}{\cellcolor{mkhi4dl}4 cam} & \multicolumn{2}{c@{\hskip2.5pt{\color{gray!40}\vrule width 0.4pt}\hskip2.5pt}}{\cellcolor{mkhi4dl}8 cam} & \multicolumn{2}{@{}c}{\cellcolor{mkperpl}2 cam} & \multicolumn{2}{c@{\hskip2.5pt{\color{gray!40}\vrule width 0.4pt}\hskip2.5pt}}{\cellcolor{mkperpl}4 cam} & \multicolumn{2}{@{}c}{\cellcolor{mksynl}2 cam} & \multicolumn{2}{c}{\cellcolor{mksynl}4 cam} & \multicolumn{2}{c@{\hskip2.5pt{\color{gray!40}\vrule width 0.4pt}\hskip2.5pt}}{\cellcolor{mksynl}8 cam} & \multicolumn{2}{@{}c}{\cellcolor{mkchargel}2 cam} & \multicolumn{2}{c}{\cellcolor{mkchargel}4 cam} & \multicolumn{2}{c@{\hskip2.5pt{\color{gray!40}\vrule width 0.4pt}\hskip2.5pt}}{\cellcolor{mkchargel}8 cam} & \multicolumn{2}{@{}c@{\hskip2.5pt{\color{gray!40}\vrule width 0.4pt}\hskip2.5pt}}{\cellcolor{mkmcrl}2 cam} & \multicolumn{2}{@{}c@{\hskip2.5pt{\color{gray!40}\vrule width 0.4pt}\hskip2.5pt}}{\cellcolor{mkkubl}10 cam} & \multicolumn{2}{@{}c@{\hskip2.5pt{\color{gray!40}\vrule width 0.4pt}\hskip2.5pt}}{\cellcolor{mkdexl}8 cam} & \multicolumn{5}{@{}c}{\cellcolor{mkavgl}} \\
Method & \cellcolor{mkhi4dl}$R\!\downarrow$ & \cellcolor{mkhi4dl}$t\!\downarrow$ & \cellcolor{mkhi4dl}$R\!\downarrow$ & \cellcolor{mkhi4dl}$t\!\downarrow$ & \cellcolor{mkhi4dl}$R\!\downarrow$ & \cellcolor{mkhi4dl}$t\!\downarrow$ & \cellcolor{mkperpl}$R\!\downarrow$ & \cellcolor{mkperpl}$t\!\downarrow$ & \cellcolor{mkperpl}$R\!\downarrow$ & \cellcolor{mkperpl}$t\!\downarrow$ & \cellcolor{mksynl}$R\!\downarrow$ & \cellcolor{mksynl}$t\!\downarrow$ & \cellcolor{mksynl}$R\!\downarrow$ & \cellcolor{mksynl}$t\!\downarrow$ & \cellcolor{mksynl}$R\!\downarrow$ & \cellcolor{mksynl}$t\!\downarrow$ & \cellcolor{mkchargel}$R\!\downarrow$ & \cellcolor{mkchargel}$t\!\downarrow$ & \cellcolor{mkchargel}$R\!\downarrow$ & \cellcolor{mkchargel}$t\!\downarrow$ & \cellcolor{mkchargel}$R\!\downarrow$ & \cellcolor{mkchargel}$t\!\downarrow$ & \cellcolor{mkmcrl}$R\!\downarrow$ & \cellcolor{mkmcrl}$t\!\downarrow$ & \cellcolor{mkkubl}$R\!\downarrow$ & \cellcolor{mkkubl}$t\!\downarrow$ & \cellcolor{mkdexl}$R\!\downarrow$ & \cellcolor{mkdexl}$t\!\downarrow$ & \cellcolor{mkavgl}$R\!\downarrow$ & \cellcolor{mkavgl}$t\!\downarrow$ & \cellcolor{mkavgl}ATE$_n\!\downarrow$ & \cellcolor{mkavgl}$f\!\downarrow$ & \cellcolor{mkavgl}FoV$\downarrow$ \\
\midrule
\multicolumn{34}{l}{\textit{Feed-forward}} \\
VGGT~\cite{wang2025vggt} & 3.6 & 3.6 & 7.4 & 7.5 & 5.2 & 5.4 & 5.3 & 6.3 & 21 & 17 & 3.1 & 6.3 & 3.6 & 6.0 & 5.5 & 7.7 & .48 & 9.7 & 4.5 & 15 & 2.5 & 8.6 & 1.1 & 5.7 & 15 & 11 & 2.6 & 3.1 & 6.3 & 7.8 & 4.3 & 5.2 & 2.2 \\
VGGT-$\Omega$~\cite{wang2026vggtomega} & .60 & 1.2 & .76 & 1.2 & .76 & 1.1 & \mkB{.68} & 1.3 & 5.1 & 4.1 & .68 & 1.5 & 1.1 & 1.8 & 1.0 & 1.6 & .25 & 3.8 & 8.8 & 11 & 4.1 & 6.3 & \mkS{.85} & \mkB{2.4} & 1.0 & .88 & \mkS{1.5} & \mkB{1.0} & 1.8 & 2.4 & .95 & 1.7 & .73 \\
Pi3X~\cite{wang2026pi3} & .61 & 1.4 & 1.2 & 1.5 & 1.7 & 2.2 & 1.1 & 2.5 & 5.7 & 6.6 & .65 & 2.5 & .79 & 1.8 & 1.0 & 1.9 & .54 & 7.0 & 8.8 & 11 & 7.9 & 10 & .89 & 3.5 & 8.8 & 8.8 & \mkS{1.5} & 1.5 & 3.2 & 4.5 & 2.1 & 6.2 & 2.3 \\
$\pi^4$~\cite{wang2026pi4} & \mkS{.54} & .89 & .69 & .94 & .71 & 1.1 & \mkS{.71} & 1.3 & \mkB{2.8} & \mkS{2.5} & 2.3 & 1.2 & \mkS{.36} & .75 & .33 & .53 & .43 & 3.0 & \mkS{.89} & 2.3 & .95 & 2.5 & \mkS{.85} & 2.9 & .32 & .26 & \mkB{1.3} & \mkS{1.1} & .94 & 1.5 & .51 & 1.2 & .50 \\
OmniX~\cite{jiang2026omnix} & 1.5 & 1.8 & 1.9 & 1.8 & 2.9 & 2.4 & 1.5 & 2.0 & 10 & 7.1 & .76 & 1.4 & .71 & 1.4 & 2.3 & 1.9 & .42 & 3.6 & 1.4 & 18 & 1.3 & 3.9 & .99 & 2.9 & 1.7 & 1.6 & 2.1 & 1.6 & 2.1 & 3.3 & 1.0 & 8.6 & 3.4 \\
ARROW-FF~\cite{fradlin2026arrow} & .90 & 1.2 & .98 & 1.1 & 1.3 & 1.4 & .98 & 1.6 & 5.0 & 3.8 & .49 & .96 & .53 & .87 & .69 & .85 & .39 & 2.6 & .99 & 1.9 & .81 & \mkS{1.9} & .87 & 3.0 & 1.6 & 1.4 & 1.9 & 1.4 & 1.4 & 1.8 & .69 & 1.3 & .52 \\
\midrule
\multicolumn{34}{l}{\textit{Monocular, per video}} \\
Any4D~\cite{karhade2026any4d} & 2.8 & 4.6 & 2.7 & 3.3 & 2.4 & 2.9 & 5.1 & 7.8 & 9.9 & 10 & 1.3 & 5.2 & 1.1 & 4.1 & 1.1 & 2.4 & 2.1 & 29 & 1.8 & 12 & 1.5 & 12 & 1.4 & 11 & 1.2 & .99 & \mkS{1.5} & \mkB{1.0} & 2.5 & 6.8 & 2.9 & 12 & 4.8 \\
V-DPM~\cite{sucar2026vdpm} & 1.2 & 3.6 & 1.3 & 2.8 & 1.2 & 2.3 & 3.1 & 5.0 & 8.9 & 8.6 & 1.4 & 7.2 & 1.2 & 5.4 & 1.1 & 3.2 & 2.0 & 33 & 1.7 & 13 & 1.5 & 13 & 1.2 & 9.6 & 1.2 & .98 & \mkS{1.5} & \mkB{1.0} & 2.0 & 6.6 & 2.6 & 7.9 & 3.3 \\
MegaSaM~\cite{li2025megasam} & .71 & 1.3 & 1.3 & 1.7 & 1.1 & 1.5 & 4.7 & 4.3 & 8.2 & 5.8 & \mkS{.37} & .38 & .46 & .65 & .48 & .49 & 1.5 & 5.8 & 1.6 & 3.3 & 1.1 & 3.3 & .86 & 2.6 & 1.2 & .99 & \mkS{1.5} & \mkB{1.0} & 1.8 & 2.3 & 1.3 & 9.2 & 4.6 \\
\midrule
\multicolumn{34}{l}{\textit{Multi-view optimization}} \\
Sun et al.~\cite{sun2026multiview} & 3.3 & 4.2 & 1.4 & 2.0 & 3.1 & 2.7 & 20 & 11 & 22 & 16 & 3.8 & 3.7 & 2.0 & 2.1 & 4.4 & 3.0 & 15 & 13 & 18 & 11 & 11 & 8.9 & 1.0 & 3.4 & 2.7 & 2.0 & \mkS{1.5} & 1.3 & 6.7 & 5.3 & 3.5 & 6.3 & 2.3 \\
Jojo4D (init only) & .77 & 1.1 & .78 & 1.0 & .85 & 1.1 & .99 & 1.2 & 5.3 & 4.3 & .42 & .61 & .50 & .80 & .51 & .56 & .60 & 4.5 & .91 & 2.1 & .73 & 2.1 & .86 & 3.1 & 1.2 & .99 & \mkS{1.5} & \mkB{1.0} & 1.3 & 1.8 & .72 & 3.1 & 1.3 \\
Jojo4D (Pi3X) & \mkB{.03} & \mkS{.03} & \mkS{.04} & \mkS{.04} & \mkS{.05} & .03 & 1.1 & .95 & 6.7 & 2.8 & \mkB{.07} & \mkS{.08} & \mkB{.04} & \mkB{.04} & .06 & \mkS{.04} & .11 & .74 & 10 & 5.5 & 5.3 & 4.0 & \mkB{.83} & \mkS{2.5} & \mkS{.09} & \mkS{.07} & 1.7 & \mkS{1.1} & 1.7 & 1.3 & .75 & \mkS{.69} & \mkS{.31} \\
Jojo4D ($\pi^4$) & \mkB{.03} & \mkS{.03} & .09 & .06 & \mkB{.02} & \mkS{.02} & .90 & \mkS{.72} & \mkS{4.1} & \mkB{1.7} & \mkB{.07} & \mkB{.07} & \mkB{.04} & \mkS{.05} & \mkS{.04} & \mkB{.03} & \mkS{.08} & \mkS{.42} & \mkB{.08} & \mkB{.12} & \mkB{.02} & \mkB{.07} & \mkB{.83} & \mkS{2.5} & \mkS{.09} & \mkS{.07} & 1.6 & \mkS{1.1} & \mkB{.75} & \mkB{.74} & \mkB{.30} & \mkB{.64} & \mkB{.29} \\
\textbf{Jojo4D (ours)} & \mkB{.03} & \mkB{.02} & \mkB{.03} & \mkB{.03} & \mkB{.02} & \mkB{.01} & .75 & \mkB{.56} & 5.4 & 3.5 & \mkB{.07} & \mkB{.07} & \mkB{.04} & \mkB{.04} & \mkB{.03} & \mkB{.03} & \mkB{.03} & \mkB{.29} & \mkB{.08} & \mkS{.16} & \mkS{.03} & \mkB{.07} & \mkB{.83} & \mkS{2.5} & \mkB{.08} & \mkB{.06} & \mkS{1.5} & \mkB{1.0} & \mkS{.80} & \mkS{.84} & \mkS{.36} & \mkB{.64} & \mkB{.29} \\
\bottomrule
\end{tabular}}
\vspace{-0.3cm}
\end{table*}
}
\end{table*}

\subsection{Point cloud reconstruction}
\label{sec:supp_points}

\paragraph{Point cloud alignment.}
To align a prediction with the ground-truth point clouds, we estimate a single similarity transform for the entire sequence. We initialize this transform with a robust least-squares fit~\cite{umeyama1991least} between sampled predicted and ground-truth points at corresponding image pixels. We then refine the transform with several iterations of trimmed ICP~\cite{besl1992icp,chetverikov2002trimmed} to obtain the best fit between the point clouds. Each of the five iterations matches sampled predicted points from every fifth time step to their nearest ground-truth points of the same time step, keeps the closest $70\%$ of these pairs and estimates the transform again. We only accept the refined transform if it reduces the mean distance of the closest $70\%$ of pairs compared to the initialization and keeps the scale within a factor of two of the initial scale.

The methods predict their points on grids of different resolutions (\cref{tab:resolutions}), which changes the density of the evaluated point clouds. Before scoring, we therefore downsample Jojo4D's point maps and upsample ARROW's point maps toward the $518$-pixel-wide grid of VGGT and Pi3X. For Jojo4D, including the anchors-only variant, we select the nearest source pixel at each target pixel center without interpolating the 3D points, retaining its original grid when already close to the target resolution. For ARROW, we use depth-aware bilinear interpolation that excludes neighbors whose depths differ from the nearest source point by more than $4\%$. Other methods retain their stored grids. 

\paragraph{Metrics.}
We report the following metrics for the reconstructed point clouds. For every setting, each metric is the median over its scenes.

\emph{Accuracy, completeness and F1.} Let $\mathcal{P}_t$ be the ground-truth point cloud of time step $t$ and $\hat{\mathcal{P}}_t$ the aligned predicted point cloud. Predicted points at pixels without a valid ground-truth depth cannot be evaluated and are excluded. With $d(x, \mathcal{S})$ denoting the distance of a point $x$ to its nearest neighbor in a point set $\mathcal{S}$, the accuracy and completeness at a distance threshold $\tau$ are
\begin{align}
    \mathrm{Acc}_\tau &= \frac{\sum_t \big| \{ x \in \hat{\mathcal{P}}_t \mid d(x, \mathcal{P}_t) < \tau \} \big|}{\sum_t | \hat{\mathcal{P}}_t |}, \label{eq:accuracy} \\
    \mathrm{Comp}_\tau &= \frac{\sum_t \big| \{ y \in \mathcal{P}_t \mid d(y, \hat{\mathcal{P}}_t) < \tau \} \big|}{\sum_t | \mathcal{P}_t |}. \label{eq:completeness}
\end{align}
Nearest neighbors are computed within each time step, and the point counts are summed over all time steps of the clip before dividing. The F1 score is their harmonic mean,
\begin{equation}
    \mathrm{F1}_\tau = \frac{2 \, \mathrm{Acc}_\tau \, \mathrm{Comp}_\tau}{\mathrm{Acc}_\tau + \mathrm{Comp}_\tau}.
    \label{eq:f1}
\end{equation}
In our tables, $\mathrm{F}_2$ and $\mathrm{F}_5$ denote the F1 scores at distance thresholds of $2$ and $5$\,cm, respectively (\cref{tab:point_cloud_results}).

\emph{AUC$_\tau$.} For a method that predicts a confidence, removing the least confident points trades completeness for accuracy. Adding the points in the order of decreasing confidence traces a curve of accuracy over completeness, and we report the area under this curve,
\begin{equation}
    \mathrm{AUC}_\tau = \int_0^{\mathrm{Comp}_\tau^{\max}} \mathrm{Acc}_\tau(\kappa) \, \mathrm{d}\kappa,
    \label{eq:auc}
\end{equation}
where $\mathrm{Acc}_\tau(\kappa)$ is the accuracy at completeness $\kappa$, and $\mathrm{Comp}_\tau^{\max}$ is the completeness with all predicted points. We evaluate cuts at $0,10,\ldots,90\%$ of the confidence ranks. Tied scores stay together, so on some VGGT clips even the strictest cut retains nearly all points. To integrate from zero completeness for every method, we connect the point at the strictest evaluated confidence cut to $100\%$ accuracy at zero completeness by a straight line. The main paper reports the averages of AUC$_2$ and AUC$_5$ over all datasets as $A_2$ and $A_5$, and \cref{tab:supp_point_distances} lists them per dataset.

\emph{Chamfer distance.} For the completion ablations, we also report the clipped Chamfer distance with $\delta = 5$\,cm,
\begin{align}
    \mathrm{CD} &= \tfrac{1}{2} \big( D_\delta(\hat{\mathcal{P}}, \mathcal{P}) + D_\delta(\mathcal{P}, \hat{\mathcal{P}}) \big), \label{eq:chamfer} \\
    D_\delta(\mathcal{A}, \mathcal{B}) &= \frac{\sum_t \sum_{x \in \mathcal{A}_t} \min\big(d(x, \mathcal{B}_t), \delta\big)}{\sum_t |\mathcal{A}_t|}. \nonumber
\end{align}

\paragraph{Results.}
Complementing the point cloud results of the main paper (\cref{tab:point_cloud_results}), \cref{tab:supp_point_distances} reports the accuracy and completeness at $2$\,cm and the accuracy--completeness AUC at $2$ and $5$\,cm per dataset. Jojo4D has the highest average F1 (\cref{tab:point_cloud_results}) and AUC (\cref{tab:supp_point_distances}) at both thresholds. Most importantly, its accuracy is much higher than that of every baseline on every dataset ($86.0\%$ versus $64.0\%$ for $\pi^4$ at $2$\,cm on average).

\Cref{tab:supp_points_mean} repeats \cref{tab:point_cloud_results} using means instead of medians over scenes. Jojo4D still has the highest average F1 at both thresholds, although the clips discussed in \cref{sec:supp_failures} lower its mean scores on Perpetua.

\begin{table*}[t]
\caption{\textbf{Point cloud accuracy, completeness and AUC per dataset.} At $2$\,cm, Jojo4D and its anchors-only variant are the two most accurate methods on every dataset. Completing the anchors raises the average completeness from $68.2\%$ to $77.5\%$, the highest of all methods, while the accuracy drops from $90.8\%$ to $86.0\%$, well above the $64.0\%$ of $\pi^4$, the most accurate baseline. Jojo4D has the highest average AUC at both thresholds. Acc and Comp are the accuracy and completeness at $2$\,cm, and $\text{AUC}_2$ and $\text{AUC}_5$ are the areas under the accuracy--completeness curve at $2$ and $5$\,cm, all in percent (higher is better). Each entry averages the setting medians of one dataset, and Avg.\ averages over the seven datasets, as in \cref{tab:point_cloud_results}. MegaSaM and Sun et al.\ predict no confidence and therefore have no AUC. Dark and light green mark the best and second best results.}
\label{tab:supp_point_distances}
\mksetup
\resizebox{\linewidth}{!}{%
\footnotesize\setlength{\tabcolsep}{2.5pt}
\begin{tabular}{l@{\hskip5pt}cccc@{\hskip2.5pt{\color{gray!40}\vrule width 0.4pt}\hskip2.5pt}cccc@{\hskip2.5pt{\color{gray!40}\vrule width 0.4pt}\hskip2.5pt}cccc@{\hskip2.5pt{\color{gray!40}\vrule width 0.4pt}\hskip2.5pt}cccc@{\hskip2.5pt{\color{gray!40}\vrule width 0.4pt}\hskip2.5pt}cccc@{\hskip2.5pt{\color{gray!40}\vrule width 0.4pt}\hskip2.5pt}cccc@{\hskip2.5pt{\color{gray!40}\vrule width 0.4pt}\hskip2.5pt}cccc@{\hskip2.5pt{\color{gray!40}\vrule width 0.4pt}\hskip2.5pt}cccc}
\toprule
 & \multicolumn{4}{@{}c@{\hskip2.5pt{\color{gray!40}\vrule width 0.4pt}\hskip2.5pt}}{\cellcolor{mkhi4d}\textbf{Hi4D}} & \multicolumn{4}{@{}c@{\hskip2.5pt{\color{gray!40}\vrule width 0.4pt}\hskip2.5pt}}{\cellcolor{mkperp}\textbf{Perpetua}} & \multicolumn{4}{@{}c@{\hskip2.5pt{\color{gray!40}\vrule width 0.4pt}\hskip2.5pt}}{\cellcolor{mksyn}\textbf{Syn4D}} & \multicolumn{4}{@{}c@{\hskip2.5pt{\color{gray!40}\vrule width 0.4pt}\hskip2.5pt}}{\cellcolor{mkcharge}\textbf{CHARGE}} & \multicolumn{4}{@{}c@{\hskip2.5pt{\color{gray!40}\vrule width 0.4pt}\hskip2.5pt}}{\cellcolor{mkmcr}\textbf{MCRL}} & \multicolumn{4}{@{}c@{\hskip2.5pt{\color{gray!40}\vrule width 0.4pt}\hskip2.5pt}}{\cellcolor{mkkub}\textbf{MV-Kubric}} & \multicolumn{4}{@{}c@{\hskip2.5pt{\color{gray!40}\vrule width 0.4pt}\hskip2.5pt}}{\cellcolor{mkdex}\textbf{DexYCB}} & \multicolumn{4}{@{}c}{\cellcolor{mkavg}\textbf{Avg.}} \\
 & \multicolumn{2}{@{}c}{\cellcolor{mkhi4dl}$2$\,cm} & \multicolumn{2}{c@{\hskip2.5pt{\color{gray!40}\vrule width 0.4pt}\hskip2.5pt}}{\cellcolor{mkhi4dl}AUC} & \multicolumn{2}{@{}c}{\cellcolor{mkperpl}$2$\,cm} & \multicolumn{2}{c@{\hskip2.5pt{\color{gray!40}\vrule width 0.4pt}\hskip2.5pt}}{\cellcolor{mkperpl}AUC} & \multicolumn{2}{@{}c}{\cellcolor{mksynl}$2$\,cm} & \multicolumn{2}{c@{\hskip2.5pt{\color{gray!40}\vrule width 0.4pt}\hskip2.5pt}}{\cellcolor{mksynl}AUC} & \multicolumn{2}{@{}c}{\cellcolor{mkchargel}$2$\,cm} & \multicolumn{2}{c@{\hskip2.5pt{\color{gray!40}\vrule width 0.4pt}\hskip2.5pt}}{\cellcolor{mkchargel}AUC} & \multicolumn{2}{@{}c}{\cellcolor{mkmcrl}$2$\,cm} & \multicolumn{2}{c@{\hskip2.5pt{\color{gray!40}\vrule width 0.4pt}\hskip2.5pt}}{\cellcolor{mkmcrl}AUC} & \multicolumn{2}{@{}c}{\cellcolor{mkkubl}$2$\,cm} & \multicolumn{2}{c@{\hskip2.5pt{\color{gray!40}\vrule width 0.4pt}\hskip2.5pt}}{\cellcolor{mkkubl}AUC} & \multicolumn{2}{@{}c}{\cellcolor{mkdexl}$2$\,cm} & \multicolumn{2}{c@{\hskip2.5pt{\color{gray!40}\vrule width 0.4pt}\hskip2.5pt}}{\cellcolor{mkdexl}AUC} & \multicolumn{2}{@{}c}{\cellcolor{mkavgl}$2$\,cm} & \multicolumn{2}{c}{\cellcolor{mkavgl}AUC} \\
Method & \cellcolor{mkhi4dl}{\scriptsize Acc} & \cellcolor{mkhi4dl}{\scriptsize Comp} & \cellcolor{mkhi4dl}{\scriptsize $\text{AUC}_2$} & \cellcolor{mkhi4dl}{\scriptsize $\text{AUC}_5$} & \cellcolor{mkperpl}{\scriptsize Acc} & \cellcolor{mkperpl}{\scriptsize Comp} & \cellcolor{mkperpl}{\scriptsize $\text{AUC}_2$} & \cellcolor{mkperpl}{\scriptsize $\text{AUC}_5$} & \cellcolor{mksynl}{\scriptsize Acc} & \cellcolor{mksynl}{\scriptsize Comp} & \cellcolor{mksynl}{\scriptsize $\text{AUC}_2$} & \cellcolor{mksynl}{\scriptsize $\text{AUC}_5$} & \cellcolor{mkchargel}{\scriptsize Acc} & \cellcolor{mkchargel}{\scriptsize Comp} & \cellcolor{mkchargel}{\scriptsize $\text{AUC}_2$} & \cellcolor{mkchargel}{\scriptsize $\text{AUC}_5$} & \cellcolor{mkmcrl}{\scriptsize Acc} & \cellcolor{mkmcrl}{\scriptsize Comp} & \cellcolor{mkmcrl}{\scriptsize $\text{AUC}_2$} & \cellcolor{mkmcrl}{\scriptsize $\text{AUC}_5$} & \cellcolor{mkkubl}{\scriptsize Acc} & \cellcolor{mkkubl}{\scriptsize Comp} & \cellcolor{mkkubl}{\scriptsize $\text{AUC}_2$} & \cellcolor{mkkubl}{\scriptsize $\text{AUC}_5$} & \cellcolor{mkdexl}{\scriptsize Acc} & \cellcolor{mkdexl}{\scriptsize Comp} & \cellcolor{mkdexl}{\scriptsize $\text{AUC}_2$} & \cellcolor{mkdexl}{\scriptsize $\text{AUC}_5$} & \cellcolor{mkavgl}{\scriptsize Acc} & \cellcolor{mkavgl}{\scriptsize Comp} & \cellcolor{mkavgl}{\scriptsize $\text{AUC}_2$} & \cellcolor{mkavgl}{\scriptsize $\text{AUC}_5$} \\
\midrule
\multicolumn{33}{l}{\textit{Feed-forward}} \\
VGGT~\cite{wang2025vggt} & 32.1 & 45.2 & 24.2 & 54.2 & 23.2 & 26.5 & 14.0 & 36.0 & 22.3 & 31.3 & 12.0 & 43.3 & 24.2 & 35.4 & 14.0 & 47.2 & 58.0 & 56.7 & 45.1 & 72.8 & 13.2 & 28.8 & 8.4 & 37.6 & 68.9 & 87.3 & 73.1 & 90.6 & 34.6 & 44.5 & 27.3 & 54.5 \\
VGGT-$\Omega$~\cite{wang2026vggtomega} & 63.4 & 69.3 & 62.3 & 84.0 & 59.2 & 63.2 & 53.4 & \mkS{84.6} & 42.7 & 48.3 & 37.6 & 67.7 & 37.5 & 41.4 & 26.7 & 55.5 & 63.5 & \mkB{66.9} & 56.3 & \mkB{87.3} & 26.0 & 44.3 & 17.7 & 62.4 & 86.9 & \mkS{94.2} & \mkB{93.3} & \mkB{98.1} & 54.2 & 61.1 & 49.6 & 77.1 \\
Pi3X~\cite{wang2026pi3} & 58.4 & 66.9 & 47.6 & 79.9 & 46.6 & 51.5 & 31.7 & 63.3 & 41.6 & 46.0 & 32.2 & 66.2 & 36.4 & 42.5 & 24.2 & 55.5 & 61.4 & 60.3 & 50.4 & 81.3 & 13.5 & 29.9 & 8.0 & 35.2 & 87.6 & 90.7 & 89.8 & 97.2 & 49.4 & 55.4 & 40.5 & 68.4 \\
$\pi^4$~\cite{wang2026pi4} & 66.0 & 69.8 & 63.3 & 85.3 & 65.0 & \mkS{68.2} & \mkS{58.9} & \mkS{84.6} & 59.9 & 57.6 & 54.5 & 74.8 & 44.7 & 47.5 & 32.7 & 65.4 & 64.5 & 63.7 & \mkS{57.2} & \mkS{86.6} & 56.9 & 76.7 & 56.6 & 91.3 & 91.4 & 93.4 & \mkS{93.1} & \mkS{97.6} & 64.0 & 68.1 & 59.5 & \mkS{83.7} \\
OmniX~\cite{jiang2026omnix} & 61.2 & 69.1 & 55.1 & 84.2 & 36.6 & 40.2 & 27.6 & 59.7 & 45.7 & 46.9 & 35.3 & 67.3 & 36.8 & 44.0 & 24.0 & 54.2 & 65.6 & \mkS{66.0} & \mkB{57.6} & 85.4 & 16.2 & 33.6 & 8.8 & 39.5 & 87.3 & 92.6 & 91.1 & \mkS{97.6} & 49.9 & 56.1 & 42.8 & 69.7 \\
ARROW-FF~\cite{fradlin2026arrow} & 65.6 & 71.3 & 61.6 & \mkS{86.5} & 50.6 & 56.2 & 38.1 & 76.9 & 60.1 & 59.5 & 53.8 & 77.0 & 41.8 & 48.3 & 30.5 & 63.1 & 63.5 & 64.2 & 53.1 & \mkS{86.6} & 24.9 & 50.0 & 17.4 & 59.5 & 89.3 & 92.7 & 92.0 & 97.1 & 56.5 & 63.2 & 49.5 & 78.1 \\
\midrule
\multicolumn{33}{l}{\textit{Monocular, per video}} \\
Any4D~\cite{karhade2026any4d} & 13.5 & 24.8 & 6.0 & 23.6 & 20.1 & 26.4 & 8.3 & 34.1 & 14.2 & 24.3 & 8.0 & 34.0 & 10.9 & 20.2 & 5.1 & 21.3 & 31.5 & 33.8 & 16.8 & 57.1 & 10.7 & 27.6 & 5.8 & 30.5 & 65.0 & 88.2 & 72.5 & 90.6 & 23.7 & 35.0 & 17.5 & 41.6 \\
V-DPM~\cite{sucar2026vdpm} & 31.0 & 47.5 & 23.8 & 56.9 & 23.5 & 29.9 & 10.7 & 40.2 & 13.2 & 24.3 & 6.6 & 29.6 & 12.8 & 24.3 & 6.3 & 24.3 & 47.1 & 51.5 & 31.9 & 68.1 & 11.3 & 25.4 & 6.8 & 32.7 & 59.4 & 82.9 & 65.7 & 85.3 & 28.3 & 40.8 & 21.7 & 48.2 \\
MegaSaM~\cite{li2025megasam} & 20.6 & 33.3 & -- & -- & 25.4 & 33.9 & -- & -- & 17.2 & 24.0 & -- & -- & 13.1 & 25.7 & -- & -- & 42.3 & 44.7 & -- & -- & 5.8 & 11.8 & -- & -- & 67.3 & 87.5 & -- & -- & 27.4 & 37.3 & -- & -- \\
\midrule
\multicolumn{33}{l}{\textit{Multi-view optimization}} \\
Sun et al.~\cite{sun2026multiview} & 25.5 & 42.3 & -- & -- & 7.2 & 9.2 & -- & -- & 17.0 & 27.3 & -- & -- & 27.8 & 37.0 & -- & -- & 62.1 & 63.7 & -- & -- & 8.5 & 24.4 & -- & -- & 85.8 & \mkB{94.7} & -- & -- & 33.4 & 42.7 & -- & -- \\
Jojo4D (anchors only) & \mkB{96.8} & \mkS{80.2} & \mkS{79.8} & 85.4 & \mkB{85.6} & 41.1 & 38.5 & 51.3 & \mkB{88.5} & \mkS{67.9} & \mkS{67.4} & \mkS{82.4} & \mkB{87.6} & \mkS{66.3} & \mkS{64.8} & \mkS{73.5} & \mkB{82.2} & 49.8 & 44.6 & 68.8 & \mkB{96.6} & \mkS{86.9} & \mkS{86.7} & \mkS{91.4} & \mkB{98.1} & 85.4 & 84.5 & 92.1 & \mkB{90.8} & \mkS{68.2} & \mkS{66.6} & 77.8 \\
\textbf{Jojo4D (ours)} & \mkS{92.6} & \mkB{88.1} & \mkB{87.2} & \mkB{94.8} & \mkS{71.4} & \mkB{71.5} & \mkB{63.3} & \mkB{86.0} & \mkS{85.4} & \mkB{69.9} & \mkB{68.9} & \mkB{85.5} & \mkS{84.7} & \mkB{67.9} & \mkB{66.0} & \mkB{75.4} & \mkS{80.0} & 62.6 & 53.9 & 82.3 & \mkS{90.4} & \mkB{92.7} & \mkB{92.1} & \mkB{97.6} & \mkS{97.4} & 89.7 & 88.4 & 93.5 & \mkS{86.0} & \mkB{77.5} & \mkB{74.2} & \mkB{87.9} \\
\bottomrule
\end{tabular}}
\end{table*}

\begin{table*}[t]
\caption{\textbf{Point cloud reconstruction, mean over scenes.} Jojo4D still has the highest average F1 at both $2$ and $5$\,cm when each setting reports the mean instead of the median used in \cref{tab:point_cloud_results}. On Perpetua, the same clips with limited view overlap that raise Jojo4D's mean camera errors also lower its mean point cloud F1 (\cref{tab:supp_cameras_mean,sec:supp_failures}). Avg.\ averages settings within each dataset, then over all seven datasets. Dark and light green mark the best and second best results.}
\label{tab:supp_points_mean}
\mksetup
\resizebox{\linewidth}{!}{%
\footnotesize\setlength{\tabcolsep}{2.5pt}
\begin{tabular}{l@{\hskip5pt}cccccc@{\hskip2.5pt{\color{gray!40}\vrule width 0.4pt}\hskip2.5pt}cccc@{\hskip2.5pt{\color{gray!40}\vrule width 0.4pt}\hskip2.5pt}cccccc@{\hskip2.5pt{\color{gray!40}\vrule width 0.4pt}\hskip2.5pt}cccccc@{\hskip2.5pt{\color{gray!40}\vrule width 0.4pt}\hskip2.5pt}cc@{\hskip2.5pt{\color{gray!40}\vrule width 0.4pt}\hskip2.5pt}w{c}{2.1em}w{c}{2.1em}@{\hskip2.5pt{\color{gray!40}\vrule width 0.4pt}\hskip2.5pt}cc@{\hskip2.5pt{\color{gray!40}\vrule width 0.4pt}\hskip2.5pt}cccc}
\toprule
 & \multicolumn{6}{@{}c@{\hskip2.5pt{\color{gray!40}\vrule width 0.4pt}\hskip2.5pt}}{\cellcolor{mkhi4d}\textbf{Hi4D}} & \multicolumn{4}{@{}c@{\hskip2.5pt{\color{gray!40}\vrule width 0.4pt}\hskip2.5pt}}{\cellcolor{mkperp}\textbf{Perpetua}} & \multicolumn{6}{@{}c@{\hskip2.5pt{\color{gray!40}\vrule width 0.4pt}\hskip2.5pt}}{\cellcolor{mksyn}\textbf{Syn4D}} & \multicolumn{6}{@{}c@{\hskip2.5pt{\color{gray!40}\vrule width 0.4pt}\hskip2.5pt}}{\cellcolor{mkcharge}\textbf{CHARGE}} & \multicolumn{2}{@{}c@{\hskip2.5pt{\color{gray!40}\vrule width 0.4pt}\hskip2.5pt}}{\cellcolor{mkmcr}\textbf{MCRL}} & \multicolumn{2}{@{}c@{\hskip2.5pt{\color{gray!40}\vrule width 0.4pt}\hskip2.5pt}}{\cellcolor{mkkub}\textbf{MV-Kubric}} & \multicolumn{2}{@{}c@{\hskip2.5pt{\color{gray!40}\vrule width 0.4pt}\hskip2.5pt}}{\cellcolor{mkdex}\textbf{DexYCB}} & \multicolumn{4}{@{}c}{\cellcolor{mkavg}\textbf{Avg.}} \\
 & \multicolumn{2}{@{}c}{\cellcolor{mkhi4dl}2 cam} & \multicolumn{2}{c}{\cellcolor{mkhi4dl}4 cam} & \multicolumn{2}{c@{\hskip2.5pt{\color{gray!40}\vrule width 0.4pt}\hskip2.5pt}}{\cellcolor{mkhi4dl}8 cam} & \multicolumn{2}{@{}c}{\cellcolor{mkperpl}2 cam} & \multicolumn{2}{c@{\hskip2.5pt{\color{gray!40}\vrule width 0.4pt}\hskip2.5pt}}{\cellcolor{mkperpl}4 cam} & \multicolumn{2}{@{}c}{\cellcolor{mksynl}2 cam} & \multicolumn{2}{c}{\cellcolor{mksynl}4 cam} & \multicolumn{2}{c@{\hskip2.5pt{\color{gray!40}\vrule width 0.4pt}\hskip2.5pt}}{\cellcolor{mksynl}8 cam} & \multicolumn{2}{@{}c}{\cellcolor{mkchargel}2 cam} & \multicolumn{2}{c}{\cellcolor{mkchargel}4 cam} & \multicolumn{2}{c@{\hskip2.5pt{\color{gray!40}\vrule width 0.4pt}\hskip2.5pt}}{\cellcolor{mkchargel}8 cam} & \multicolumn{2}{@{}c@{\hskip2.5pt{\color{gray!40}\vrule width 0.4pt}\hskip2.5pt}}{\cellcolor{mkmcrl}2 cam} & \multicolumn{2}{@{}c@{\hskip2.5pt{\color{gray!40}\vrule width 0.4pt}\hskip2.5pt}}{\cellcolor{mkkubl}10 cam} & \multicolumn{2}{@{}c@{\hskip2.5pt{\color{gray!40}\vrule width 0.4pt}\hskip2.5pt}}{\cellcolor{mkdexl}8 cam} & \multicolumn{4}{@{}c}{\cellcolor{mkavgl}} \\
Method & \cellcolor{mkhi4dl}$\text{F}_2\!\uparrow$ & \cellcolor{mkhi4dl}$\text{F}_5\!\uparrow$ & \cellcolor{mkhi4dl}$\text{F}_2\!\uparrow$ & \cellcolor{mkhi4dl}$\text{F}_5\!\uparrow$ & \cellcolor{mkhi4dl}$\text{F}_2\!\uparrow$ & \cellcolor{mkhi4dl}$\text{F}_5\!\uparrow$ & \cellcolor{mkperpl}$\text{F}_2\!\uparrow$ & \cellcolor{mkperpl}$\text{F}_5\!\uparrow$ & \cellcolor{mkperpl}$\text{F}_2\!\uparrow$ & \cellcolor{mkperpl}$\text{F}_5\!\uparrow$ & \cellcolor{mksynl}$\text{F}_2\!\uparrow$ & \cellcolor{mksynl}$\text{F}_5\!\uparrow$ & \cellcolor{mksynl}$\text{F}_2\!\uparrow$ & \cellcolor{mksynl}$\text{F}_5\!\uparrow$ & \cellcolor{mksynl}$\text{F}_2\!\uparrow$ & \cellcolor{mksynl}$\text{F}_5\!\uparrow$ & \cellcolor{mkchargel}$\text{F}_2\!\uparrow$ & \cellcolor{mkchargel}$\text{F}_5\!\uparrow$ & \cellcolor{mkchargel}$\text{F}_2\!\uparrow$ & \cellcolor{mkchargel}$\text{F}_5\!\uparrow$ & \cellcolor{mkchargel}$\text{F}_2\!\uparrow$ & \cellcolor{mkchargel}$\text{F}_5\!\uparrow$ & \cellcolor{mkmcrl}$\text{F}_2\!\uparrow$ & \cellcolor{mkmcrl}$\text{F}_5\!\uparrow$ & \cellcolor{mkkubl}$\text{F}_2\!\uparrow$ & \cellcolor{mkkubl}$\text{F}_5\!\uparrow$ & \cellcolor{mkdexl}$\text{F}_2\!\uparrow$ & \cellcolor{mkdexl}$\text{F}_5\!\uparrow$ & \cellcolor{mkavgl}$\text{F}_2\!\uparrow$ & \cellcolor{mkavgl}$\text{F}_5\!\uparrow$ & \cellcolor{mkavgl}$\text{A}_2\!\uparrow$ & \cellcolor{mkavgl}$\text{A}_5\!\uparrow$ \\
\midrule
\multicolumn{33}{l}{\textit{Feed-forward}} \\
VGGT~\cite{wang2025vggt} & 38.1 & 63.3 & 30.1 & 52.0 & 37.3 & 60.2 & 42.1 & 68.9 & 14.3 & 30.5 & 28.7 & 50.1 & 30.4 & 51.1 & 27.0 & 46.9 & 34.6 & 58.1 & 21.2 & 44.2 & 23.7 & 47.1 & 59.0 & 82.2 & 18.8 & 43.3 & 76.2 & 89.8 & 38.9 & 60.4 & 28.5 & 53.3 \\
VGGT-$\Omega$~\cite{wang2026vggtomega} & 66.1 & 85.4 & 64.7 & 83.7 & 67.9 & 84.7 & \mkS{67.1} & \mkS{88.9} & 48.0 & \mkS{73.8} & 43.6 & 69.1 & 47.3 & 70.3 & 49.0 & 71.8 & 39.5 & 65.6 & 31.9 & 53.8 & 38.5 & 59.7 & 65.4 & 86.4 & 33.4 & 67.4 & 90.5 & 95.9 & 56.6 & 78.0 & 49.4 & 75.3 \\
Pi3X~\cite{wang2026pi3} & 63.6 & 86.7 & 58.8 & 81.6 & 57.3 & 79.5 & 59.7 & 80.6 & 34.8 & 59.6 & 40.6 & 66.5 & 46.5 & 69.8 & 40.5 & 67.1 & 40.9 & 63.8 & 32.2 & 53.2 & 33.5 & 56.1 & 64.4 & 85.6 & 19.3 & 48.6 & 88.9 & 95.6 & 51.1 & 72.6 & 40.6 & 66.0 \\
$\pi^4$~\cite{wang2026pi4} & 66.7 & 85.5 & 67.1 & 84.8 & 71.5 & 87.7 & \mkS{67.1} & 87.1 & \mkS{60.3} & \mkB{81.4} & 51.6 & 70.9 & 59.0 & 76.6 & 60.8 & 77.2 & 42.0 & 66.2 & 36.9 & 59.2 & 45.3 & 67.1 & \mkS{66.6} & 86.8 & 65.1 & 89.2 & \mkS{92.4} & \mkB{96.9} & 65.0 & 83.2 & 58.8 & \mkS{81.5} \\
OmniX~\cite{jiang2026omnix} & 60.1 & 84.1 & 60.8 & 83.8 & 60.5 & 83.3 & 49.1 & 76.0 & 25.2 & 48.8 & 42.8 & 68.9 & 50.1 & 73.8 & 46.6 & 72.1 & 39.6 & 62.6 & 29.8 & 51.8 & 37.2 & 59.5 & 66.4 & \mkB{87.5} & 21.5 & 50.4 & 89.8 & 95.5 & 51.0 & 72.7 & 42.0 & 67.8 \\
ARROW-FF~\cite{fradlin2026arrow} & 69.3 & \mkS{88.1} & 69.3 & 87.3 & 68.3 & 86.4 & 57.6 & 85.2 & 42.4 & 70.7 & 53.2 & 72.8 & 57.7 & 75.6 & 57.0 & 75.3 & 39.1 & 62.1 & 38.2 & 59.9 & 46.2 & 66.2 & 66.4 & \mkS{86.9} & 32.5 & 66.1 & 91.1 & 95.3 & 58.0 & 78.7 & 49.3 & 75.2 \\
\midrule
\multicolumn{33}{l}{\textit{Monocular, per video}} \\
Any4D~\cite{karhade2026any4d} & 16.1 & 40.4 & 17.4 & 39.7 & 18.6 & 41.5 & 24.4 & 51.9 & 19.8 & 43.9 & 19.0 & 41.6 & 20.2 & 44.3 & 21.4 & 43.6 & 12.6 & 29.8 & 12.8 & 29.9 & 16.0 & 33.9 & 32.4 & 59.1 & 16.1 & 40.8 & 74.3 & 85.7 & 28.0 & 49.8 & 18.9 & 40.2 \\
V-DPM~\cite{sucar2026vdpm} & 41.3 & 68.9 & 31.9 & 58.9 & 37.9 & 65.1 & 28.9 & 55.5 & 21.3 & 45.2 & 17.5 & 38.9 & 17.8 & 39.2 & 17.0 & 36.9 & 19.5 & 40.0 & 11.3 & 27.8 & 17.7 & 37.0 & 54.1 & 76.5 & 16.8 & 40.8 & 68.4 & 85.4 & 33.6 & 55.8 & 24.1 & 48.5 \\
MegaSaM~\cite{li2025megasam} & 44.8 & 68.5 & 15.7 & 38.1 & 17.0 & 40.9 & 29.0 & 50.4 & 25.8 & 47.1 & 36.2 & 58.6 & 9.6 & 27.7 & 13.2 & 34.9 & 20.3 & 40.2 & 11.8 & 25.5 & 13.8 & 29.0 & 48.9 & 72.9 & 7.7 & 20.4 & 76.2 & 88.0 & 31.6 & 50.2 & -- & -- \\
\midrule
\multicolumn{33}{l}{\textit{Multi-view optimization}} \\
Sun et al.~\cite{sun2026multiview} & 23.3 & 47.8 & 33.8 & 54.2 & 34.8 & 53.1 & 9.8 & 25.0 & 7.7 & 20.2 & 13.9 & 31.6 & 25.8 & 45.8 & 27.7 & 49.5 & 26.5 & 46.7 & 23.2 & 42.3 & 30.4 & 48.3 & 59.5 & 82.2 & 12.1 & 31.1 & 89.8 & 94.6 & 35.7 & 52.9 & -- & -- \\
Jojo4D (anchors only) & \mkS{80.6} & 87.2 & \mkS{82.7} & \mkS{87.6} & \mkS{90.4} & \mkS{93.7} & 59.9 & 69.7 & 53.4 & 65.2 & \mkS{66.6} & \mkS{81.1} & \mkS{75.6} & \mkS{88.2} & \mkS{79.9} & \mkS{91.1} & \mkB{71.8} & \mkS{81.5} & \mkS{70.0} & \mkS{78.2} & \mkB{77.2} & \mkS{84.3} & 59.8 & 77.7 & \mkB{91.2} & \mkS{95.1} & 90.9 & 95.6 & \mkS{75.7} & \mkS{84.8} & \mkS{65.4} & 76.6 \\
\textbf{Jojo4D (ours)} & \mkB{86.2} & \mkB{95.3} & \mkB{84.9} & \mkB{92.0} & \mkB{90.7} & \mkB{95.7} & \mkB{73.3} & \mkB{89.4} & \mkB{63.9} & \mkB{81.4} & \mkB{67.3} & \mkB{84.4} & \mkB{76.0} & \mkB{89.5} & \mkB{80.0} & \mkB{91.6} & \mkS{71.6} & \mkB{81.7} & \mkB{70.1} & \mkB{78.6} & \mkS{77.0} & \mkB{84.4} & \mkB{68.9} & \mkB{87.5} & \mkS{90.4} & \mkB{96.0} & \mkB{92.7} & \mkS{96.0} & \mkB{79.3} & \mkB{89.9} & \mkB{72.6} & \mkB{86.0} \\
\bottomrule
\end{tabular}}
\vspace{-0.3cm}
\end{table*}

\subsection{Calibrated cameras}
\label{sec:supp_calibrated}

\Cref{tab:supp_calibrated} repeats the camera evaluation of \cref{tab:cameras} with the ground-truth intrinsics as input. Pi3X receives them as an intrinsics prior and Any4D as camera rays. MegaSaM and Jojo4D hold them fixed in bundle adjustment, and Sun~\etal use them in place of the Pi3X intrinsics of the main comparison. Scenes, metrics and averaging follow \cref{tab:cameras}, without the intrinsics errors. Jojo4D has the lowest pose errors in every setting except on MultiCamRobolab, where the rotation errors of all methods except Any4D lie between $0.85^\circ$ and $0.95^\circ$. Errors in the ground-truth camera poses may explain part of this discrepancy (\cref{sec:supp_gt_accuracy}). Compared with the estimated intrinsics in \cref{tab:cameras}, the known intrinsics reduce the average rotation error of Jojo4D from $0.36^\circ$ to $0.27^\circ$, mostly on DexYCB, where it drops from $1.5^\circ$ to $0.79^\circ$.

\begin{table*}[t]
\caption{\textbf{Camera estimation with known intrinsics.} With ground-truth intrinsics, the pose errors of Jojo4D decrease further, most strongly on DexYCB, and Jojo4D has the lowest errors on every dataset except MultiCamRobolab, where the ground-truth camera poses may explain part of the measured rotation error (\cref{sec:supp_gt_accuracy}). All methods receive the ground-truth intrinsics as input. Shaded gray rows above each method repeat its errors with estimated intrinsics from \cref{tab:cameras} for comparison and are not ranked. Each setting reports the median over its scenes. Avg.\ first averages settings within each dataset, then averages over all seven datasets. Dark and light green mark the best and second best results. $R$ and $t$ are in degrees, ATE$_n$ in percent, and lower is better.}
\label{tab:supp_calibrated}
\mksetup
\resizebox{\linewidth}{!}{%
\footnotesize\setlength{\tabcolsep}{2.5pt}
\begin{tabular}{l@{\hskip5pt}cccccc@{\hskip2.5pt{\color{gray!40}\vrule width 0.4pt}\hskip2.5pt}cccc@{\hskip2.5pt{\color{gray!40}\vrule width 0.4pt}\hskip2.5pt}cccccc@{\hskip2.5pt{\color{gray!40}\vrule width 0.4pt}\hskip2.5pt}cccccc@{\hskip2.5pt{\color{gray!40}\vrule width 0.4pt}\hskip2.5pt}cc@{\hskip2.5pt{\color{gray!40}\vrule width 0.4pt}\hskip2.5pt}cc@{\hskip2.5pt{\color{gray!40}\vrule width 0.4pt}\hskip2.5pt}cc@{\hskip2.5pt{\color{gray!40}\vrule width 0.4pt}\hskip2.5pt}ccc}
\toprule
 & \multicolumn{6}{@{}c@{\hskip2.5pt{\color{gray!40}\vrule width 0.4pt}\hskip2.5pt}}{\cellcolor{mkhi4d}\textbf{Hi4D}} & \multicolumn{4}{@{}c@{\hskip2.5pt{\color{gray!40}\vrule width 0.4pt}\hskip2.5pt}}{\cellcolor{mkperp}\textbf{Perpetua}} & \multicolumn{6}{@{}c@{\hskip2.5pt{\color{gray!40}\vrule width 0.4pt}\hskip2.5pt}}{\cellcolor{mksyn}\textbf{Syn4D}} & \multicolumn{6}{@{}c@{\hskip2.5pt{\color{gray!40}\vrule width 0.4pt}\hskip2.5pt}}{\cellcolor{mkcharge}\textbf{CHARGE}} & \multicolumn{2}{@{}c@{\hskip2.5pt{\color{gray!40}\vrule width 0.4pt}\hskip2.5pt}}{\cellcolor{mkmcr}\textbf{MCRL}} & \multicolumn{2}{@{}c@{\hskip2.5pt{\color{gray!40}\vrule width 0.4pt}\hskip2.5pt}}{\cellcolor{mkkub}\textbf{MV-Kubric}} & \multicolumn{2}{@{}c@{\hskip2.5pt{\color{gray!40}\vrule width 0.4pt}\hskip2.5pt}}{\cellcolor{mkdex}\textbf{DexYCB}} & \multicolumn{3}{@{}c}{\cellcolor{mkavg}\textbf{Avg.}} \\
 & \multicolumn{2}{@{}c}{\cellcolor{mkhi4dl}2 cam} & \multicolumn{2}{c}{\cellcolor{mkhi4dl}4 cam} & \multicolumn{2}{c@{\hskip2.5pt{\color{gray!40}\vrule width 0.4pt}\hskip2.5pt}}{\cellcolor{mkhi4dl}8 cam} & \multicolumn{2}{@{}c}{\cellcolor{mkperpl}2 cam} & \multicolumn{2}{c@{\hskip2.5pt{\color{gray!40}\vrule width 0.4pt}\hskip2.5pt}}{\cellcolor{mkperpl}4 cam} & \multicolumn{2}{@{}c}{\cellcolor{mksynl}2 cam} & \multicolumn{2}{c}{\cellcolor{mksynl}4 cam} & \multicolumn{2}{c@{\hskip2.5pt{\color{gray!40}\vrule width 0.4pt}\hskip2.5pt}}{\cellcolor{mksynl}8 cam} & \multicolumn{2}{@{}c}{\cellcolor{mkchargel}2 cam} & \multicolumn{2}{c}{\cellcolor{mkchargel}4 cam} & \multicolumn{2}{c@{\hskip2.5pt{\color{gray!40}\vrule width 0.4pt}\hskip2.5pt}}{\cellcolor{mkchargel}8 cam} & \multicolumn{2}{@{}c@{\hskip2.5pt{\color{gray!40}\vrule width 0.4pt}\hskip2.5pt}}{\cellcolor{mkmcrl}2 cam} & \multicolumn{2}{@{}c@{\hskip2.5pt{\color{gray!40}\vrule width 0.4pt}\hskip2.5pt}}{\cellcolor{mkkubl}10 cam} & \multicolumn{2}{@{}c@{\hskip2.5pt{\color{gray!40}\vrule width 0.4pt}\hskip2.5pt}}{\cellcolor{mkdexl}8 cam} & \multicolumn{3}{@{}c}{\cellcolor{mkavgl}} \\
Method & \cellcolor{mkhi4dl}$R\!\downarrow$ & \cellcolor{mkhi4dl}$t\!\downarrow$ & \cellcolor{mkhi4dl}$R\!\downarrow$ & \cellcolor{mkhi4dl}$t\!\downarrow$ & \cellcolor{mkhi4dl}$R\!\downarrow$ & \cellcolor{mkhi4dl}$t\!\downarrow$ & \cellcolor{mkperpl}$R\!\downarrow$ & \cellcolor{mkperpl}$t\!\downarrow$ & \cellcolor{mkperpl}$R\!\downarrow$ & \cellcolor{mkperpl}$t\!\downarrow$ & \cellcolor{mksynl}$R\!\downarrow$ & \cellcolor{mksynl}$t\!\downarrow$ & \cellcolor{mksynl}$R\!\downarrow$ & \cellcolor{mksynl}$t\!\downarrow$ & \cellcolor{mksynl}$R\!\downarrow$ & \cellcolor{mksynl}$t\!\downarrow$ & \cellcolor{mkchargel}$R\!\downarrow$ & \cellcolor{mkchargel}$t\!\downarrow$ & \cellcolor{mkchargel}$R\!\downarrow$ & \cellcolor{mkchargel}$t\!\downarrow$ & \cellcolor{mkchargel}$R\!\downarrow$ & \cellcolor{mkchargel}$t\!\downarrow$ & \cellcolor{mkmcrl}$R\!\downarrow$ & \cellcolor{mkmcrl}$t\!\downarrow$ & \cellcolor{mkkubl}$R\!\downarrow$ & \cellcolor{mkkubl}$t\!\downarrow$ & \cellcolor{mkdexl}$R\!\downarrow$ & \cellcolor{mkdexl}$t\!\downarrow$ & \cellcolor{mkavgl}$R\!\downarrow$ & \cellcolor{mkavgl}$t\!\downarrow$ & \cellcolor{mkavgl}ATE$_n\!\downarrow$ \\
\midrule
\multicolumn{32}{l}{\textit{Feed-forward}} \\
\rowcolor{black!5}\textcolor{gray}{\textit{Pi3X, estimated $K$}} & \textcolor{gray}{.49} & \textcolor{gray}{1.2} & \textcolor{gray}{.83} & \textcolor{gray}{1.2} & \textcolor{gray}{1.0} & \textcolor{gray}{1.3} & \textcolor{gray}{.82} & \textcolor{gray}{2.0} & \textcolor{gray}{1.7} & \textcolor{gray}{2.8} & \textcolor{gray}{.50} & \textcolor{gray}{1.8} & \textcolor{gray}{.62} & \textcolor{gray}{1.6} & \textcolor{gray}{.85} & \textcolor{gray}{1.7} & \textcolor{gray}{.49} & \textcolor{gray}{5.6} & \textcolor{gray}{.93} & \textcolor{gray}{5.6} & \textcolor{gray}{1.1} & \textcolor{gray}{5.6} & \textcolor{gray}{.90} & \textcolor{gray}{2.9} & \textcolor{gray}{7.8} & \textcolor{gray}{8.8} & \textcolor{gray}{1.5} & \textcolor{gray}{1.5} & \textcolor{gray}{2.0} & \textcolor{gray}{3.4} & \textcolor{gray}{1.5} \\
Pi3X~\cite{wang2026pi3} & .49 & 1.3 & .77 & 1.3 & .84 & 1.2 & .70 & 1.5 & 1.4 & 2.5 & .37 & 1.3 & \mkS{.32} & 1.2 & .67 & 1.2 & .31 & 3.8 & .84 & 3.7 & 1.0 & 4.2 & \mkB{.85} & \mkB{2.5} & 4.5 & 6.1 & 1.6 & 1.2 & 1.4 & 2.6 & 1.1 \\
\midrule
\multicolumn{32}{l}{\textit{Monocular, per video}} \\
\rowcolor{black!5}\textcolor{gray}{\textit{Any4D, estimated $K$}} & \textcolor{gray}{2.3} & \textcolor{gray}{4.4} & \textcolor{gray}{2.3} & \textcolor{gray}{3.0} & \textcolor{gray}{2.0} & \textcolor{gray}{2.7} & \textcolor{gray}{3.0} & \textcolor{gray}{7.2} & \textcolor{gray}{3.1} & \textcolor{gray}{5.7} & \textcolor{gray}{.98} & \textcolor{gray}{4.9} & \textcolor{gray}{1.0} & \textcolor{gray}{4.0} & \textcolor{gray}{.89} & \textcolor{gray}{2.3} & \textcolor{gray}{1.9} & \textcolor{gray}{26} & \textcolor{gray}{1.6} & \textcolor{gray}{9.6} & \textcolor{gray}{1.4} & \textcolor{gray}{10} & \textcolor{gray}{1.4} & \textcolor{gray}{7.9} & \textcolor{gray}{1.1} & \textcolor{gray}{.92} & \textcolor{gray}{1.5} & \textcolor{gray}{1.1} & \textcolor{gray}{1.7} & \textcolor{gray}{5.5} & \textcolor{gray}{2.2} \\
Any4D~\cite{karhade2026any4d} & 1.5 & 3.4 & 1.7 & 2.5 & 1.5 & 2.2 & 4.4 & 7.3 & 4.6 & 6.3 & 1.2 & 4.2 & 1.0 & 3.4 & .94 & 1.9 & 1.3 & 23 & 1.4 & 8.7 & 1.1 & 9.6 & 1.6 & 7.4 & 1.1 & .92 & \mkS{1.5} & 1.1 & 1.8 & 5.1 & 2.1 \\
\rowcolor{black!5}\textcolor{gray}{\textit{MegaSaM, estimated $K$}} & \textcolor{gray}{.64} & \textcolor{gray}{1.1} & \textcolor{gray}{.68} & \textcolor{gray}{1.1} & \textcolor{gray}{.76} & \textcolor{gray}{1.2} & \textcolor{gray}{.35} & \textcolor{gray}{.86} & \textcolor{gray}{.96} & \textcolor{gray}{1.8} & \textcolor{gray}{.28} & \textcolor{gray}{.33} & \textcolor{gray}{.35} & \textcolor{gray}{.57} & \textcolor{gray}{.37} & \textcolor{gray}{.42} & \textcolor{gray}{.84} & \textcolor{gray}{2.6} & \textcolor{gray}{1.2} & \textcolor{gray}{2.0} & \textcolor{gray}{.92} & \textcolor{gray}{1.8} & \textcolor{gray}{.87} & \textcolor{gray}{2.5} & \textcolor{gray}{1.1} & \textcolor{gray}{.92} & \textcolor{gray}{1.5} & \textcolor{gray}{1.0} & \textcolor{gray}{.87} & \textcolor{gray}{1.4} & \textcolor{gray}{.59} \\
MegaSaM~\cite{li2025megasam} & \mkS{.33} & \mkS{.46} & \mkS{.50} & \mkS{.80} & \mkS{.61} & \mkS{.97} & \mkS{.54} & \mkS{.84} & \mkS{.95} & \mkS{1.8} & \mkS{.25} & \mkS{.32} & .33 & \mkS{.56} & \mkS{.36} & \mkS{.40} & \mkS{.23} & \mkS{1.9} & \mkS{.65} & \mkS{1.3} & \mkS{.57} & \mkS{1.1} & .88 & 2.8 & 1.1 & .92 & \mkS{1.5} & \mkS{1.0} & \mkS{.78} & \mkS{1.2} & \mkS{.48} \\
\midrule
\multicolumn{32}{l}{\textit{Multi-view optimization}} \\
\rowcolor{black!5}\textcolor{gray}{\textit{Sun~\etal, estimated $K$}} & \textcolor{gray}{1.2} & \textcolor{gray}{2.2} & \textcolor{gray}{1.2} & \textcolor{gray}{1.6} & \textcolor{gray}{1.3} & \textcolor{gray}{1.7} & \textcolor{gray}{2.6} & \textcolor{gray}{4.3} & \textcolor{gray}{7.3} & \textcolor{gray}{11} & \textcolor{gray}{1.1} & \textcolor{gray}{1.9} & \textcolor{gray}{1.0} & \textcolor{gray}{1.5} & \textcolor{gray}{1.3} & \textcolor{gray}{1.3} & \textcolor{gray}{.86} & \textcolor{gray}{6.1} & \textcolor{gray}{2.0} & \textcolor{gray}{3.7} & \textcolor{gray}{1.2} & \textcolor{gray}{3.4} & \textcolor{gray}{.91} & \textcolor{gray}{2.8} & \textcolor{gray}{1.4} & \textcolor{gray}{1.7} & \textcolor{gray}{1.5} & \textcolor{gray}{1.2} & \textcolor{gray}{1.8} & \textcolor{gray}{3.0} & \textcolor{gray}{1.7} \\
Sun et al.~\cite{sun2026multiview} & .92 & 2.0 & 1.2 & 1.3 & 1.3 & 1.4 & 2.5 & 4.3 & 9.5 & 11 & 1.0 & 1.8 & .99 & 1.4 & 1.2 & 1.3 & .80 & 5.6 & 2.0 & 3.3 & 1.2 & 3.2 & \mkS{.87} & 3.1 & \mkS{.82} & \mkS{.61} & 1.6 & \mkS{1.0} & 1.8 & 2.8 & 1.7 \\
\rowcolor{black!5}\textcolor{gray}{\textit{Jojo4D (ours), estimated $K$}} & \textcolor{gray}{.02} & \textcolor{gray}{.02} & \textcolor{gray}{.01} & \textcolor{gray}{.01} & \textcolor{gray}{.01} & \textcolor{gray}{.01} & \textcolor{gray}{.09} & \textcolor{gray}{.21} & \textcolor{gray}{.10} & \textcolor{gray}{.17} & \textcolor{gray}{.04} & \textcolor{gray}{.06} & \textcolor{gray}{.03} & \textcolor{gray}{.03} & \textcolor{gray}{.02} & \textcolor{gray}{.02} & \textcolor{gray}{.02} & \textcolor{gray}{.13} & \textcolor{gray}{.02} & \textcolor{gray}{.06} & \textcolor{gray}{.01} & \textcolor{gray}{.04} & \textcolor{gray}{.84} & \textcolor{gray}{2.4} & \textcolor{gray}{.07} & \textcolor{gray}{.06} & \textcolor{gray}{1.5} & \textcolor{gray}{1.0} & \textcolor{gray}{.36} & \textcolor{gray}{.54} & \textcolor{gray}{.13} \\
\textbf{Jojo4D (ours)} & \mkB{.01} & \mkB{.01} & \mkB{.01} & \mkB{.005} & \mkB{.004} & \mkB{.003} & \mkB{.09} & \mkB{.20} & \mkB{.09} & \mkB{.17} & \mkB{.01} & \mkB{.03} & \mkB{.01} & \mkB{.01} & \mkB{.004} & \mkB{.01} & \mkB{.01} & \mkB{.12} & \mkB{.02} & \mkB{.04} & \mkB{.01} & \mkB{.03} & .95 & \mkS{2.7} & \mkB{.01} & \mkB{.01} & \mkB{.79} & \mkB{.49} & \mkB{.27} & \mkB{.49} & \mkB{.13} \\
\bottomrule
\end{tabular}}
\end{table*}

\section{Point Tracking on TAPVid-MV}
\label{sec:supp_tracking}

On TAPVid-MV~\cite{koppula2026tapvidmv}, a tracker must recover the 3D trajectory of a point marked in one view of a synchronized recording. We run each tracker on the same videos and change only which method supplies its cameras and depth maps: the benchmark's VGGT-$\Omega$, $\pi^4$, ARROW-FF, or Jojo4D with ARROW. We use the standard camera estimation settings. To provide denser depth maps for tracking, we also fill pixels at depth boundaries or with low ARROW confidence.

We use TAPVid-MV's standard input resolution ($512$ pixels on the longer side) and provide no ground-truth geometry. We evaluate every benchmark sequence in DROID, Waymo, PACE, Hi4D, and Perpetua. For EgoExo4D, the TAPVid-MV authors shared a pre-release with annotations for $9$ of the $18$ sequences. At the time of evaluation, annotations for the remaining EgoExo4D sequences and for Harmony4D were not yet available to us.

\Cref{tab:supp_tracking} compares CoTracker3~\cite{karaev2025cotracker3}, MVTracker~\cite{rajic2025mvtracker}, and TAPIP3D~\cite{zhang2025tapip3d}. We lift CoTracker3's 2D tracks to 3D with each reconstruction. We also run MVTracker fine-tuned on Perpetua (FT). Jojo4D fails on one of the $200$ sequences, on Waymo. We keep the failed sequence in the average and assign zero to all three tracking metrics for each tracker. VGGT-$\Omega$, $\pi^4$, and ARROW-FF have no failures. With Jojo4D instead of VGGT-$\Omega$, all three original trackers improve on all three metrics on Hi4D, Perpetua, and EgoExo4D, but score lower on DROID and Waymo.

Trackers using Jojo4D achieve state-of-the-art 3D tracking accuracy on Hi4D, Perpetua and the EgoExo4D pre-release in both world and camera coordinates. TAPIP3D with Jojo4D also ranks first on average across the six datasets. On PACE, it leads in both camera-space metrics and comes within $0.3$ percentage points of the world-space record ($44.7\%$ versus $45.0\%$). On Hi4D, TAPIP3D's world-space accuracy with Jojo4D is more than twice the previous best ($48.7\%$ versus $20.8\%$).

\begin{table*}[t]
\mksetup
\caption{\textbf{3D point tracking accuracy on TAPVid-MV.} With Jojo4D, TAPIP3D ranks first on average in both world and camera coordinates. Its average world-space accuracy rises from $23.9\%$ with VGGT-$\Omega$ to $33.4\%$ with Jojo4D. W measures 3D tracking accuracy in world coordinates, N in the non-query cameras, and Q in the query camera (percent, higher is better). We average over sequences within each dataset and give each dataset equal weight in the average. The benchmark row averages the best scores from each subset. Dark and light green mark the best and second best. For EgoExo4D ($^{*}$), we rerun TAPIP3D with VGGT-$\Omega$, the benchmark's top tracker on the original annotations, on the nine reviewed sequences. Fine-tuned MVTracker is unranked on its training dataset, Perpetua, and in the averages that include it.}
\label{tab:supp_tracking}
\footnotesize\setlength{\tabcolsep}{2pt}
\resizebox{\ifdim\width>\linewidth\linewidth\else\width\fi}{!}{%
\begin{tabular}{@{}ll*{7}{ccc}@{}}
\toprule
 &  & \multicolumn{3}{c}{\textbf{Avg.}} & \multicolumn{3}{c}{\textbf{DROID}} & \multicolumn{3}{c}{\textbf{EgoExo4D}} & \multicolumn{3}{c}{\textbf{PACE}} & \multicolumn{3}{c}{\textbf{Hi4D}} & \multicolumn{3}{c}{\textbf{Waymo}} & \multicolumn{3}{c}{\textbf{Perpetua}} \\
\cmidrule(lr){3-5}\cmidrule(lr){6-8}\cmidrule(lr){9-11}\cmidrule(lr){12-14}\cmidrule(lr){15-17}\cmidrule(lr){18-20}\cmidrule(lr){21-23}
Tracker & Reconstruction & W$\,\uparrow$ & N$\,\uparrow$ & Q$\,\uparrow$ & W$\,\uparrow$ & N$\,\uparrow$ & Q$\,\uparrow$ & W$\,\uparrow$ & N$\,\uparrow$ & Q$\,\uparrow$ & W$\,\uparrow$ & N$\,\uparrow$ & Q$\,\uparrow$ & W$\,\uparrow$ & N$\,\uparrow$ & Q$\,\uparrow$ & W$\,\uparrow$ & N$\,\uparrow$ & Q$\,\uparrow$ & W$\,\uparrow$ & N$\,\uparrow$ & Q$\,\uparrow$ \\
\midrule
\multicolumn{2}{@{}l}{Best in TAPVid-MV~\cite{koppula2026tapvidmv}} & 24.3 & 22.9 & 27.4 & \mkB{6.0} & 12.4 & \mkS{\underline{13.1}} & 54.1$^{*}$ & 47.0$^{*}$ & 56.1$^{*}$ & \mkB{45.0} & 35.7 & 36.5 & 20.8 & 16.4 & 19.6 & 3.7 & 12.1 & 16.5 & 16.1 & 13.4 & 22.5 \\
\midrule
\multirow{4}{*}{CoTracker3} & VGGT-$\Omega$ & 21.2 & 18.3 & 25.8 & 5.7 & 10.7 & 12.7 & 49.6 & 43.0 & 54.3 & 42.8 & 34.7 & 35.3 & 16.3 & 12.2 & 18.7 & 2.4 & 4.2 & 14.8 & 10.6 & 5.1 & 18.9 \\
 & $\pi^4$ & 22.4 & 20.4 & 28.1 & 1.9 & 11.3 & 13.0 & 52.3 & 48.0 & 57.2 & 41.5 & 33.4 & 34.3 & 20.4 & 17.6 & 24.4 & 4.3 & 4.5 & 16.4 & 14.3 & 7.6 & 23.5 \\
 & ARROW-FF & 22.2 & 19.0 & 27.2 & 0.4 & 2.5 & 6.8 & 55.7 & 47.7 & 58.9 & 41.3 & 34.2 & 35.8 & 22.0 & 20.2 & 27.5 & 2.0 & 3.0 & 8.8 & 11.8 & 6.6 & 25.3 \\
 & \textbf{Jojo4D (ours)} & 28.1 & 24.5 & \mkS{\underline{31.6}} & 3.6 & 5.3 & 6.4 & 66.6 & 59.1 & \mkS{\underline{61.7}} & 42.8 & 37.5 & \mkS{\underline{38.2}} & 37.0 & 34.9 & \mkS{\underline{51.5}} & 0.4 & 0.7 & 3.5 & 18.1 & 9.7 & 28.2 \\
\midrule
\multirow{4}{*}{MVTracker} & VGGT-$\Omega$ & 23.2 & 22.0 & 24.7 & 5.2 & 11.0 & 11.0 & 51.8 & 46.0 & 50.7 & 43.7 & 35.2 & 35.4 & 19.1 & 15.1 & 17.8 & 3.4 & 11.2 & 15.1 & 16.1 & 13.4 & 18.4 \\
 & $\pi^4$ & 24.9 & 25.0 & 27.6 & 1.7 & \mkS{\underline{13.4}} & 12.3 & 54.7 & 51.0 & 54.7 & 40.2 & 31.8 & 32.6 & 24.6 & 22.1 & 24.0 & \mkS{\underline{6.0}} & \mkS{\underline{12.2}} & 16.8 & 22.2 & \mkS{\underline{19.8}} & 25.2 \\
 & ARROW-FF & 24.1 & 22.2 & 25.4 & 0.4 & 2.8 & 6.0 & 58.4 & 51.3 & 55.2 & 40.8 & 34.5 & 35.1 & 24.0 & 22.0 & 24.6 & 2.8 & 6.1 & 8.5 & 18.0 & 16.6 & 23.1 \\
 & \textbf{Jojo4D (ours)} & \mkS{\underline{30.1}} & \mkS{\underline{27.5}} & 29.6 & 3.5 & 5.5 & 5.2 & 67.2 & 60.5 & 59.8 & 42.0 & 36.7 & 37.1 & 39.0 & 35.9 & 41.7 & 0.5 & 1.9 & 3.7 & \mkB{28.1} & \mkB{24.2} & \mkS{\underline{30.3}} \\
\midrule
\multirow{4}{*}{MVTracker (FT)} & VGGT-$\Omega$ & \textcolor{gray}{24.6} & \textcolor{gray}{23.3} & \textcolor{gray}{25.8} & \mkB{6.0} & 12.4 & 11.8 & 51.8 & 46.4 & 51.0 & \mkB{45.0} & 35.3 & 35.8 & 20.8 & 16.4 & 18.8 & 3.7 & 12.1 & 15.9 & \textcolor{gray}{20.2} & \textcolor{gray}{17.3} & \textcolor{gray}{21.5} \\
 & $\pi^4$ & \textcolor{gray}{26.8} & \textcolor{gray}{27.2} & \textcolor{gray}{29.2} & 1.8 & \mkB{14.7} & 12.8 & 55.9 & 52.0 & 55.9 & 42.2 & 32.9 & 33.6 & 26.4 & 24.1 & 25.4 & \mkB{6.3} & \mkB{13.4} & \mkS{\underline{17.9}} & \textcolor{gray}{28.3} & \textcolor{gray}{25.9} & \textcolor{gray}{29.8} \\
 & ARROW-FF & \textcolor{gray}{25.6} & \textcolor{gray}{23.6} & \textcolor{gray}{26.7} & 0.5 & 2.8 & 6.2 & 58.9 & 51.9 & 56.1 & 43.0 & 35.1 & 35.8 & 26.1 & 24.2 & 26.1 & 2.9 & 6.4 & 8.8 & \textcolor{gray}{22.1} & \textcolor{gray}{21.3} & \textcolor{gray}{27.1} \\
 & \textbf{Jojo4D (ours)} & \textcolor{gray}{33.1} & \textcolor{gray}{30.0} & \textcolor{gray}{32.0} & 3.8 & 6.4 & 5.8 & \mkS{\underline{69.2}} & \mkS{\underline{61.3}} & 61.0 & 43.5 & \mkS{\underline{37.9}} & \mkS{\underline{38.2}} & \mkS{\underline{45.0}} & \mkS{\underline{41.0}} & 46.5 & 0.5 & 2.1 & 3.9 & \textcolor{gray}{36.4} & \textcolor{gray}{31.5} & \textcolor{gray}{36.5} \\
\midrule
\multirow{4}{*}{TAPIP3D} & VGGT-$\Omega$ & 23.9 & 21.7 & 27.2 & \mkS{\underline{5.8}} & 11.6 & 12.7 & 54.1 & 47.0 & 56.1 & 44.1 & 35.5 & 35.8 & 20.5 & 15.8 & 19.6 & 3.5 & 10.1 & 16.5 & 15.5 & 10.0 & 22.2 \\
 & $\pi^4$ & 26.0 & 25.0 & 30.4 & 2.1 & 13.3 & \mkB{13.2} & 57.3 & 52.5 & 59.9 & 43.2 & 34.5 & 35.3 & 26.0 & 23.8 & 26.3 & 5.9 & 10.5 & \mkB{18.4} & 21.4 & 15.5 & 29.3 \\
 & ARROW-FF & 25.0 & 22.1 & 28.8 & 0.4 & 2.7 & 6.8 & 60.2 & 51.6 & 61.3 & 43.4 & 35.4 & 36.9 & 27.3 & 26.2 & 29.3 & 2.7 & 5.4 & 9.2 & 16.0 & 11.5 & 29.1 \\
 & \textbf{Jojo4D (ours)} & \mkB{33.4} & \mkB{29.9} & \mkB{34.7} & 3.9 & 6.1 & 6.2 & \mkB{74.7} & \mkB{65.6} & \mkB{65.4} & \mkS{\underline{44.7}} & \mkB{38.9} & \mkB{39.5} & \mkB{48.7} & \mkB{47.7} & \mkB{57.1} & 0.5 & 1.7 & 4.0 & \mkS{\underline{27.8}} & 19.6 & \mkB{35.9} \\
\bottomrule
\end{tabular}}
\end{table*}

To distinguish camera errors from depth errors, \cref{tab:supp_tapvid_geometry} evaluates the reconstructions supplied to the trackers. TAPVid-MV measures intrinsics error $K$ as the mean angle between predicted and ground-truth viewing rays. Its pose scores are AUC at $30^\circ$ for frame pairs from the same camera ($E_C$) or from different cameras ($E_W$), excluding pairs with negligible ground-truth translation. Its depth scores average the fraction of valid ground-truth pixels reconstructed within thresholds from $1$ to $10$\,cm, counting missing predictions as misses. $D_C$ aligns each video separately, whereas $D_W$ fits one similarity across all cameras.

\begin{table*}[t]
\mksetup
\caption{\textbf{Camera and depth quality on TAPVid-MV.} Jojo4D gives the most accurate relative poses within a video on average ($E_C$), while $\pi^4$ leads across cameras ($E_W$). Lower is better for intrinsics error $K$ (degrees), higher for the pose and depth scores (percent). Each dataset has equal weight in the average. As in TAPVid-MV, we report depth on PACE, Hi4D, and Perpetua. Dark and light green mark the best and second-best scores.}
\label{tab:supp_tapvid_geometry}
\footnotesize\setlength{\tabcolsep}{2pt}
\resizebox{\ifdim\width>\linewidth\linewidth\else\width\fi}{!}{%
\begin{tabular}{@{}l*{29}{c}@{}}
\toprule
 & \multicolumn{5}{c}{\textbf{Avg.}} & \multicolumn{3}{c}{\textbf{DROID}} & \multicolumn{3}{c}{\textbf{EgoExo4D}} & \multicolumn{5}{c}{\textbf{PACE}} & \multicolumn{5}{c}{\textbf{Hi4D}} & \multicolumn{3}{c}{\textbf{Waymo}} & \multicolumn{5}{c}{\textbf{Perpetua}} \\
\cmidrule(lr){2-6}\cmidrule(lr){7-9}\cmidrule(lr){10-12}\cmidrule(lr){13-17}\cmidrule(lr){18-22}\cmidrule(lr){23-25}\cmidrule(lr){26-30}
Reconstruction & $K\!\downarrow$ & $E_C\!\uparrow$ & $E_W\!\uparrow$ & $D_C\!\uparrow$ & $D_W\!\uparrow$ & $K\!\downarrow$ & $E_C\!\uparrow$ & $E_W\!\uparrow$ & $K\!\downarrow$ & $E_C\!\uparrow$ & $E_W\!\uparrow$ & $K\!\downarrow$ & $E_C\!\uparrow$ & $E_W\!\uparrow$ & $D_C\!\uparrow$ & $D_W\!\uparrow$ & $K\!\downarrow$ & $E_C\!\uparrow$ & $E_W\!\uparrow$ & $D_C\!\uparrow$ & $D_W\!\uparrow$ & $K\!\downarrow$ & $E_C\!\uparrow$ & $E_W\!\uparrow$ & $K\!\downarrow$ & $E_C\!\uparrow$ & $E_W\!\uparrow$ & $D_C\!\uparrow$ & $D_W\!\uparrow$ \\
\midrule
VGGT-$\Omega$ & \mkB{0.79} & 81.0 & 90.4 & 47.0 & \mkS{\underline{43.6}} & \mkB{1.39} & 56.0 & \mkB{85.9} & \mkS{\underline{0.29}} & 88.1 & 98.9 & 0.78 & 72.5 & 89.4 & 74.3 & 70.5 & 0.95 & 92.6 & 95.6 & 45.0 & 40.7 & 0.92 & \mkS{\underline{93.1}} & \mkS{\underline{86.6}} & 0.41 & 83.7 & 86.1 & \mkS{\underline{21.8}} & \mkS{\underline{19.7}} \\
$\pi^4$ & 0.94 & \mkS{\underline{86.9}} & \mkB{92.5} & \mkB{53.4} & \mkB{50.5} & 2.77 & \mkS{\underline{69.8}} & \mkS{\underline{84.4}} & \mkB{0.17} & \mkS{\underline{94.0}} & \mkB{99.9} & 0.89 & 71.8 & 89.5 & 74.7 & 71.3 & \mkS{\underline{0.90}} & \mkS{\underline{97.0}} & \mkS{\underline{98.0}} & \mkB{54.5} & \mkB{53.0} & \mkS{\underline{0.73}} & \mkB{94.2} & \mkB{89.5} & 0.16 & \mkS{\underline{94.9}} & \mkB{94.0} & \mkB{31.0} & \mkB{27.2} \\
ARROW-FF & \mkS{\underline{0.80}} & 78.3 & 87.0 & 42.7 & 39.5 & \mkS{\underline{2.03}} & 37.4 & 61.3 & 0.31 & 85.5 & 98.5 & \mkS{\underline{0.71}} & \mkS{\underline{72.6}} & \mkS{\underline{90.5}} & \mkB{75.3} & \mkB{71.6} & 0.91 & 93.7 & 96.2 & 37.3 & 32.4 & \mkB{0.70} & 92.3 & 86.5 & \mkS{\underline{0.11}} & 87.9 & 88.8 & 15.5 & 14.5 \\
\textbf{Jojo4D (ours)} & 0.99 & \mkB{89.4} & \mkS{\underline{90.8}} & \mkS{\underline{48.9}} & 43.3 & 2.74 & \mkB{73.5} & 82.3 & 0.34 & \mkB{98.6} & \mkS{\underline{99.5}} & \mkB{0.66} & \mkB{76.0} & \mkB{92.0} & \mkS{\underline{74.9}} & \mkS{\underline{71.4}} & \mkB{0.02} & \mkB{100.0} & \mkB{100.0} & \mkS{\underline{52.0}} & \mkS{\underline{47.3}} & 2.14 & 90.9 & 80.1 & \mkB{0.04} & \mkB{97.5} & \mkS{\underline{91.0}} & 19.9 & 11.1 \\
\bottomrule
\end{tabular}}
\end{table*}

\section{Comparison with Structure from Motion and SLAM}
\label{sec:supp_baselines}

\label{sec:supp_sfm}

\Cref{tab:supp_sfm} compares the cameras of Jojo4D with those of established structure-from-motion and SLAM pipelines, which are designed for static scenes or single videos. Unlike the main tables, it uses only the $48$ clips of the camera ablations, six each with two and four cameras on Hi4D, Perpetua, Syn4D and CHARGE.  A clip counts as a failure if the method crashes, returns no reconstruction, or leaves at least one camera without any registered frame, and failed clips enter their setting median with infinite error. If only some frames of a camera are missing, each receives the pose of the nearest registered frame of the same video. 

\paragraph{SLAM.}
DROID-SLAM~\cite{teed2021droidslam} tracks a single moving camera with learned optical flow and dense bundle adjustment. ViPE~\cite{huang2025vipe} builds on the same kind of optimization and adds metric depth priors and masks for moving objects. MetricTrail~\cite{gao2026metrictrail} tracks a monocular stream using metric depth and local bundle adjustment. All three expect one continuous video, so we concatenate the videos of all cameras into one stream. Each switch to the next video breaks the continuous camera motion that these methods assume, which largely explains their high rotation errors.

\paragraph{Classical structure from motion.}
COLMAP~\cite{schonberger2016colmap} and GLOMAP~\cite{pan2024glomap} treat all frames as one image collection, with one simple radial camera model per video, initialized with the focal length $1.2\max(W,H)$ for images of width $W$ and height $H$. We run both methods on the same verified correspondences in two variants. The first matches SuperPoint features~\cite{detone2018superpoint} with LightGlue~\cite{lindenberger2023lightglue} (SP+LG) using the standard hloc~\cite{sarlin2019hloc} setting, at most $1{,}600$ pixels on the long side and $4{,}096$ features per image. It matches each frame with the frames up to ten time steps away in the same video and with those $16$, $32$, $64$ and $128$ frames away. Across cameras, it matches the synchronized views and the ten most similar images under MegaLoc retrieval. The second variant uses RoMa~v2~\cite{edstedt2026romav2} matches at Jojo4D's matching resolution on the image pairs of Jojo4D's matching graph (\cref{sec:matching_graph_suppl}). COLMAP and GLOMAP build tracks from sparse keypoints, so we keep the dense matches only at about $2{,}500$ fixed grid points per image. Each grid point then serves as one keypoint across all image pairs. Both use COLMAP's two-view verification with up to $20{,}000$ RANSAC trials and a minimum inlier ratio of $0.1$. Finally, we run the triangulation and global refinement of COLMAP from our initialization and matches, in place of the track building and bundle adjustment of Jojo4D. This variant achieves the lowest rotation error among the baselines, yet all baselines remain substantially less accurate than Jojo4D.

\paragraph{Learned structure from motion.}
MP-SfM~\cite{pataki2025mpsfm} extends incremental structure from motion with monocular depth and normal priors. For a fair comparison, it receives the same RoMa~v2 matches on the image pairs of Jojo4D's matching graph and is initialized with the per-video intrinsics of our camera initialization. GlueMap~\cite{pan2026gluemap} combines a learned two-view model with global mapping and supports several video sequences of one scene, pairing frames sequentially within each sequence. We run it in this multi-sequence mode with one sequence per camera. Since Jojo4D exploits synchronization, we also evaluate GlueMap with all synchronized image pairs added, kept through its pair filtering and given the same priority that it assigns to consecutive frames. While this lowers its rotation error from $1.52^\circ$ to $1.06^\circ$, it remains more than ten times higher than that of Jojo4D. Because of their long runtime, we evaluate both methods only on the two-camera clips.

We run VidMap~\cite{pataki2026vidmap} on the concatenated camera videos, reversing every other video and keeping every frame as a keyframe. At camera switches, we remove sequential matches and disable the sequential link that VidMap would otherwise add during rotation averaging. Only matches between retrieved image pairs connect different videos.

\begin{table*}[t]
\caption{\textbf{Camera registration on dynamic scenes.} Jojo4D successfully registers every clip and has the lowest pose errors. Success counts clips with a pose for every camera. Because of their long runtime, we evaluate GlueMap and MP-SfM on the two-camera clips only. $R$, $t$ and FoV are in degrees, ATE$_n$ and $f$ in percent. Dark and light green mark the best and second best results.}
\label{tab:supp_sfm}
\mksetup
\footnotesize\setlength{\tabcolsep}{2.5pt}
\resizebox{\ifdim\width>\linewidth\linewidth\else\width\fi}{!}{%
\begin{tabular}{@{}lrrrrrr@{\hspace{9pt}}rrrrrr@{}}
\toprule
 & \multicolumn{6}{c}{Two cameras (24 clips)} & \multicolumn{6}{c}{Four cameras (24 clips)} \\
\cmidrule(lr){2-7}\cmidrule(l){8-13}
Method & Success$\!\uparrow$ & $R\downarrow$ & $t\downarrow$ & ATE$_n\downarrow$ & $f\downarrow$ & FoV$\downarrow$ & Success$\!\uparrow$ & $R\downarrow$ & $t\downarrow$ & ATE$_n\downarrow$ & $f\downarrow$ & FoV$\downarrow$ \\
\midrule
\multicolumn{13}{@{}l}{\textit{SLAM}} \\
DROID-SLAM (concat) & \mkB{24/24} & 29.36 & 18.44 & 13.08 & 1.92 & 0.66 & \mkB{24/24} & 66.17 & 37.09 & 26.83 & 2.11 & 0.73 \\
ViPE (concat) & \mkB{24/24} & 24.99 & 15.67 & 12.99 & 2.14 & 0.68 & \mkB{24/24} & 64.95 & 32.59 & 25.54 & 2.07 & 0.66 \\
MetricTrail (concat) & \mkB{24/24} & 29.66 & 25.71 & 16.71 & 13.11 & 6.28 & \mkB{24/24} & 72.84 & 37.86 & 24.65 & 17.50 & 8.86 \\
\midrule
\multicolumn{13}{@{}l}{\textit{Classical SfM}} \\
COLMAP (SP+LG) & \mkB{24/24} & 17.65 & 9.18 & 9.32 & 4.23 & 1.18 & \mkS{23/24} & 20.51 & 10.95 & 7.31 & 4.31 & 1.93 \\
GLOMAP (SP+LG) & 22/24 & 12.22 & 10.00 & 9.88 & 1.54 & 0.55 & 22/24 & 33.59 & 16.38 & 14.26 & 2.79 & 1.15 \\
COLMAP (RoMa v2, Jojo4D pairs) & \mkB{24/24} & 0.93 & 1.62 & 0.57 & 0.69 & 0.25 & \mkB{24/24} & 0.64 & 0.93 & 0.43 & 0.23 & 0.10 \\
GLOMAP (RoMa v2, Jojo4D pairs) & \mkB{24/24} & 0.66 & \mkS{1.29} & 3.06 & \mkS{0.26} & \mkS{0.10} & \mkB{24/24} & 1.14 & 1.10 & 1.63 & \mkS{0.22} & \mkS{0.09} \\
\midrule
\multicolumn{13}{@{}l}{\textit{Learned SfM}} \\
VidMap & \mkB{24/24} & 0.99 & 2.00 & 0.56 & 3.08 & 1.03 & \mkB{24/24} & 8.79 & 7.10 & 3.33 & 2.99 & 1.02 \\
GlueMap & \mkS{23/24} & 1.52 & 2.50 & 1.24 & 2.20 & 0.78 & \multicolumn{6}{c}{} \\
GlueMap + synchronized pairs & \mkB{24/24} & 1.06 & 2.48 & 1.19 & 1.44 & 0.52 & \multicolumn{6}{c}{} \\
MP-SfM & \mkB{24/24} & 0.84 & 3.76 & 1.28 & 1.42 & 0.52 & \multicolumn{6}{c}{} \\
\midrule
\multicolumn{13}{@{}l}{\textit{From our initialization}} \\
COLMAP (from our init.) & \mkS{23/24} & \mkS{0.40} & 1.36 & \mkS{0.54} & 0.37 & 0.14 & \mkS{23/24} & \mkS{0.59} & \mkS{0.77} & \mkS{0.35} & 0.31 & 0.12 \\
\textbf{Jojo4D (ours)} & \mkB{24/24} & \mkB{0.08} & \mkB{0.12} & \mkB{0.03} & \mkB{0.06} & \mkB{0.02} & \mkB{24/24} & \mkB{0.16} & \mkB{0.23} & \mkB{0.09} & \mkB{0.07} & \mkB{0.02} \\
\bottomrule
\end{tabular}}
\end{table*}

\section{Monocular Setting}
\label{sec:supp_single_camera}

For \cref{tab:supp_single_camera}, we process the two videos of each two-camera clip separately. Camera and point metrics, alignment and resampling follow the main paper. Sun~\etal~\cite{sun2026multiview} do not support single-camera input. We also evaluate monocular camera estimation on $18$ clips from MPI Sintel~\cite{butler2012sintel} with the evaluation protocol of MegaSaM and report the median over the clips (\cref{tab:supp_single_camera}). Jojo4D has the lowest camera errors, followed by MegaSaM, and the highest $\text{F}_{1\%}$. At the looser $2\%$ threshold, $\pi^4$ has a slightly higher F1.

\begin{table*}[t]
\caption{\textbf{Single-camera input.} Jojo4D has the lowest average pose errors and the highest average F1, and on MPI Sintel it has the lowest camera errors and the highest $\text{F}_{1\%}$. Each setting reports the median over its one-camera videos, and Avg.\ averages over the five datasets of our benchmark. $R$ and $t$ are in degrees, $\text{F}_2$ and $\text{F}_5$ in percent. The Sintel columns are medians over $18$ clips under the protocol of MegaSaM~\cite{li2025megasam} and are not part of Avg. There, $R$ and $t$ are errors between consecutive frames, $t$ and ATE are in percent of the distance between the first and last ground-truth cameras, and the F1 thresholds are $1$ and $2$\,\% of each frame's median depth. Dark and light green mark the best and second best results.}
\label{tab:supp_single_camera}
\mksetup
\resizebox{\linewidth}{!}{%
\footnotesize\setlength{\tabcolsep}{2.5pt}
\begin{tabular}{l@{\hskip5pt}cccc@{\hskip2.5pt{\color{gray!40}\vrule width 0.4pt}\hskip2.5pt}cccc@{\hskip2.5pt{\color{gray!40}\vrule width 0.4pt}\hskip2.5pt}cccc@{\hskip2.5pt{\color{gray!40}\vrule width 0.4pt}\hskip2.5pt}cccc@{\hskip2.5pt{\color{gray!40}\vrule width 0.4pt}\hskip2.5pt}cccc@{\hskip2.5pt{\color{gray!40}\vrule width 0.4pt}\hskip2.5pt}cccc@{\hskip2.5pt{\color{gray!40}\vrule width 0.4pt}\hskip2.5pt}ccccc}
\toprule
 & \multicolumn{4}{@{}c@{\hskip2.5pt{\color{gray!40}\vrule width 0.4pt}\hskip2.5pt}}{\cellcolor{mkhi4d}\textbf{Hi4D}} & \multicolumn{4}{@{}c@{\hskip2.5pt{\color{gray!40}\vrule width 0.4pt}\hskip2.5pt}}{\cellcolor{mkperp}\textbf{Perpetua}} & \multicolumn{4}{@{}c@{\hskip2.5pt{\color{gray!40}\vrule width 0.4pt}\hskip2.5pt}}{\cellcolor{mksyn}\textbf{Syn4D}} & \multicolumn{4}{@{}c@{\hskip2.5pt{\color{gray!40}\vrule width 0.4pt}\hskip2.5pt}}{\cellcolor{mkcharge}\textbf{CHARGE}} & \multicolumn{4}{@{}c@{\hskip2.5pt{\color{gray!40}\vrule width 0.4pt}\hskip2.5pt}}{\cellcolor{mkmcr}\textbf{MCRL}} & \multicolumn{4}{@{}c@{\hskip2.5pt{\color{gray!40}\vrule width 0.4pt}\hskip2.5pt}}{\cellcolor{mkavg}\textbf{Avg.}} & \multicolumn{5}{@{}c}{\cellcolor{mkdex}\textbf{Sintel}} \\
Method & \cellcolor{mkhi4dl}$R\!\downarrow$ & \cellcolor{mkhi4dl}$t\!\downarrow$ & \cellcolor{mkhi4dl}$\text{F}_2\!\uparrow$ & \cellcolor{mkhi4dl}$\text{F}_5\!\uparrow$ & \cellcolor{mkperpl}$R\!\downarrow$ & \cellcolor{mkperpl}$t\!\downarrow$ & \cellcolor{mkperpl}$\text{F}_2\!\uparrow$ & \cellcolor{mkperpl}$\text{F}_5\!\uparrow$ & \cellcolor{mksynl}$R\!\downarrow$ & \cellcolor{mksynl}$t\!\downarrow$ & \cellcolor{mksynl}$\text{F}_2\!\uparrow$ & \cellcolor{mksynl}$\text{F}_5\!\uparrow$ & \cellcolor{mkchargel}$R\!\downarrow$ & \cellcolor{mkchargel}$t\!\downarrow$ & \cellcolor{mkchargel}$\text{F}_2\!\uparrow$ & \cellcolor{mkchargel}$\text{F}_5\!\uparrow$ & \cellcolor{mkmcrl}$R\!\downarrow$ & \cellcolor{mkmcrl}$t\!\downarrow$ & \cellcolor{mkmcrl}$\text{F}_2\!\uparrow$ & \cellcolor{mkmcrl}$\text{F}_5\!\uparrow$ & \cellcolor{mkavgl}$R\!\downarrow$ & \cellcolor{mkavgl}$t\!\downarrow$ & \cellcolor{mkavgl}$\text{F}_2\!\uparrow$ & \cellcolor{mkavgl}$\text{F}_5\!\uparrow$ & \cellcolor{mkdexl}$R\!\downarrow$ & \cellcolor{mkdexl}$t\!\downarrow$ & \cellcolor{mkdexl}ATE$\!\downarrow$ & \cellcolor{mkdexl}$\text{F}_{1\%}\!\uparrow$ & \cellcolor{mkdexl}$\text{F}_{2\%}\!\uparrow$ \\
\midrule
\multicolumn{30}{l}{\textit{Feed-forward}} \\
VGGT & .76 & 1.85 & 39.0 & 65.6 & .99 & 4.43 & 48.4 & 77.2 & .36 & 2.31 & 33.4 & 59.1 & .35 & 9.31 & 29.8 & 56.8 & .45 & 3.87 & 54.2 & 80.1 & .58 & 4.35 & 41.0 & 67.7 & .177 & 2.70 & 3.77 & 13.2 & 28.8 \\
VGGT-$\Omega$ & \mkS{.29} & 1.07 & 66.5 & 87.5 & .37 & 1.17 & \mkS{68.3} & 89.3 & .15 & 1.09 & 44.9 & 69.6 & \mkS{.21} & 3.74 & 34.5 & 56.0 & \mkS{.42} & \mkS{3.02} & 63.8 & \mkS{87.1} & \mkS{.29} & 2.02 & 55.6 & 77.9 & .090 & 1.00 & 1.42 & 16.0 & 31.6 \\
Pi3X & .37 & 1.18 & 67.9 & 89.1 & .38 & 1.86 & 66.2 & \mkB{89.9} & .23 & 1.94 & 45.9 & 68.7 & .41 & 6.03 & \mkS{37.4} & 62.8 & .43 & 3.35 & 61.1 & 86.3 & .36 & 2.87 & 55.7 & 79.4 & .106 & 1.58 & 1.82 & 16.8 & 37.4 \\
$\pi^4$ & .42 & \mkS{.96} & 63.6 & 85.3 & .39 & 1.27 & 67.4 & 86.9 & .15 & 1.00 & 50.9 & 71.1 & .30 & 3.68 & 37.2 & \mkS{62.9} & \mkB{.40} & 3.14 & 61.2 & 84.3 & .33 & 2.01 & 56.1 & 78.1 & .087 & .87 & 1.01 & \mkS{20.6} & \mkB{42.2} \\
OmniX & .91 & 1.29 & \mkS{69.8} & \mkS{90.2} & .56 & 1.43 & 56.7 & 85.9 & .31 & 1.17 & 45.7 & 69.6 & .31 & 3.71 & 36.7 & 59.9 & .45 & 3.14 & 65.0 & 86.6 & .51 & 2.15 & 54.8 & 78.4 & .096 & .68 & 1.45 & 17.3 & 33.2 \\
ARROW-FF & .52 & 1.10 & 69.7 & 90.0 & .52 & 1.44 & 58.3 & 87.3 & .16 & .82 & \mkS{56.3} & \mkS{77.0} & .26 & 2.48 & 37.2 & 60.6 & .45 & 3.28 & \mkB{66.6} & \mkB{87.9} & .38 & 1.83 & \mkS{57.6} & \mkS{80.6} & .140 & .99 & 1.89 & 12.1 & 30.9 \\
\midrule
\multicolumn{30}{l}{\textit{Monocular, per video}} \\
Any4D & 2.21 & 4.64 & 15.8 & 39.7 & 2.50 & 7.65 & 26.2 & 53.1 & .89 & 8.06 & 25.7 & 53.5 & 1.54 & 30.3 & 10.6 & 26.6 & 1.11 & 12.6 & 34.9 & 69.0 & 1.65 & 12.6 & 22.6 & 48.4 & .540 & 6.64 & 8.83 & 11.1 & 25.5 \\
V-DPM & .95 & 3.26 & 43.6 & 72.3 & 1.61 & 4.84 & 33.7 & 60.6 & 1.09 & 11.0 & 28.5 & 52.5 & 1.79 & 38.2 & 12.5 & 31.8 & .75 & 10.6 & 49.4 & 75.5 & 1.24 & 13.6 & 33.5 & 58.5 & .590 & 4.58 & 8.68 & 11.2 & 23.0 \\
MegaSaM & .45 & .99 & 45.2 & 68.6 & \mkS{.26} & \mkS{.60} & 25.5 & 57.8 & \mkS{.06} & \mkS{.28} & 39.5 & 63.0 & .42 & \mkS{1.48} & 18.3 & 40.2 & .45 & 3.06 & 41.3 & 69.3 & .33 & \mkS{1.28} & 34.0 & 59.8 & \mkS{.022} & \mkS{.20} & \mkS{.28} & 16.2 & 30.0 \\
\midrule
\textbf{Jojo4D (ours)} & \mkB{.07} & \mkB{.09} & \mkB{80.0} & \mkB{92.5} & \mkB{.09} & \mkB{.28} & \mkB{69.5} & \mkS{89.5} & \mkB{.02} & \mkB{.06} & \mkB{66.1} & \mkB{84.9} & \mkB{.04} & \mkB{.29} & \mkB{61.7} & \mkB{77.6} & \mkS{.42} & \mkB{2.90} & \mkS{65.8} & 83.6 & \mkB{.13} & \mkB{.73} & \mkB{68.6} & \mkB{85.6} & \mkB{.021} & \mkB{.11} & \mkB{.14} & \mkB{35.6} & \mkS{39.3} \\
\bottomrule
\end{tabular}}
\end{table*}

\section{Failure Cases}
\label{sec:supp_failures}

\begin{figure*}[t]
\centering
\includegraphics[width=\textwidth]{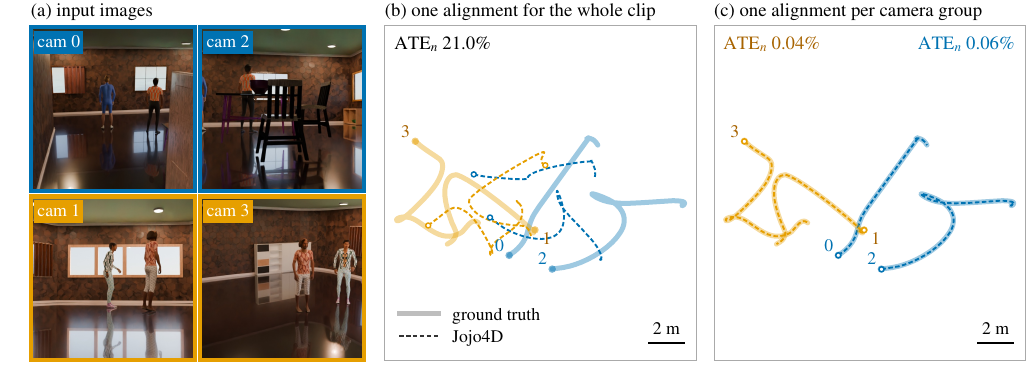}
\caption{\textbf{Limited view overlap.} On the four-camera Perpetua clip diningroom03, cameras $0$ and $2$ (blue) never observe a common part of the scene with cameras $1$ and $3$ (orange). (a) Input images at the first frame. (b) Top view of the camera trajectories of Jojo4D (dashed) and the ground truth (solid) after one similarity alignment for the whole clip, as for ATE$_n$. (c) The same trajectories with one alignment per camera group. Each group is reconstructed accurately, but the two groups are placed incorrectly relative to each other.}
\label{fig:perpetua_overlap}
\end{figure*}

\paragraph{Limited view overlap.}
Jojo4D places the camera trajectories relative to each other through matches between different cameras at the same instant. Dense matching is robust to large baselines, but it fails when cameras see the same surfaces only from very different viewpoints or share no view at all. Some rooms of the four-camera Perpetua setting contain such camera groups. Jojo4D then reconstructs each group accurately but places the groups incorrectly relative to each other (\cref{fig:perpetua_overlap}). This causes all three camera failures of Jojo4D on the benchmark (ATE$_n$ above $10\%$) and dominates the means in \cref{tab:supp_cameras_mean}.

These errors already originate in the initialization, and without matches between the groups, the bundle adjustment cannot correct them. Initialized from $\pi^4$ instead, Jojo4D places the groups correctly on two of the three clips. On the third, the two groups never share a view, and every method fails. A future remedy is to detect such groups from the number of reliable matches between cameras at the same instant and to report them as separate reconstructions.

\begin{figure*}[t]
\centering
\includegraphics[width=\textwidth]{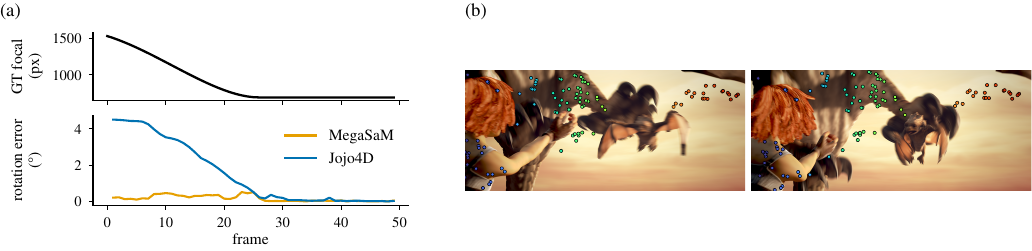}
\caption{\textbf{Failure cases on MPI Sintel.} (a) On the zooming clip cave\_2, the ground-truth focal length (top) halves during the first $26$ frames. The rotation error of Jojo4D between consecutive frames (bottom) is high during the zoom and drops to the level of MegaSaM once the focal length stays fixed. (b) Matches between consecutive frames of temple\_3 that pass our certainty and round-trip checks, with $100$ drawn. RoMa assigns low certainty to the distant sky, so most retained matches lie on the moving character and dragon.}
\label{fig:sintel_failures}
\end{figure*}

\paragraph{Changing intrinsics.}
Jojo4D assumes one pinhole camera with fixed intrinsics per video. Zoom, digital stabilization that crops and rescales each frame, and strong lens distortion violate this model, and the camera poses absorb the remaining error. On MPI Sintel, the zooming clip cave\_2 causes our largest rotation error, which rises exactly while the focal length changes (\cref{fig:sintel_failures}a). Per-frame focal lengths or a distortion model in the bundle adjustment would address this, but the additional freedom can destabilize the optimization when the intrinsics are in fact constant. If the input is known to contain videos with time-variant intrinsics, this option can be enabled per camera. Enabling per-frame focal lengths on cave\_2 lowers its RPE-R from $2.28^\circ$ to $0.05^\circ$.

\paragraph{Strong motion.}
The tracks that constrain the cameras of a video over time must lie on static parts of the scene. When moving objects dominate the images and the static background is distant or hard to match, most of the retained matches lie on the moving objects (\cref{fig:sintel_failures}b) causing the camera estimates to degrade. In addition, strong motion blur and very fast camera rotation reduce the number of correct matches between frames, so tracks break and the cameras of a video lose their connection over time.

\paragraph{Static single cameras.}
A camera that does not move sees the scene from one viewpoint, so its images provide no parallax. With a single static camera, neither the depth of the scene nor the motion of dynamic objects can be triangulated, and the reconstruction falls back on the depth predicted by the feed-forward model.

\section{Additional Qualitative Examples}
\label{sec:supp_qualitative}

\begin{figure*}[t]
    \centering
    \includegraphics[width=\linewidth]{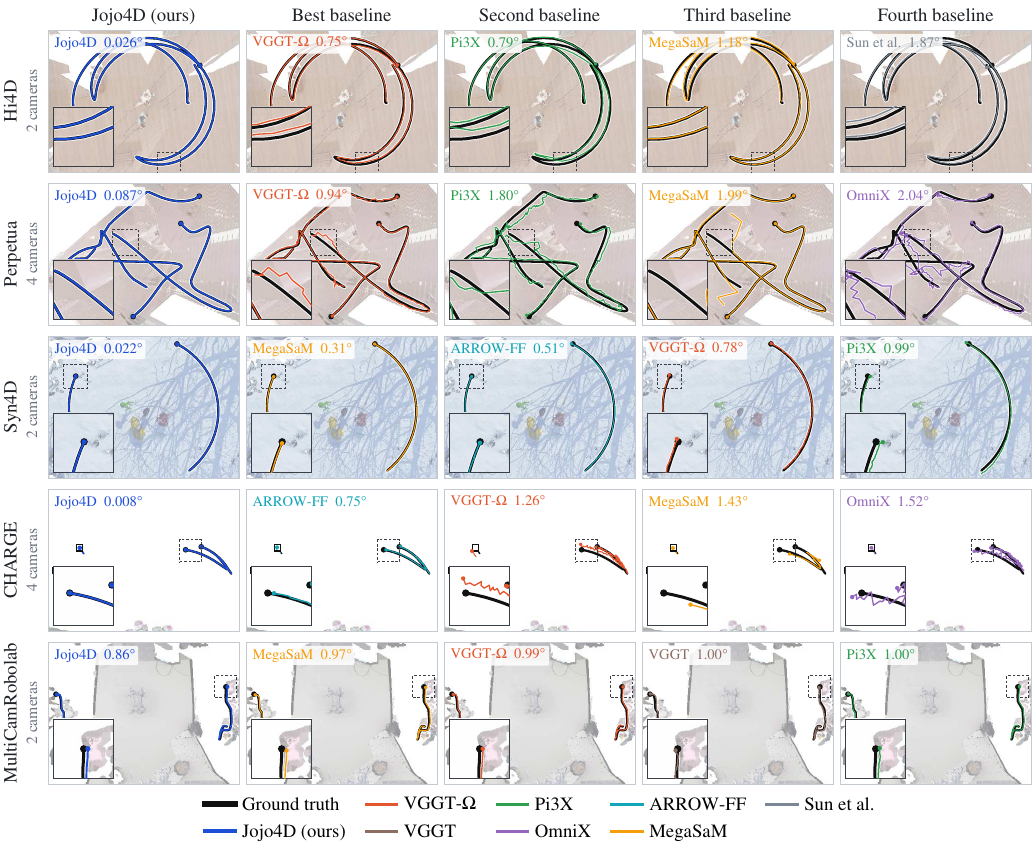}
    \vspace{-0.5cm}
    \caption{\textbf{Additional camera trajectories.} Camera trajectories estimated by Jojo4D (left column) and by the four baselines with the lowest RPE-R on each scene (remaining columns, best first), aligned to the ground truth (black). The number in each panel is the RPE-R in degrees.}
    \label{fig:supp_qual_traj}
\end{figure*}

\begin{figure*}[t]
    \centering
    \includegraphics[width=\linewidth]{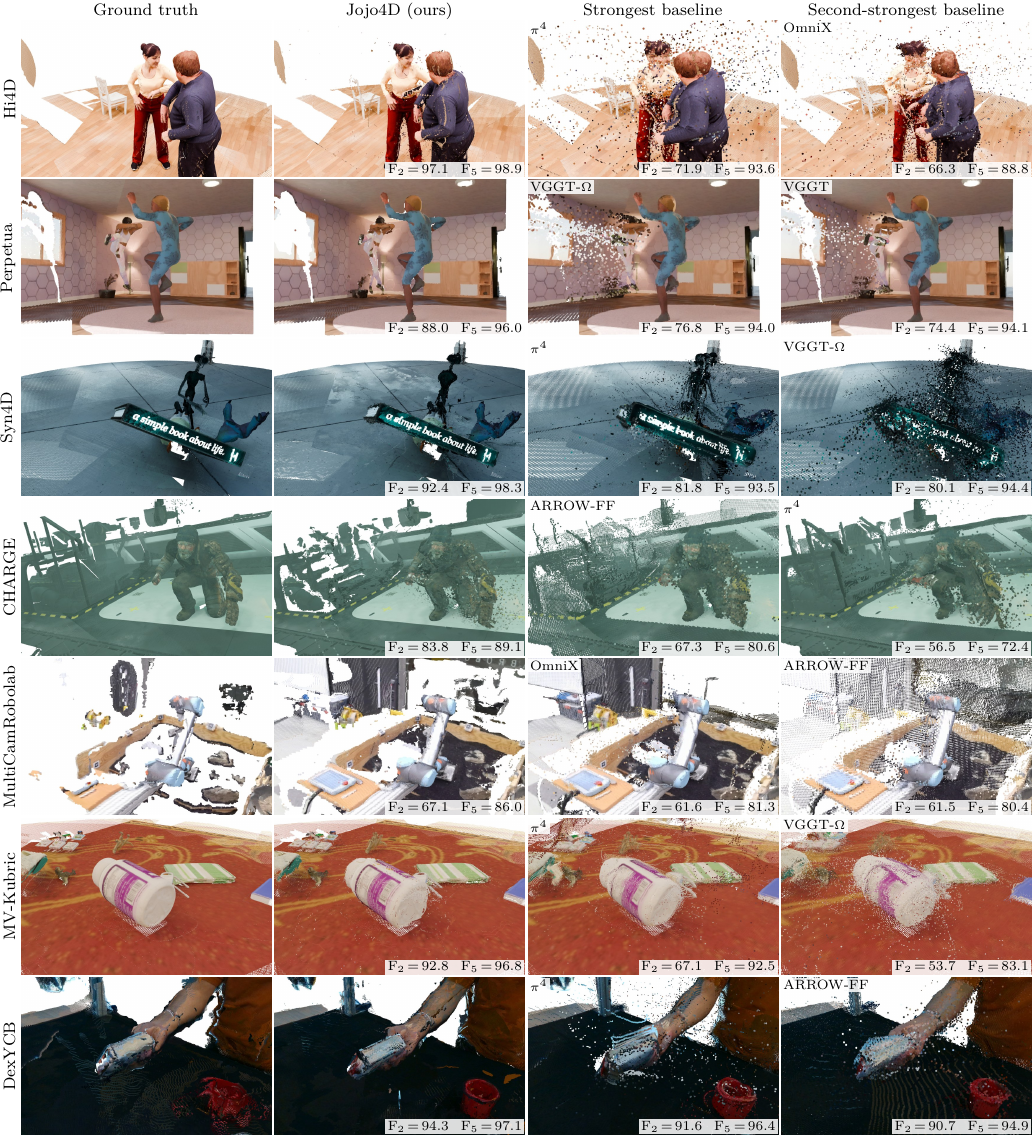}
    \caption{\textbf{Additional point clouds.} Each row shows one instant of a clip from a single view, with the ground truth, Jojo4D and the two baselines with the highest $\text{F}_2$ score on that clip. Tags report $\text{F}_2$ and $\text{F}_5$ (\%) over the whole clip.}
    \label{fig:supp_qual_points}
\end{figure*}

\Cref{fig:supp_qual_traj,fig:supp_qual_points} show additional qualitative examples of camera estimation and point cloud reconstruction on the benchmark datasets.

\begin{figure*}[t]
    \centering
    \includegraphics[width=\linewidth]{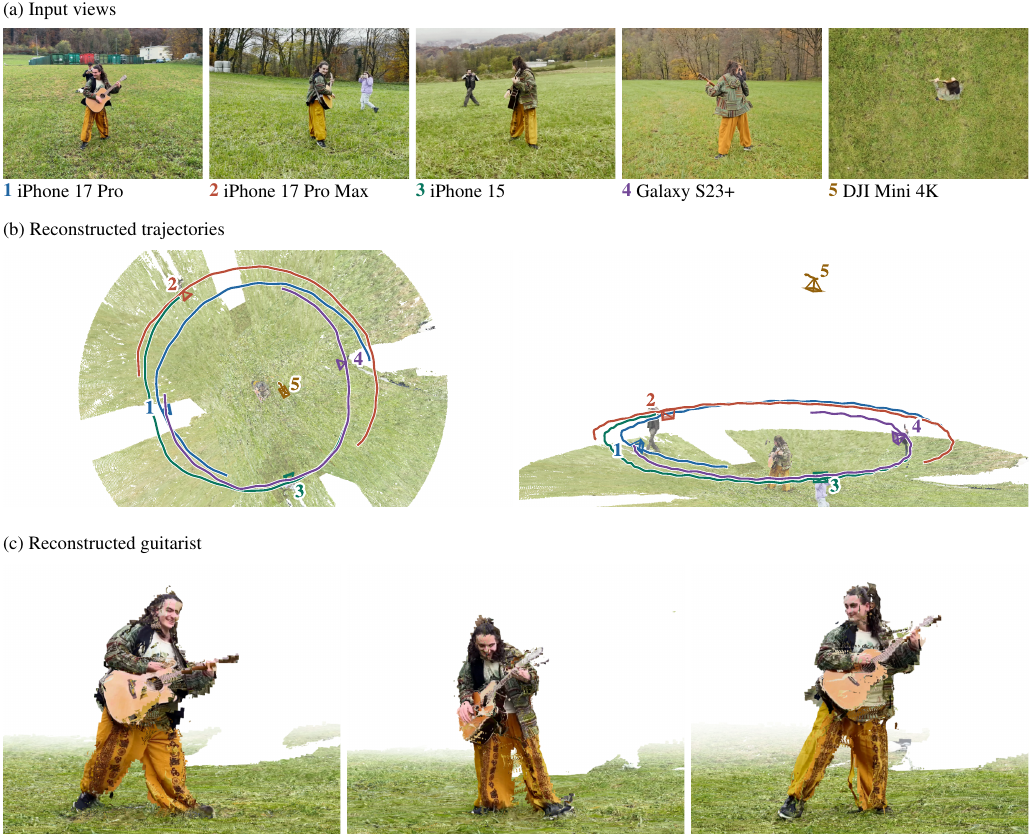}
    \caption{\textbf{Reconstructing a guitarist from handheld and aerial footage.} (a) One frame of each device. (b) The reconstructed camera trajectories from above and from the side. The phones move around the guitarist while the drone stays above them. Each path traces one device over the full recording, with its pose marked at the instant shown in (a). (c) The reconstructed guitarist at three instants along the camera orbit.}
    \label{fig:supp_outdoor_capture}\label{fig:supp_real_demo}
\end{figure*}

\label{sec:supp_real_demo}
Beyond the benchmark, we tested Jojo4D on an outdoor guitar performance filmed without a calibrated rig. Four handheld phones followed the guitarist while a drone filmed from above (\cref{fig:supp_outdoor_capture}). The compressed phone footage includes camera shake and motion blur, and the views differ in exposure and color. We aligned the five videos by hand to a clap at the start of the recording, so small timing offsets may remain.

From $598$ frames per camera at $30$\,fps, Jojo4D estimates all five camera trajectories and reconstructs one point cloud per time step. The supplementary webpage shows the reconstruction in motion alongside the five recordings.

\end{document}